\documentclass[final,3p,times]{elsarticle}

\usepackage{amssymb}
\usepackage{amsthm}
\usepackage{amsmath}
\usepackage{amsfonts}
\usepackage{bm}
\usepackage{graphicx}
\usepackage{float}
\usepackage{subfig}
\usepackage{slantsc}
\usepackage{multirow}
\usepackage[normalem]{ulem}
\usepackage{marginnote}
\usepackage{xcolor} 
\usepackage[normalem]{ulem}

\journal{}

\begin{document}

\begin{frontmatter}



\title{Augmented neural forms with parametric boundary-matching operators for solving ordinary differential equations}


\author{Adam D. Kypriadis}
\ead{a.kypriadis@uoi.gr}
\author{Isaac E. Lagaris}
\ead{lagaris@uoi.gr}
\author{Aristidis Likas}
\ead{arly@uoi.gr}
\author{Konstantinos E. Parsopoulos\corref{cor1}}
\ead{kostasp@uoi.gr}
\cortext[cor1]{Corresponding author}
\affiliation{organization={Department of Computer Science and Engineering, University of Ioannina},
city={GR-45110 Ioannina},
country={Greece}}


\begin{abstract}
\noindent
Approximating solutions of ordinary and partial differential equations constitutes a significant challenge. Complementarily to numerical analysis methods, neural forms have emerged as a valuable solution methodology to this problem. Based on functional expressions that inherently depend on neural networks, neural forms are specifically designed to precisely satisfy the prescribed initial or boundary conditions of the problem, while providing the approximate solutions in closed form. Departing from the fundamentally important class of ordinary differential equations, the present work aims to refine and validate the neural forms methodology, paving the ground for further developments in more challenging fields. The main contributions are as follows. 
First, it introduces a formalism for systematically crafting proper neural forms with adaptable boundary matches that are amenable to optimization. Second, it describes a novel technique for converting problems with Neumann or Robin conditions into equivalent problems with parametric Dirichlet conditions. Third, it outlines a method for determining an upper bound on the absolute deviation from the exact solution. The proposed augmented neural forms approach was tested on a set of diverse problems, encompassing first- and second-order ordinary differential equations, as well as first-order systems. Stiff differential equations have been considered as well. The resulting solutions were subjected to assessment against existing exact solutions, solutions derived through the common penalized neural method, and solutions obtained via contemporary numerical analysis methods. 
The reported results demonstrate that the augmented neural forms not only satisfy the boundary and initial conditions exactly, but also provide closed-form solutions that facilitate high-quality interpolation and controllable overall precision. These attributes are essential for expanding the application field of neural forms to more challenging problems that are described by partial differential equations.
\end{abstract}



\begin{keyword}
Ordinary differential equations \sep neural networks \sep neural forms
\end{keyword}

\end{frontmatter}



\section{Introduction}
\label{sec:intro}

Neural networks have been successfully employed for the  solution of diverse problems in science and engineering. Their universal approximation property, when using activation functions of specific type~\mbox{\cite{CYBENKO_1989,HORNIK_1989,PARK_1991},} renders them particularly suitable for modeling tasks. Within this framework, neural networks have been employed to model solutions for ordinary differential equations (ODEs) and partial differential equations (PDEs) as well. 
Although the first relevant neural methods appeared in the mid-90s~\cite{Lee_1990,MEADE_1994,DISSANAYAKE_1994,LAGARIS_1997, LAGARIS_1998,LAGARIS_2000}, their utilization remained limited for a period of about 20 years. This stagnation can be attributed to several factors, including the lack of widely accessible specialized software, as well as the limited computing power available at the time.
The recent advancements in deep learning, the emergence of sophisticated software platforms (such as TensorFlow, Keras, and PyTorch) and the advent of high-performance computing technologies (including GPUs and parallel multicore multiprocessor clusters) have led to a resurgence of interest in neural differential equation solvers. Thus, a number of alternative techniques, based on a plethora of neural network types, have been developed and are currently  available~\cite{MCFALL_2009, MALL_2016, MALL_2017, BLEKAS_2017, BERG_2018, SIRIGNANO_2018, RAISSI_2019, LAGARI_2020, BLEKAS_2020, MORTARI_2020,AARTS_2001}, along with new finite difference based  solvers such as ODIL~\cite{ODIL}.

The primary goal of neural methods is the construction of a parametric model that can be tuned to produce a solution with a prescribed degree of accuracy. This is carried out by minimizing a properly designed \textit{error function} (also known as \textit{loss} or \textit{cost function}) within the space of the model parameters. 
In addition, differential equations are accompanied by boundary and/or initial conditions that  must be satisfied. 
From here on we will use the term \textit{conditions} to refer either to boundary or initial conditions, unless otherwise specified. 
It is important to realize that a trial solution which fails to meet the prescribed conditions cannot be considered as an adequate approximation to the true solution of the problem under consideration.

A common approach treats the conditions as equality constraints when minimizing the error function. In this case, constrained optimization methods (interior point, active set, augmented Lagrangian, etc.) that exactly satisfy the constraints should be used. However, their application appears to be complicated and time-consuming as well.  Alternatively,  equality constraints may be approximately treated via penalty methods using unconstrained optimization. Penalty terms are added to the original error function in an attempt to inhibit constraint violations. 
In fact, this is by far the most commonly followed approach, perhaps due to reasons of implementation convenience and simplicity of interpretation.
If the parametric trial solution is crafted in such a way so as to satisfy exactly the specified conditions regardless of the parameter values, 
the constraints are eliminated, thereby allowing for the application of efficient unconstrained optimization methods. 
In the case of ODEs, as well as of PDEs that are defined inside rectangular domains, this approach is preferable because the exact satisfaction of the conditions allows the method to attain higher accuracy as pointed out in~\cite{MULLER_2022}. 
In contrast, for PDEs within domains of complex geometry, the design of proper trial solutions that satisfy prescribed conditions is challenging, and thus the alternative approximate penalty approach has been frequently preferred.  

In cases where the conditions are treated as constraints, the trial solution typically consists of a single neural network. On the other hand, the constraint elimination approach requires more elaborate functional constructs. To this end, \textit{neural forms} were indirectly introduced in~\cite{LAGARIS_1998} and further refined in~\cite{LAGARI_2020}. In both cases, a rigid polynomial expression was used to account for the conditions. An alternative neural form type was introduced in~\cite{MCFALL_2009}, involving a function representing the distance of a domain point from the boundary, an idea that was further extended in~\cite{BERG_2018} for the case of Dirichlet conditions. 
Neural forms that exactly satisfy the ODE conditions are obviously not unique. Therefore, the question concerning the existence of a possibly optimal choice naturally arises. This is one of the issues addressed here by introducing the \textit{augmented neural forms} that are presented in the following paragraphs. 

The contributions of the present work can be summarized as follows:
\begin{enumerate}[(a)]
\item A flexible, parametric neural form replaces the rigid polynomial boundary or initial match used in~\cite{LAGARIS_1998,LAGARI_2020}.
\item A technique is introduced for transforming Neumann or Robin to equivalent parametric Dirichlet conditions.
\item A methodology is introduced for evaluating the quality of the approximate solution by determining a reliable upper bound of its absolute deviation from the exact solution.
\end{enumerate}
The performance of the proposed approach was assessed on first- and second-order ODEs, as well as on first-order systems, subject to Dirichlet, Neumann, mixed Dirichlet-Neumann, Robin, and Cauchy conditions. The obtained solutions were compared to existing exact solutions, to solutions obtained by the penalized neural method, and to solutions obtained by the \texttt{ode113} numerical solver of \textsc{Matlab}\textsuperscript{\textregistered}~\cite{matlabodesuite}, which is based on a state-of-the-art variable-step, variable-order Adams-Bashforth-Moulton method. The conducted analysis verified that the proposed augmented neural forms can provide robust, closed-form solutions with exact satisfaction of the prescribed conditions, along with excellent generalization performance.

The rest of the article is organized as follows: Section~\ref{ODE_SOLVE} briefly reviews neural ODE solvers, and introduces the neural forms concept. Section~\ref{OPT_NF} elaborates on the augmented neural forms framework for first- and second-order ODEs. Section~\ref{REDUCT} analyzes the transformation of Neumann or Robin conditions to equivalent parametric Dirichlet conditions. Section~\ref{UPPER_BOUND} provides a comprehensive analysis of obtaining a reliable upper bound for the absolute deviation between the approximate and the exact solution. Section~\ref{sec:ExperSetting} describes in detail the experimental setup. 
Section~\ref{sec:results} presents the obtained solutions and a comparison thereof, as mentioned above.
Finally, Section~\ref{CONCLUDE} provides an overview of the contributions of this work, along with comments, conclusions, and suggestions for further research.


\section{Neural forms for solving ordinary differential equations} 
\label{ODE_SOLVE}

\noindent
Consider the general form of a second-order ODE:
\begin{equation}
\mathcal{L}_x \psi(x) = f(x), \quad x \in [a,b],
\label{GENDE}
\end{equation}
with boundary conditions (BCs):
\begin{equation}
\mathcal{B}_a \psi(x) |_a = \xi_a, \qquad \mathcal{B}_b \psi(x) |_b = \xi_b, 
\label{BCIC_1}
\end{equation}
or initial conditions (ICs):
\begin{equation}
\psi(x) |_a = \xi_0, \qquad \psi'(x) |_a = \xi_1,
\label{BCIC_2}
\end{equation}
where $\mathcal{L}_x$ is a second-order differential operator, $\mathcal{B}_a$ and $\mathcal{B}_b$ are the corresponding boundary condition operators, and $\xi$'s are prescribed constants. An appropriate neural model for a trial solution $\Psi_{T}$ that approximates the exact solution $\psi(x)$ can be designed in many ways. The simplest and most common approach sets the trial solution equal to a neural network $N(x,w)$ with weights $w$, i.e.: 
\begin{equation}
\psi(x) \approx \Psi_T(x,w) \triangleq N(x,w), 
\label{eq:singleNeuralNetwork}
\end{equation}
and requires that both the ODE and the accompanying conditions are satisfied. This is accomplished by finding suitable values for the weights  $w$ so that:
\begin{equation}
\int_a^b dx \, \left[ \mathcal{L}_x \Psi_T(x,w)-f(x) \right]^2 = 0,
\label{RESID}
\end{equation}
subject to either:
\begin{equation}
\mathcal{B}_a \Psi_T(x,w) |_a = \xi_a, \qquad \mathcal{B}_b \Psi_T(x,w) |_b = \xi_b, \hskip 10 pt \hbox{(case of BCs)}
\label{CONSTRAINTS_1}
\end{equation}
or:
\begin{equation}
\Psi_T(x,w) |_a = \xi_0, \qquad \Psi'_T(x,w) |_a = \xi_1.
 \hskip 18 pt \hbox{(case of ICs)}
\label{CONSTRAINTS_2}
\end{equation}
This problem is typically addressed by minimizing the left-hand side of Eq.~(\ref{RESID}) under the conditions of Eqs.~(\ref{CONSTRAINTS_1})~or~(\ref{CONSTRAINTS_2}), using constrained optimization methods. 
The \textit{baseline neural method}  \cite{AARTS_2001,RAISSI_2019}, adds penalties to the error function which assumes the following form:
\begin{equation}
E(w) \triangleq \int_a^b dx \, \left[ \mathcal{L}_x \Psi_T(x,w)-f(x) \right]^2 + \zeta 
\begin{cases}
  [(\mathcal{B}_a \Psi_T(x,w) |_a - \xi_a)^2+(\mathcal{B}_b \Psi_T(x,w) |_b - \xi_b)^2], & \text{for BCs},\\ 
   [(\Psi_T(x,w) |_a - \xi_0)^2+(\Psi'_T(x,w) |_a - \xi_1)^2], & \text{for ICs},
\end{cases}
\label{eq:baseline_error}
\end{equation}
and employs unconstrained optimization, with $\zeta > 0$ being a positive penalty-regulating coefficient.\\
Alternatively, neural forms cast the trial solution as a sum of two terms:
\begin{equation}
\psi(x) \approx \Psi_{T}(x,w) \triangleq A(x) + G(x,w),
\label{BASIC_MODEL}
\end{equation}
where $A(x)$ is a smooth function matching the boundary or initial conditions, henceforth referred to as the \textit{boundary match} or \textit{initial match}, respectively. Moreover, $G(x,w)$ is a parametric function typically depending on a neural network with null contribution to the  problem's conditions, i.e.:
\[
\begin{array}{llcll}
\forall w, & \mathcal{B}_a G(x,w) |_a = 0, & \hbox{and} & \mathcal{B}_b G(x,w) |_b = 0,  & \text{for BCs},\\ [4pt]
 & & \text{or} & & \\ [4pt]
\forall w, & G(x,w) |_a = 0, & \hbox{and}  & G'(x,w) |_a = 0, & \text{for ICs}.
\end{array}
\]
Note that, with the above formulation, the trial solution satisfies the prescribed ODE conditions by construction. 

The functions $A(x)$ and $G(x,w)$ are closely related, and their design is based on a common matching operator. Let $\mathcal{P}_x$ be a boundary matching operator defined as:
\begin{equation*}
\mathcal{P}_x \psi(x)  \triangleq A(x).
\end{equation*}
We can easily verify that this operator is not unique. In fact, there is an infinite number of alternative forms. For instance, for the case of Dirichlet boundary conditions, $\mathcal{P}_x$ may be given by a linear relation as follows:
\begin{equation}
\mathcal{P}_x \psi(x) \triangleq \psi(a) \, \frac{x-b}{a-b} + \psi(b) \, \frac{x-a}{b-a},
\label{RIGID_1}
\end{equation}
or as any of the nonlinear alternatives:
\begin{equation}
\mathcal{P}_x \psi(x) \triangleq \psi(a) \, \left( \frac{x-b}{a-b} \right)^n + \psi(b) \, \left( \frac{x-a}{b-a} \right)^n, \quad n =2,3,\ldots
\label{RIGID_2}
\end{equation}
Starting from the identity:
\begin{equation*}
\psi(x) = \mathcal{P}_x \psi(x) + \left( 1-\mathcal{P}_x \right) \psi(x),
\label{IDENTITY}
\end{equation*}
a plausible trial solution based on a neural network $N(x,w)$ may be cast as:
\begin{eqnarray*}
\Psi_T(x,w) &=& \mathcal{P}_x \psi(x) + (1-\mathcal{P}_x) N(x,w) \nonumber \\ 
 &=& N(x,w) + \mathcal{P}_x \left( \psi(x)-N(x,w) \right),
\label{NFTRIAL}
\end{eqnarray*} 
and, hence, the $G(x,w)$ function in Eq.~(\ref{BASIC_MODEL}) can be constructed as:
\begin{equation*}
G(x,w)= \left( 1-\mathcal{P}_x \right) N(x,w).
\end{equation*}
Then, an approximate solution of the ODE is obtained by minimizing the error function:
\begin{equation}
E(w) \triangleq \int_a^b dx \, \left[ \mathcal{L}_x \Psi_T(x,w)-f(x) \right]^2,
\label{ERROR_NF}
\end{equation}
with respect to the neural network's weight vector $w$. 


\section{Augmented neural forms for different types of  boundary and initial conditions} 
\label{OPT_NF}

\noindent
As illustrated in Eqs.~(\ref{RIGID_1}) and~(\ref{RIGID_2}), the function $A(x)$ is not unique. Therefore, we can take a further step by representing it as a parametric function $A(x,\theta)$ that satisfies the problem's conditions for any values of the parameters vector $\theta$. The error function of Eq.~(\ref{ERROR_NF}) can then be written as:
\begin{equation}
E(w,\theta) \triangleq \int dx \left[ \mathcal{L}_x \Psi_T(x,w,\theta)-f(x) \right]^2, 
\label{FULLYPARAM}
\end{equation}
and minimized with respect to both $w$ and $\theta$, leading to an optimized choice for the match $A(x,\theta)$ and, consequently, for the trial neural form solution. In the following paragraphs, we elaborate on this concept for the cases of first-order ODEs and systems, as well as for second-order ODEs.


\subsection{Augmented neural forms for first-order ordinary differential equations and systems}
\label{1stORDER}

\noindent
Let the first-order ODE:
\begin{equation}
\mathcal{L}_x \psi(x) = f(x), \quad x \in [a,b],
\label{FIRSTORDER}
\end{equation}
with the initial condition $\psi(a) = \xi_a$. 
The functions $A(x,\theta)$, $G(x,w,\theta)$, and $\Psi_T(x,w,\theta)$, are then given by:
\begin{eqnarray*}
A(x,\theta) &=& \mathcal{P}_x(\theta)\psi(x) ~~\triangleq~~ \psi(a) \left( F(x,\theta)-F(a,\theta)+1 \right),\\
G(x,w,\theta) &=& (1-\mathcal{P}_x (\theta)) N(x,w),\\
\Psi_T(x,w,\theta) &=& A(x,\theta)+G(x,w,\theta),
\end{eqnarray*}
where $F(x,\theta)$ is a smooth, bounded function for  $x\in[a,b]$. Note that $F(x,\theta)$ may be a neural network, i.e., $F(x,\theta) = N_1(x,\theta)$. The associated error function is then given by:
\begin{equation*}
E(w,\theta) \triangleq \int dx \left[ \mathcal{L}_x\Psi_T(x,w,\theta)-f(x) \right]^2.
\end{equation*}
Generalizing the concepts above, let a system of first-order ODEs:
\[
\mathcal{L} \Psi(x) = f(x),  
\]
with $\Psi(x), f(x) \in R^n$, where:
\[
\mathcal{L} \Psi(x) \triangleq \frac{d}{dx}\Psi(x) + F(x,\Psi), 
\]
and $F \in R^n$. Then, the initial matches are defined as: 
\begin{equation*}
A_i(x,\theta_i) ~=~ \mathcal{P}_i (\theta_i) \psi_i(x) ~=~ \psi_i(a) \left( \tilde{N}_i(x,\theta_i)-\tilde{N}_i(a,\theta_i)+1 \right), \quad i=1,2,\ldots,n,
\end{equation*}
and, respectively, we have:
\begin{equation*}
G_i(x,\theta_i,w_i) = \left( 1-\mathcal{P}_i (\theta_i) \right) N_i(x,w_i),
\end{equation*}
where $\tilde{N}_i(x,\theta_i)$ denotes the corresponding neural network for the $i$-th initial match, and $N_i(x,w_i)$ denotes the network for the $G_i$ function associated with the $i$-th trial solution. Hence:
\begin{equation*}
\Psi_T^{[i]}(x,w,\theta) = A_i(x,\theta_i) + G_i(x,w_i,\theta_i), \quad \forall i=1,2,\ldots,n,
\end{equation*}
are the trial solutions for the above system of simultaneous equations.


\subsection{Augmented neural forms for second-order ordinary differential equations}
\label{2ndORDER}

\noindent
Second-order ODEs may be accompanied by Dirichlet, Neumann, mixed Dirichlet-Neumann, Robin, or Cauchy conditions. Each condition type corresponds to a different boundary matching operator, hence, we examine each case separately.

\subsubsection{Case of Dirichlet conditions}

\noindent
Consider the following ODE:
\begin{equation}
\mathcal{L}_x \psi(x) = f(x), \quad x \in [a,b],
\label{DIR_ODE}
\end{equation}
with Dirichlet boundary conditions:
\begin{equation*}
\psi(a)=\xi_a, \qquad \psi(b)=\xi_b,
\end{equation*}
where $\mathcal{L}_x$ is a second-order differential operator. Also, consider a function $A_\psi(x)$ defined for all $x \in [a,b]$, satisfying 
$
A_\psi(a) = \psi(a)$ and $A_\psi(b) = \psi(b).
$
Then, there are diverse ways to construct $A_\psi(x)$. For example, it may be crafted as:
\begin{equation}
A_\psi(x) = \psi(a) \, \left (\frac{x-b}{a-b}+F(x)-F(a)\frac{x-b}{a-b}-F(b)\frac{x-a}{b-a}\right ) + 
\psi(b) \, \left (\frac{x-a}{b-a}+H(x)-H(a)\frac{x-b}{a-b}-H(b)\frac{x-a}{b-a}\right ),
\label{BASIC_A}
\end{equation}
with $F(x)$ and $H(x)$ be smooth, bounded functions in $x$. Let us employ the form of Eq.~(\ref{BASIC_A}) with:
\begin{equation}
F(x) = N_1(x,\theta_1), \qquad H(x) = N_2(x,\theta_2),
\end{equation} 
where $N_1(x,\theta_1)$ and $N_2(x,\theta_2)$ are neural networks with weight vectors $\theta_1$ and $\theta_2$, respectively.
Then, the associated Dirichlet boundary match operator $\mathcal{P}_x$ is given as: 
\begin{align}
\mathcal{P}_x(\theta_1,\theta_2) \psi(x) ~\triangleq~ & \psi(a) \, \left (\frac{x-b}{a-b}+N_1(x,\theta_1)-N_1(a,\theta_1)\frac{x-b}{a-b}-N_1(b,\theta_1)\frac{x-a}{b-a}\right ) ~ + \nonumber \\
 & \psi(b) \, \left (\frac{x-a}{b-a}+N_2(x,\theta_2)-N_2(a,\theta_2)\frac{x-b}{a-b}-N_2(b,\theta_2)\frac{x-a}{b-a}\right ).
\end{align}
Thus, we have:
\begin{equation} 
G(x,\theta_1,\theta_2,w) = \left[ 1-\mathcal{P}_x(\theta_1,\theta_2) \right] N(x,w),
\label{DIR_G}
\end{equation}
yielding the trial solution:
\begin{equation} 
\Psi_T(x,\theta_1,\theta_2,w) = \mathcal{P}_x(\theta_1,\theta_2) \, \psi(x)  +  G(x,\theta_1,\theta_2,w),
\label{DIR_PSI}
\end{equation}
which is a proper neural form for the case of Dirichlet conditions.


\subsubsection{Case of mixed Dirichlet-Neumann conditions}

\noindent
Let us now assume the ODE of Eq.~(\ref{DIR_ODE}) with mixed conditions:
\begin{equation*}
\psi(a)=\xi_a ~~\textrm{(Dirichlet)}, \qquad \psi'(b)=\xi_b ~~\textrm{(Neumann)}.
\end{equation*}
A plausible boundary match is given as:
\begin{align*}
\mathcal{P}_x(\theta_1,\theta_2) \psi(x) ~\triangleq~ & \psi(a) \, \left[ 1+N_1(x,\theta_1)-N_1(a,\theta_1)-(x-a)N'_1(b,\theta_1)\right] ~+  \\
 & \psi'(b) \, \left[  (x-a)+N_2(x,\theta_2)-N_2(a,\theta_2)-(x-a)N'_2(b,\theta_2)\right].
 \label{PREF_MIXED}
\end{align*}
The function $G(x,\theta_1,\theta_2,w)$ and the trial solution $\Psi_T(x,\theta_1,\theta_2,w)$ for this case are given as in Eqs.~(\ref{DIR_G}) and~(\ref{DIR_PSI}), respectively.


\subsubsection{Case of Neumann conditions}

\noindent
Let the ODE of Eq.~(\ref{DIR_ODE}) with Neumann conditions:
\begin{equation*}
\psi'(a) = \xi_a, \qquad \psi'(b) = \xi_b.
\end{equation*} 
The boundary match operator assumes the following form:
\begin{align*} 
\mathcal{P}_x(\theta_1,\theta_2) \psi(x) ~\triangleq~ & \psi'(a) \, \left ( \frac{(x-b)^2}{2(a-b)}+N_1(x,\theta_1)-N'_1(a,\theta_1)\frac{(x-b)^2}{2(a-b)}-N'_1(b,\theta_1)\frac{(x-a)^2}{2(b-a)} \right ) ~+ \nonumber \\
 & \psi'(b) \, \left ( \frac{(x-a)^2}{2(b-a)}+N_2(x,\theta_2)-N'_2(a,\theta_2)\frac{(x-b)^2}{2(a-b)}-N'_2(b,\theta_2)\frac{(x-a)^2}{2(b-a)} \right ).
\end{align*}
Again, the functions $G(x,\theta_1,\theta_2,w)$, $\Psi_T(x,\theta_1,\theta_2,w)$, retain the form of Eqs.~(\ref{DIR_G}) and~(\ref{DIR_PSI}), respectively.


\subsubsection{Case of Cauchy conditions}

\noindent
Cauchy conditions are initial conditions of the form:
\begin{equation*}
\psi(a) = \xi_0, \hbox{\ and\ } \psi'(a) = \xi_1.
\end{equation*}
The following initial matching operator is admissible:
\begin{align*}
\mathcal{P}_x(\theta_1,\theta_2) \psi(x) ~\triangleq~ & \psi(a) \, \left[ 1+N_1(x,\theta_1)-N_1(a,\theta_1)-(x-a)N'_1(a,\theta_1)\right] ~+ \nonumber \\
 & \psi'(a) \, \left[ (x-a)+N_2(x,\theta_2)-N_2(a,\theta_2)-(x-a)N'_2(a,\theta_2)\right].
\end{align*}
Equations~(\ref{DIR_G}) and~(\ref{DIR_PSI}) hold in this case too.


\subsubsection{Case of Robin conditions}

\noindent
Robin conditions are defined as linear combinations of the form:
\begin{equation}
\lambda \, \psi(a) + \mu \, \psi'(a) = \xi_a, \qquad 
\gamma \, \psi(b) + \delta \, \psi'(b) = \xi_b.
\label{ROBC}
\end{equation}
The boundary match operator in this case is defined as:
\begin{equation*}
\mathcal{P}_x(\theta_1,\theta_2) \psi(x) \triangleq \left[ \lambda \, \psi(a) + \mu \, \psi'(a) \right] \, F(x,\theta_1) \, + \left[ \gamma \, \psi(b) + \delta \, \psi'(b) \right] \, H(x,\theta_2),
\end{equation*}
where:
\begin{equation*}
F(x,\theta_1) = N_1(x,\theta_1) + F_1(\theta_1)(2x-a-b) +F_2(\theta_1)(x-a)(x-b),
\end{equation*}
with:
\begin{align*}
F_1(\theta_1) ~=~ & \frac{\delta \left[ 1-\lambda N_1(a,\theta_1)-\mu N'_1(a,\theta_1) \right] -\mu \left[ \gamma N_1(b,\theta_1)+\delta N'_1(b,\theta_1)\right]}{\delta \left[ 2\mu -\lambda (b-a) \right] + \mu \left[ 2\delta+\gamma (b-a) \right]}, \\ 
 & \\
F_2(\theta_1) ~=~ & \frac{-\gamma N_1(b,\theta_1)-\delta N'_1(b,\theta_1)-\left[ 2\delta +\gamma (b-a)\right] F_1(\theta_1)}{\delta (b-a)},
\end{align*}
and:
\begin{equation*}
H(x,\theta_2) = N_2(x,\theta_2) + H_1(\theta_2)(2x-a-b) +H_2(\theta_2)(x-a)(x-b),
\end{equation*}
with:
\begin{align*}
H_1(\theta_2) ~=~ & \frac{\delta \left[ -\lambda N_2(a,\theta_2)-\mu N'_2(a,\theta_2)\right] + \mu \left[ 1-\gamma N_2(b,\theta_2)-\delta N'_2(b,\theta_2)\right]}{\delta \left[2\mu -\lambda (b-a)\right]+\mu \left[2\delta+\gamma (b-a)\right]}, \\
 & \\
H_2(\theta_2) ~=~ & \frac{1-\gamma N_2(b,\theta_2)-\delta N'_2(b,\theta_2)-\left[2\delta +\gamma (b-a)\right] H_1(\theta_2)}{\delta (b-a)},
\end{align*}
while Eqs.~(\ref{DIR_G}) and~(\ref{DIR_PSI}) provide the trial solution. 


\section{Reductive transformation of Neumann conditions to equivalent Dirichlet boundary conditions}
\label{REDUCT}

\noindent
Instead of maintaining different boundary match operators for all condition types, neural forms for Dirichlet conditions of the form $\psi(a) = \xi_a$, $\psi(b) = \xi_b$, can be solely considered. In this case, a mixed problem with conditions $\psi'(a) = \xi_a$ and $\psi(b) = \xi_b$ can be tackled by replacing $\psi(a)$ with a suitable expression, such that the Dirichlet-type neural form satisfies the mixed conditions.
Let us demonstrate this idea with a simple example, using the following non-parametric (rigid) boundary matching operator:
\begin{equation}
\mathcal{P}_x \psi(x) = \psi(a) \, \left( \frac{x-b}{b-a} \right)^2 + \psi(b) \, \left(\frac{x-a}{b-a}\right)^2.
\label{NPBO}
\end{equation} 
The trial solution is given as:
\begin{equation}
\Psi_T(x,\theta) = N(x,\theta) + \left[ \psi(a)-N(a,\theta) \right] \, \left( \frac{x-b}{b-a} \right)^2 + \left[ \psi(b)-N(b,\theta) \right] \, \left( \frac{x-a}{b-a} \right)^2, 
\label{PSITRIAL} 
\end{equation}
and its first derivative with respect to $x$ is:
\begin{equation*}
\Psi'_T(x,\theta) = N'(x,\theta)  + 2 \, \left[ \psi(a)-N(a,\theta) \right] \, \frac{x-b}{(b-a)^2} + 2 \, \left[ \psi(b)-N(b,\theta) \right] \, \frac{x-a}{(b-a)^2}.    
\end{equation*} 
Demanding that $\Psi'_T(a,\theta) = \psi'(a)$ and solving for $\psi(a)$ yields:
\begin{equation*}
\psi(a) = N(a,\theta) + \frac{b-a}{2} \, \left[ N'(a,\theta)-\psi'(a) \right].
\end{equation*}
Substituting $\psi(a)$ in Eq.~(\ref{PSITRIAL}) results in the neural form that satisfies the mixed boundary conditions. 
We provide the following result: 
\begin{align*}
\psi(a) ~=~ & \frac{(b-a) \, \left[ \xi_a - \mu \, N'(a,\theta) \right] - 2 \, \mu \, N(a,\theta)}{(b-a) \, \lambda - 2 \, \mu}, \\
 & \\
\psi(b) ~=~ & \frac{(b-a) \, \left[ \xi_b - \delta \, N'(b,\theta) \right] + 2 \, \gamma N(b,\theta)}{(b-a) \, \gamma + 2 \, \delta}.
\end{align*}
for the case of Robin conditions of Eq.~(\ref{ROBC}).


\section{Goodness of the approximate solution}
\label{UPPER_BOUND}

\noindent
Assessing the quality of an approximate solution is of fundamental importance. Relevant research outcomes have recently been reported in~\cite{Protopapas2023}. In the following paragraphs, we present a perturbation approach that can yield, under certain weak assumptions, a reliable upper bound for the absolute deviation $|\Psi_T(x)-\psi(x)|$ 
of the obtained solution from the exact one.
More specifically, let: 
\[
\Psi_T(x) = \mathcal{P}_x(\theta^*)\psi(x) + (1-\mathcal{P}_x(\theta^*)) \, N(x,w^*),
\]
be the trial solution obtained by an augmented neural form. The corresponding error is positive and may be written as $E(\Psi_T)=s^2$,
where $s$ is a scalar. We now construct a new trial solution by adding a perturbation $\eta(x,\gamma)$ to the obtained trial solution $\Psi_T(x)$. In order to preserve the satisfaction of the prescribed conditions, the perturbation should be of the form:
\[
\eta(x,\gamma) = (1-\mathcal{P}_x(\theta^*)) \, N_s(x,\gamma),
\]
with $N_s(x,\gamma)$ be a neural network. The perturbed trial solution, i.e.,  $\Psi_T(x) + \eta(x,\gamma)$, produces the error function $E(\Psi_T+\eta)$ that is minimized  with respect to the perturbation parameters $\gamma$, starting from an initial vector $\gamma^{(0)}$ that satisfies $\eta\left(x,\gamma^{(0)}\right)=0$. The minimized  error value, corresponding to an optimum vector $\gamma$, is  $E(\Psi_T+\eta) \triangleq \delta^2 < s^2 $.

Consider a second order ODE of the form $ \psi''(x)-f(x,\psi,\psi') = 0 $. The perturbed error value is then given as:
\begin{equation}
E(\Psi_T+\eta) = \int \left[\Psi''_T(x) + \eta''(x,\gamma) - f(x,\Psi_T+\eta,\Psi_T'+\eta')\right]^2 dx.
\label{ERROR_ETA}
\end{equation}
Assuming that  $\eta(x,\gamma)$ is a small perturbation, we may consider the first-order approximation: 
\begin{equation}\label{LINEAR}
f(x,\Psi_T+\eta,\Psi_T'+\eta') \approx f(x,\Psi_T,\Psi_T') + \eta \, \frac{\partial f}{\partial \Psi_T} + \eta' \, \frac{\partial f}{\partial \Psi_T'}.
\end{equation}
Substituting in Eq.~(\ref{ERROR_ETA}), we derive that:
\begin{equation}
\delta^2 = s^2 + 2 \, \int (\Psi''_T-f) \, \left( \eta''-\eta \, \frac{\partial f}{\partial \Psi_T} - \eta' \, \frac{\partial f}{\partial \Psi_T'} \right) dx.
\label{DELTA_ERROR}
\end{equation}
If there exists an $\eta_{ex}$ such that $E(\Psi_T+\eta_{ex})=0$, then:
\begin{equation}
s^2 + 2 \, \int(\Psi''_T-f) \, \left[ \eta_{ex}'' - \eta_{ex} \, \frac{\partial f}{\partial \Psi_T} - \eta_{ex}' \, \frac{\partial f}{\partial \Psi_T'} \right] dx = 0.
\label{ZERO_ERROR}
\end{equation}
Combining Eqs.~(\ref{DELTA_ERROR}) and~(\ref{ZERO_ERROR}) yields:
\begin{equation}
\left| \eta_{ex}(x) \right| = \frac{\left| \eta(x,\gamma^*) \right|}{1-\frac{\delta^2}{s^2}}
\end{equation}
This provides a reliable estimate for an upper bound of $|\eta_{ex}(x)|$
as long as the assumed linear approximation in Eq.~(\ref{LINEAR}) is valid. In turn, this implies that $ \eta(x,\gamma)$ must be a small perturbative correction.


\section{Experimental setting}
\label{sec:ExperSetting}

\noindent
The following section provides a comprehensive presentation of the experimental assessment procedure of the proposed augmented neural forms. More specifically, it includes detailed information on the selected test problems, the dataset generation schemes, the neural architecture, and the performance measures.

\subsection{Test problems}
\label{sec:TestProblems}

\noindent
The test problems included a number of first-order and second-order ODEs, as well as first-order ODE systems, accompanied by boundary and initial conditions of various types that are frequently encountered in real world problems. 

\begin{enumerate}[(1)]
\item \textit{Test Problem 1} (first-order ODE)\\
This is a stiff ODE \cite{HAIRER_1999} defined as:
\begin{align*}
\psi'(x) ~=~ & -50 \, \left( \psi(x)-\cos(x) \right), \quad x \in [0,1.5],\\
\psi(0) ~=~ & 0.15,
\label{STIFF_HAIRER_1999}
\end{align*}
with exact solution:
\begin{equation*}
\psi(x) = \left(0.15-\frac{2500}{2501}\right)\, \exp(-50 \, x) + \frac{50}{2501} \, \sin(x) + \frac{2500}{2501} \, \cos(x).
\end{equation*}
\item \textit{Test Problem 2} (second-order, nonlinear ODE)\\
This problem is defined as:
\begin{equation*}
\psi''(x) - \psi(x) \, \psi'(x) - \psi^3(x) = 0, \quad x \in [0,9.5],
\end{equation*}
with exact solution:
\[
\psi(x) = (10-x)^{-1}.
\]
In our experiments, it has been considered as a boundary value problem with Dirichlet, Neumann, mixed, and Robin conditions, and also as an initial value problem with Cauchy conditions. 

\begin{figure}[t]
\centering
\subfloat[Test Problem 1]{\includegraphics[scale=0.2]{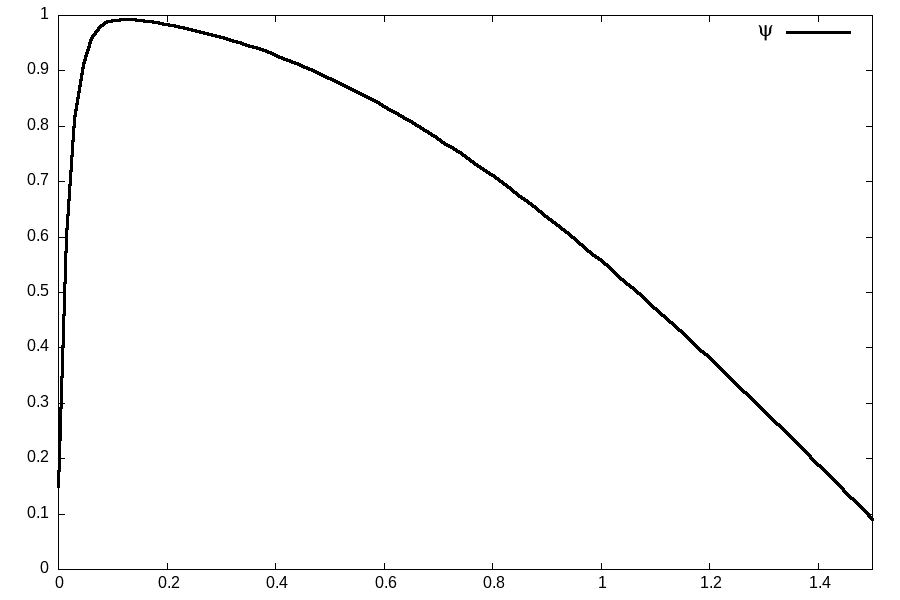}} 
\subfloat[Test Problem 2]{\includegraphics[scale=0.2]{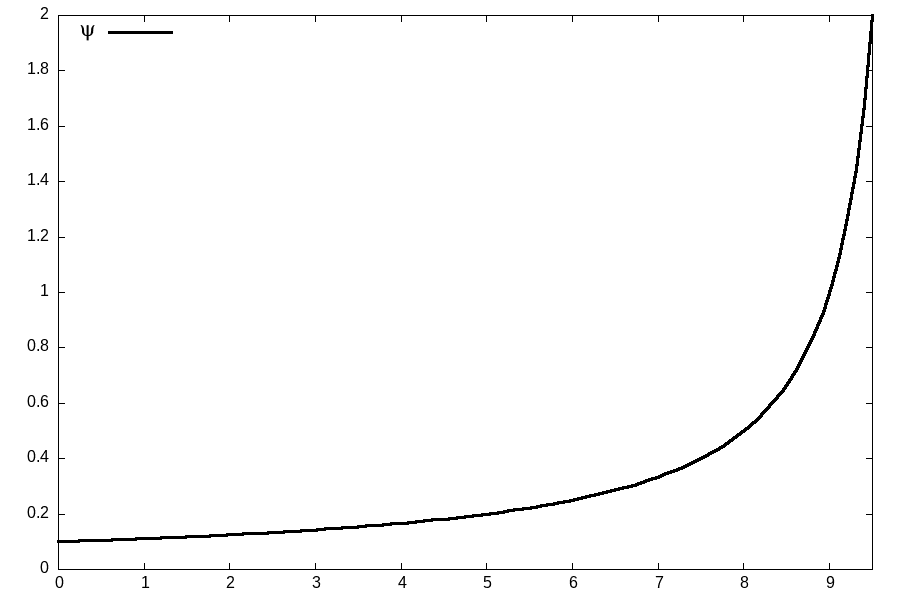}} 
\subfloat[Test Problem 3]{\includegraphics[scale=0.2]{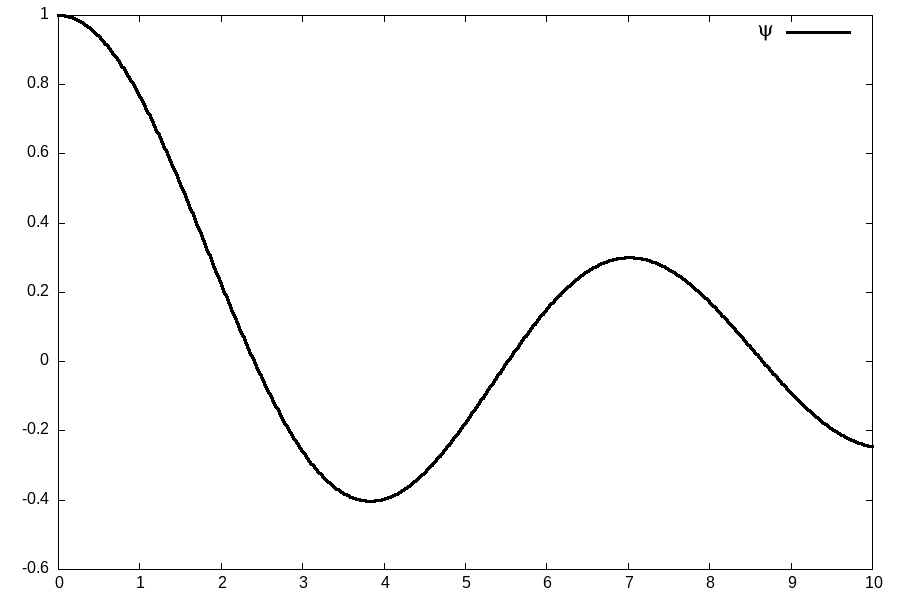}} \\
\subfloat[Test Problem 4]{\includegraphics[scale=0.2]{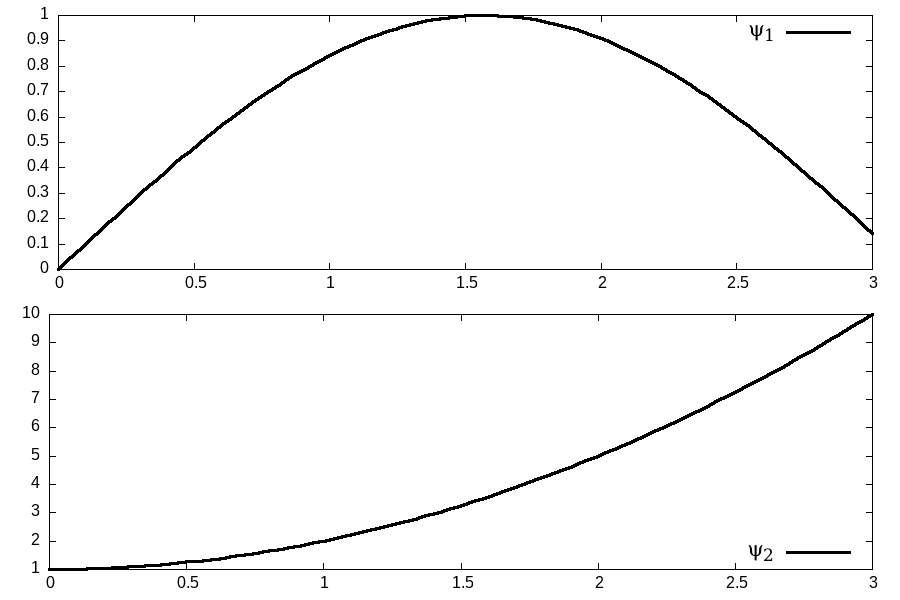}}
\subfloat[Test Problem 5]{\includegraphics[scale=0.2]{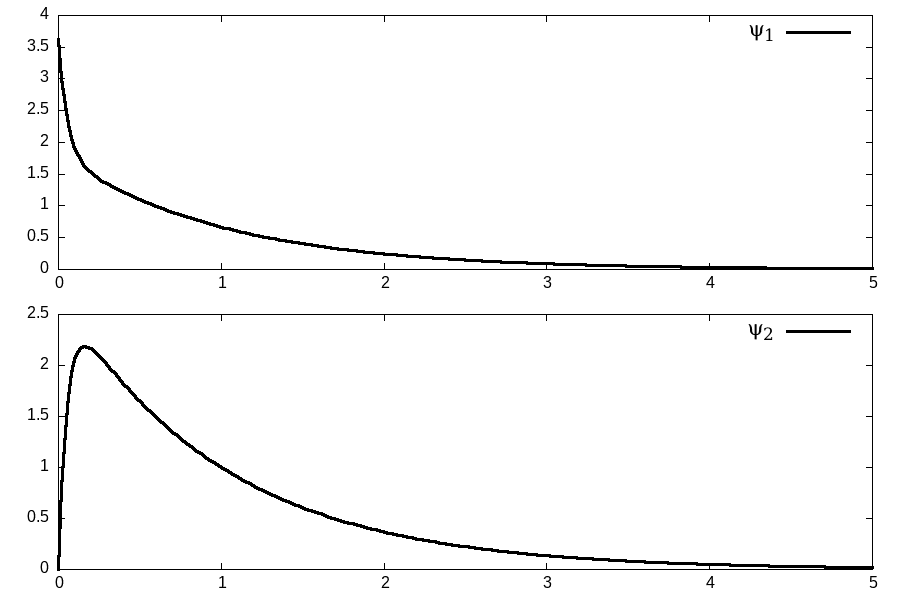}} 
\subfloat[Test Problem 6]{\includegraphics[scale=0.2]{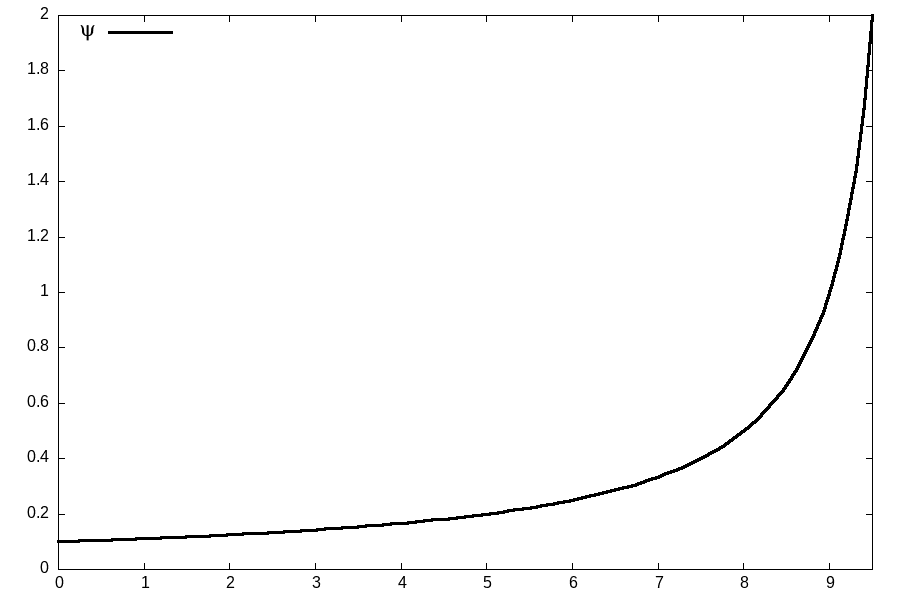}} \\
\caption{Plots of the exact solutions for all test problems.}
\label{fig:figs}
\end{figure}

\item \textit{Test Problem 3} (second-order ODE)\\
This is Bessel's equation subject to Cauchy initial conditions:
\begin{equation}
x^2 \, \psi''(x) + x \, \psi'(x) + (x^2-\nu^2) \, \psi(x) = 0, \quad x \in [0,10],
\label{BESSEL}
\end{equation}
\begin{equation*}
\psi(0)=1, \qquad  \psi'(0)=0,
\end{equation*}
The exact solution is Bessel's function of first kind $J_{\nu}(x)$, which is non-singular at the origin. Here, the case $\nu=0$ was considered. 
\item \textit{Test Problem 4} (first-order system)\\
This is a nonlinear system of two equations defined as~\cite{LAGARIS_1998}:
\begin{equation}
\begin{aligned}
\psi'_1(x) ~=~ & \cos(x) +\psi_1^2(x) + \psi_2(x) - \left[ 1+x^2+\sin^2(x) \right], \\
\psi'_2(x) ~=~ & 2 \, x - \left( 1+x^2 \right) \, \sin(x) + \psi_1(x) \, \psi_2(x),\\
 & \psi_1(0)=0, \qquad  \psi_2(0)=1, 
\end{aligned}
\qquad x \in [0,3],
\end{equation}
with exact solution:
\[
\psi_1(x) = \sin(x), \quad  \psi_2(x) = 1+ x^2.
\]
The trial solution is adapted for the fist-order system as:
\begin{align*}
\Psi_{T_1}(x,w_1,\theta_1) ~=~ & N_1(x,w_1) + \left[ 1 + \tilde{N}_1(x,\theta_1) - \tilde{N}_1(0,\theta_1) \right] \, \left[ \psi_1(0)-N_1(0,w_1) \right], \\
\Psi_{T_2}(x,w_2,\theta_2) ~=~ & N_2(x,w_2) + \left[ 1 + \tilde{N}_2(x,\theta_2)-\tilde{N}_2(0,\theta_2) \right] \, \left[ \psi_2(0)-N_2(0,w_2) \right],
\end{align*}
where $N_1$, $\tilde{N}_1$, and $N_2$, $\tilde{N}_2$, are the neural networks involved in the trial solutions $\Psi_{T_1}$ and $\Psi_{T_2}$, respectively.
\item \textit{Test Problem 5} (first-order system)\\
This is a stiff system defined as~\cite{HAIRER_1999}:
\begin{equation}
\begin{aligned}
\psi'_1(x) ~=~ & -10 \, \psi_1(x) + 6 \, \psi_2(x), \\
\psi'_2(x) ~=~ & 13.5 \, \psi_1(x) - 10 \, \psi_2(x),
\end{aligned}
\qquad x \in [0,5],
\end{equation}
\[
\psi_1(0) = \frac{4}{3} \, \exp(1.0), \qquad  \psi_2(0) = 0,
\]
with exact solution:
\begin{align*}
\psi_1(x) ~=~ & \frac{2}{3} \, \exp(1.0) \, \left( \exp(-x) + \exp(-19x) \right), \\
\psi_2(x) ~=~ & \exp(1.0) \left( \exp(-x) - \exp(-19x) \right).
\end{align*}
\item \textit{Test Problem 6} (reduction of second-order ODE to first-order system)\\
This is a nonlinear system derived by reducing the second-order ODE of Test Problem 2 as follows:
\begin{equation}
\begin{aligned} 
\psi'_1(x) ~=~ & \psi_2(x), \\
\psi'_2(x) ~=~ & \psi_1(x) \, \psi_2(x) + \psi_1^3(x),
\end{aligned}
\qquad x \in [0,9.5],
\label{SYSTEM_MAIN_NL}
\end{equation}
This problem was considered under various types of boundary conditions, namely Dirichlet, Neumann, mixed Dirichlet-Neumann, as well as with Cauchy initial conditions.

\end{enumerate}

\noindent
The exact solutions for all test problems are depicted in Fig.~\ref{fig:figs}.


\subsection{Dataset generation}
\label{sec:Dataset}

\noindent
For each test problem, two sets of data points were generated, namely a \textit{training dataset} of size $M_{\rm tr}$ that was used to train the neural forms, and a \textit{test dataset} of size $M_{\rm ts}$ that was used to assess the generalization quality. For the training dataset generation, two grid types were considered with data points $x_i \in [a,b]$:
\begin{enumerate}[(1)]
\item \textit{Equidistant points}:
\[
x_i = a + (i-1) \, \frac{b-a}{M_{\rm tr}-1}, \qquad i=1,2,\ldots,M_{\rm tr}.
\]
\item \textit{Chebyshev points}:
\[
x_i = \frac{a+b}{2} + \frac{b-a}{2} \, \cos \left( \frac{(2 \, i-1) \, \pi}{2 \, M_{\rm tr}} \right), \qquad i=1,2,\ldots,M_{\rm tr}.
\]
\end{enumerate}
The two grid types offer different point distributions inside the interval of interest. Chebyshev grids get denser towards the end points and are known to suppress the Runge phenomenon. Gauss-Legendre grids were tested as well, offering almost identical results to those of the Chebyshev grid. Hence, they were omitted from our study. 

\begin{table}[t] 
\centering
\small
\begin{tabular}{cccccccc}
\hline
Test & \multicolumn{3}{c}{Training points ($M_{\rm tr}$)} & Test points ($M_{\rm ts}$) & \multicolumn{3}{c}{Neural parameters ($|\theta|$)} \\
Problem & low & medium & high &  & low & medium & high \\
\hline
1 &  40 &  80 & 160 & 1000 &  36 &  72 & 144 \\
2 &  90 & 180 & 270 & 1000 &  90 & 180 & 270 \\
3 &  90 & 180 & 270 & 1000 &  90 & 180 & 270 \\
4 &  70 & 130 & 250 & 1000 &  60 & 120 & 240 \\
5 &  70 & 130 & 250 & 1000 &  60 & 120 & 240 \\
6 & 180 & 270 & 360 & 1000 & 180 & 270 & 360 \\
\hline
\end{tabular}
\caption{Number of training points ($M_{\rm tr}$), test points ($M_{\rm ts}$), and neural form parameters ($|\theta|$) per test problem.}
\label{tab:datapoints}
\end{table}

The size of the training set is always problem dependent and, thus, regularly determined through preliminary trial-and-error experimentation. Naturally, complex problems require higher number of training points.  Table~\ref{tab:datapoints} reports the corresponding numbers in our experiments. For each test problem, three levels (low, medium, and high) were considered for $M_{\rm tr}$. On the other hand, the test set consisted of 
$M_{\rm ts} = 1000$ equidistant points in all cases.


\subsection{Neural network architecture}
\label{sec:NetParams}

\noindent
In all test problems, a feedforward  neural network with one hidden layer and sigmoid nodes was employed. Such a network admits a scalar $x \in \mathbb{R}$ as input, and produces a scalar output as follows: 
\begin{equation}
N(x,\theta) = \sum_{i=1}^{K} a_{i} \, \sigma(w_{i} x+b_{i}), 
\label{eq:SigmoidalNN}
\end{equation}
where $K$ is the number of neurons, 
$\displaystyle \sigma(z) = [1+\exp(-z)]^{-1}$\  is the sigmoid activation function,
and $\theta$ is the network's parameter vector of size $|\theta|$. 
For a network of $K$ neurons, we have $|\theta| = 3 K$ parameters:
\begin{equation*}
\theta = \Big( \underbrace{a_{1},w_{1},b_{1}}_{neuron~1}, \underbrace{a_{2},w_{2},b_{2}}_{neuron~2},\ldots,\underbrace{a_{K},w_{K},b_{K}}_{neuron~K} \Big)
\end{equation*}
Our implementation was coded in \texttt{Fortran}, while the training was performed by the \texttt{Merlin} optimization platform~\cite{MERLIN_1998, MERLIN_2004}, which offers a variety of robust and efficient minimization routines. The minimization algorithms that were used in our experiments were (\textit{i}) a pattern search method based on alternating variables, (\textit{ii}) the irregular simplex method of Nelder and Mead~\cite{NELDER_MEAD_1965}, (\textit{iii}) Powell's version of BFGS (TOLMIN)~\cite{POWELL_1989} with Goldfarb-Idnani factors and weak Wolfe conditions for the line search, and (\textit{iv}) a trust-region version of BFGS based on Powell's dogleg technique \cite{Powell_1970}. All experiments were performed on the PRECIOUS high performance computing infrastructure \cite{precious}, which consists of 12 computing nodes, each of which utilizes 2 Intel(R) Xeon(R) Gold 5220R processors with 48 computing cores. 
The number of neural form parameters used in our experiments are  reported also in Table~\ref{tab:datapoints}.


\subsection{Solution quality criteria}
\label{sec:SolutionQualityCriteria}

\noindent
An ideal criterion for the quality of an approximate solution is its difference from the exact solution. Given that, in previously unmet problems, the exact solution is not available, an alternative quality criterion can be the size of the ODE residuals. However, this measure may not exhibit the same behavior as the solutions' difference. In our experiments, all the selected test problems had known exact solutions, hence allowing accurate comparisons and evaluation of our approximate solutions.


Let $M$ denote the number of data points over which the performance is evaluated (i.e., $M=M_{\rm tr}$ for training, and $M=M_{\rm ts}$ for testing), $\psi(x_i)$ be the exact solution at $x_i$, and $\Psi_T(x_i,\Theta)$ be the trial solution at $x_i$, where $\theta$ refers collectively to the neural form parameters. Then, the employed performance measures were as follows:
\begin{enumerate}[(1)]
\item \textit{Measures based on the residuals}
\begin{align}
\text{MSE} = \frac{1}{M} \, \sum_{i=1}^{M} \, \left[ \mathcal{L}_x\Psi_T(x_i,\Theta)-f(x_i) \right]^2 & 
\qquad \text{(Mean squared error)}
\label{eq:MSE}\\
\text{MXE} = \max_{i} \, \left\{ \left[ \mathcal{L}_x\Psi_T(x_i,\Theta)-f(x_i) \right]^2 \right\} & 
\qquad \text{(Maximum squared error)}
\label{eq:MXE}
\end{align}
\item \textit{Measures based on estimated deviation}
\begin{align}
\textrm{MSED} = \frac{1}{M} \, \sum_{i=1}^{M} \, \left[ \eta_{ex}(x_i) \right]^2 & 
\qquad \text{(Mean squared estimated deviation)}
\label{eq:MSED} \\
\textrm{MXED} = \max_{i} \, \left\{ \left[ \eta_{ex}(x_i) \right]^2 \right\} & 
\qquad \text{(Maximum squared estimated deviation)}
\label{eq:MXED}
\end{align}
\item \textit{Measures based on exact deviation}
\begin{align}
\textrm{MSD} = \frac{1}{M} \, \sum_{i=1}^{M} \, \left[ \Psi_T(x_i,\Theta) - \psi(x_i) \right]^2 & 
\qquad \text{(Mean squared deviation)}
\label{eq:MSD} \\
\textrm{MXD} = \max_{i} \, \left\{ \left[ \Psi_T(x_i,\Theta)-\psi(x_i) \right]^2 \right\} & 
\qquad \text{(Maximum squared deviation)}
\label{eq:MXD}
\end{align}
\end{enumerate}
Note that measures based on ODE residuals indicate precision over the ODE's satisfaction, while measures based on the estimated and exact deviation indicate the distance from the exact solution $\psi(x)$.

\begin{table}[h!] 
\centering
\footnotesize
\begin{tabular}{cccccccccccc}  
\hline
 & & & & & & & \multicolumn{2}{c}{Training dataset} & & \multicolumn{2}{c}{Test dataset \hskip 40pt \,} \\
\cline{8-9} \cline{11-12}
Problem & Conditions & Method & $|\theta|$ & $M_{\rm tr}$ & Grid &  & MSD & MXD & & MSD & MXD \\ 
\hline
1 & D & Neural form & 144 & 160 & Ch & & 0.22e-17 & 0.36e-16 & & 0.25e-18 & 0.36e-16 \\
  & & ode113 & & 300 & Ad & & 0.58e-28 & 0.18e-26 & & 0.12e-23 & 0.99e-21 \\
  & & Neural network & 36 & 80 & Ch & & 0.16e-17 & 0.35e-16 & & 0.25e-18 & 0.36e-16 \\    
\hline
2 & D-D & Neural form & 90 & 270 & Ch & & 0.83e-23 & 0.98e-22 & & 0.12e-22 & 0.98e-22 \\
  & & Neural network & 90 & 180 & Ch & & 0.90e-23 & 0.68e-22 & & 0.13e-22 & 0.69e-22 \\
\cline{2-12}
& D-N & Neural form & 90 & 90 & Ch & & 0.17e-21 & 0.14e-20 & & 0.24e-21 & 0.14e-20 \\
  & & Neural network & 90 & 180 & Ch & & 0.37e-22 & 0.33e-21 & & 0.49e-22 & 0.33e-21 \\  
\cline{2-12}
& N-N & Neural form & 90 & 270 & Ch & & 0.61e-23 & 0.25e-22 & & 0.65e-23 & 0.26e-22 \\
  & & Neural network & 180 & 180 & Ch & & 0.19e-22 & 0.15e-21 & & 0.27e-22 & 0.15e-21 \\
\cline{2-12}  
& C & Neural form & 90 & 180 & Ch & & 0.43e-22 & 0.58e-21 & & 0.13e-22 & 0.58e-21 \\
  & & Neural network & 90 & 270 & Ch & & 0.75e-20 & 0.10e-18 & & 0.19e-20 & 0.10e-18 \\
\cline{2-12}   
& R & Neural form & 180 & 270 & Ch & & 0.40e-23 & 0.37-22 & & 0.59e-23 & 0.37e-22 \\
  & & Neural network & 180 & 270 & Ch & & 0.20e-22 & 0.17e-21 & & 0.29e-22 & 0.17e-21 \\
\hline
3 & C & Neural form & 180 & 180 & Eq & & 0.45e-18 & 0.21e-17 & & 0.45e-18 & 0.21e-17 \\
  & & ode113 & & 215 & Ad & & 0.31e-29 & 0.28e-28 & & 0.12e-16 & 0.73e-14 \\
  & & Neural network & 180 & 270 & Eq & & 0.38e-18 & 0.18e-17 & & 0.38e-18 & 0.18e-17 \\
\hline
4 & D & Neural form & 240 & 250 & Ch & $\psi_1$ & 0.15e-18 & 0.12e-17 & & 0.61e-19 & 0.12e-17 \\
  & & & & & & $\psi_2$ & 0.12e-17 & 0.95e-17 & & 0.43e-18 & 0.95e-17 \\
  & & ode113 & & 109 & Ad & $\psi_1$ & 0.20e-24 & 0.51e-23 & & 0.91e-24 & 0.63e-22 \\
  & & & & &  & $\psi_2$ & 0.14e-23 & 0.41e-22 & & 0.20e-23 & 0.75e-22 \\
  & & Neural network & 240 & 250 & Ch & $\psi_1$ & 0.91e-07 & 0.69e-06 & & 0.36e-07 & 0.70e-06 \\
  & & & & & & $\psi_2$ & 0.68e-06 & 0.57e-05 & & 0.26e-06 & 0.57e-05 \\  
\hline
5 & D & Neural form & 240 & 250 & Ch & $\psi_1$ & 0.30e-21 & 0.17e-20 & & 0.42e-21 & 0.17e-20 \\
  & & & & & & $\psi_2$ & 0.29e-21 & 0.18e-20 & & 0.36e-21 & 0.18e-20 \\
  & & ode113 & & 321 & Ad & $\psi_1$ & 0.52e-29 & 0.86e-28 & & 0.72e-24 & 0.37e-21  \\
  & & & & &  & $\psi_2$ & 0.16e-28 & 0.31e-27 & & 0.16e-23 & 0.84e-21 \\
  & & Neural network & 240 & 250 & Ch & $\psi_1$ & 0.13e-21 & 0.95e-21 & & 0.17e-21 & 0.95e-21 \\
  & & & & & & $\psi_2$ & 0.48e-21 & 0.30e-20 & & 0.67e-21 & 0.30e-20 \\  
\hline
6 & D-D & Neural form & 270 & 360 & Ch & & 0.20e-22 & 0.10e-21 & & 0.27e-22 & 0.10e-21 \\
  & & ode113 & & 203 & Ad & & 0.13e-26 & 0.56e-26 & & 0.63e-22 & 0.52e-19 \\
  & & Neural network & 180 & 180 & Ch & & 0.30e-22 & 0.21e-21 & & 0.41e-22 & 0.21e-21 \\  
\cline{2-12}  
& D-N & Neural form & 360 & 360 & Ch & & 0.13e-20 & 0.95e-20 & & 0.18e-20 & 0.96e-20 \\
  & & ode113 & & & Ad & & 0.11e-26 & 0.46e-26 & & 0.63e-22 & 0.52e-19 \\
  & & Neural network & 360 & 360 & Ch & &  0.20e-21 & 0.13e-20 & & 0.29e-21 & 0.13e-20 \\  
\cline{2-12}  
& N-N & Neural form & 360 & 360 & Ch & &  0.63e-23 & 0.56e-22 & & 0.82e-23 & 0.57e-22 \\
  & & ode113 & & 203 & Ad & & 0.42e-27 & 0.14e-26 & & 0.63e-22 & 0.52e-19 \\
  & & Neural network & 270 & 270 & Ch & &  0.13e-21 & 0.94e-21 & & 0.17e-21 & 0.94e-21 \\  
\cline{2-12}  
& C & Neural form & 270 & 360 & Ch & & 0.24e-19 & 0.34e-18 & & 0.62e-20 & 0.34e-18 \\
  & & ode113 & & 203 & Ad & & 0.27e-24 & 0.42e-23 & & 0.64e-22 & 0.53e-19 \\
  & & Neural network & 360 & 360 & Ch & &  0.72e-19 & 0.99e-18 & & 0.18e-19 & 0.99e-18 \\  
\hline
\multicolumn{11}{l}{\scriptsize Abbreviations: D=Dirichlet, N=Neumann, D-N=Mixed Dirichlet-Neumann, R=Robin, C=Cauchy, Ch=Chebyshev, Eq=Equidistant, Ad=Adaptive}
\end{tabular}
\caption{Mean and maximum squared deviation for the best neural form, the best ode113 results and the best neural network per test problem and condition type. The corresponding grid type, number of neural form parameters $|\theta|$, and number of training points $M_{\rm tr}$ are also reported.}
\label{tab:compactRes}
\end{table}

\section{Experimental results}
\label{sec:results}

\noindent
Our experimental setting was designed to assess the performance of the proposed methodology across a diverse range of problems and computational scenarios that are commonly encountered in practice. More specifically, both equidistant grids and Chebyshev grids were used to generate training points. Three levels (low, medium, and high) were considered for the grid-point density as well as for the network size (total number of parameters) as reported in Table~\ref{tab:datapoints}. In order to prevent overtraining and its subsequent degradation in generalization quality, scenarios where the number of network parameters exceeded the number of grid points were excluded. Thus, a total number of 12 distinct scenarios were ultimately considered. 

For each training scenario and test problem, 100 independent experiments were conducted and the obtained solutions were assessed according to the quality measures presented in Section~\ref{sec:SolutionQualityCriteria}. A computational budget of 220000 function evaluations proved to be sufficient for training in all test problems. Note that Test Problem~2 was solved also using the reductive transformation of Section~\ref{REDUCT}, without any significant difference. The results of the proposed neural forms were compared with the corresponding results obtained by:
\begin{enumerate}[(i)]
\item A baseline neural method where the trial solution is a standalone neural network and penalty terms are added to the error function, in order to account for the prescribed conditions according to the form of Eq.~(\ref{eq:baseline_error}) with $\zeta =1$.
\item The \texttt{ode113} numerical solver of \textsc{Matlab}\textsuperscript{\textregistered}~\cite{matlabodesuite}, which is a variable-step, variable-order  Adams-Bashforth-Moulton PECE solver of orders 1 to 13.  In this case, the boundary value problems were handled via shooting, while the performance over the test datasets was computed using spline interpolation. 
\end{enumerate}
Given the large volume of results, Table~\ref{tab:compactRes} presents a condensed selection that includes the mean and maximum squared deviations from the exact solutions for the best neural form architecture, the \texttt{ode113} solver with quintic splines interpolation, and the top-performing standalone neural network. The detailed results for each individual test problem are reported in Table~\ref{tab:ode113_results} of~\ref{appendix:ode113} for the \texttt{ode113} solver, in Tables~\ref{tab:Res_Prob_1_SNN}-\ref{tab:Res_Prob_6_NN_SNN} of~\ref{appendix:NN} for the baseline neural method (standalone neural network), and in Tables~\ref{tab:Res_Prob_1}-\ref{tab:Res_Prob_6_NN} of~\ref{appendix:NF} for the augmented neural forms.

The reported results clearly demonstrate that the proposed augmented neural forms offered high-quality solutions for all test problems. As expected, the \texttt{ode113} solver achieved excellent precision at the points of its adaptive grid. However, on the test points where interpolation is necessary, cubic and 7-th order splines yielded inferior results than neural forms. Only quintic splines demonstrated superior performance, as can be verified by inspecting Table~\ref{tab:ode113_results}. It shall be noted that the optimal interpolation choice for this type of numerical methods is not \textit{a priori} known, hence requiring a trial-and-error experimentation by the user. Also, it shall be underlined that, in contrast to numerical methods, neural forms are interpolants themselves and, thus, they can be differentiated, integrated, or otherwise processed in a straightforward manner. This is an important feature that the traditional numerical solvers lack, thereby emphasizing the complementarity of neural forms to the available numerical solvers.

\begin{table}[h!] 
\centering
\small
\resizebox{1.\columnwidth}{!}{%
\begin{tabular}{ccccccccccccc}  
\hline
Test & Condition & Neural & \multicolumn{3}{c}{Training Parameters} & &  \multicolumn{5}{c}{Mean Squared Deviation (MSD)} & Wilcoxon \\
\cline{4-6} \cline{8-12}
Problem & Type & Method & $|\theta|$ & $M_{\rm tr}$ & Grid &  & min & max & mean & median & std & $p$-value \\ 
\hline
1 & D & Neural form & 144 & 160 & Ch & & 0.25e-18 & 0.30e-15 & 0.53e-17 & 0.49e-18 & 0.31e-16 & \multirow{ 2}{*}{0.30e-05}\\
& & Neural network  & 36 & 80 & Ch & & 0.25e-18 & 0.38e-10 & 0.47e-12 & 0.23e-17 & 0.39e-11 &\\  
\hline
2 & D-D & {Neural form } & 90 & 270 & Ch & & 0.12e-22 & 0.30e-13 & 0.46e-15 & 0.13e-17 & 0.31e-14 &\multirow{ 2}{*}{0.23e-05}\\
  & & Neural network  & 90 & 180 & Ch & & 0.13e-22 & 0.12e-15 & 0.37e-17 & 0.38e-19 & 0.16e-16 &\\  
\cline{2-13}
& D-N & {Neural form } & 90 & 90 & Ch & & 0.24e-21 & 0.16e-11 & 0.28e-13 & 0.18e-17 & 0.17e-12 &\multirow{ 2}{*}{0.19e-13}\\
  & & Neural network  & 90 & 180 & Ch & & 0.49e-22 & 0.37e-16 & 0.10e-17 & 0.60e-19 & 0.50e-17 &\\  
\cline{2-13}
& N-N & {Neural form } & 90 & 270 & Ch & & 0.65e-23 & 0.10e-11 & 0.18e-13 & 0.45e-19 & 0.13e-12 & \multirow{ 2}{*}{0.20e-03}\\
  & & Neural network  & 180 & 180 & Ch & & 0.27e-22 & 0.72e-17 & 0.16e-18 & 0.10e-19 & 0.75e-18 &\\
\cline{2-13}  
& C & {Neural form } & 90 & 180 & Ch & & 0.13e-22 & 0.70e-09 & 0.79e-11 & 0.58e-16 & 0.70e-10 &\multirow{ 2}{*}{0.76e-09}\\
  & & Neural network  & 90 & 270 & Ch & & 0.29e-22 & 0.22e-12 & 0.46e-14 & 0.12e-17 & 0.30e-13 &\\
\cline{2-13}   
& R & Neural form & 180 & 270 & Ch & & 0.45e-23 & 0.41e-19 & 0.13e-20 & 0.24e-21  & 0.47e-20  &\multirow{ 2}{*}{0.31e-17}\\
  & & Neural network  & 180 & 270 & Ch & & 0.29e-22 & 0.53e-17 & 0.11e-18 & 0.28e-20 & 0.70e-18 &\\
\hline
3 & C & Neural form & 180 & 180 & Eq & & 0.21e-19 & 0.17e-06 & 0.45e-08 & 0.28e-15 & 0.24e-07 &\multirow{ 2}{*}{0.12e-09}\\
  & & Neural network  & 180 & 270 & Eq & & 0.24e-19 & 0.24e-09 & 0.27e-11 & 0.19e-16 & 0.24e-10 &\\
\hline
4 & D & Neural form & 240 & 250 & Ch & $\psi_1$ & 0.16e-20 & 0.30e-13 & 0.65e-14 & 0.69e-17 & 0.40e-13 &\multirow{ 2}{*}{0.25e-33} \\
& & Neural network  & 240 & 250 & Ch & $\psi_1$ & 0.85e-09 & 0.34e-06 & 0.86e-07 & 0.64e-07 & 0.68e-07 &\\
  & & Neural form & 240 & 250 & Ch & $\psi_2$ & 0.11e-19 & 0.21e-11 & 0.46e-13 & 0.48e-16 & 0.28e-12 &\multirow{ 2}{*}{0.25e-33}\\
  & & Neural network & 240 & 250 & Ch & $\psi_2$ & 0.60e-08 & 0.24e-05 & 0.60e-06 & 0.45e-06 & 0.47e-06 &\\  
\hline
5 & D & Neural form  & 240 & 250 & Ch & $\psi_1$ & 0.42e-21 & 0.39e-17 & 0.13e-18 & 0.13e-19 & 0.46e-18 & \multirow{ 2}{*}{0.39e-12}\\
 & & Neural network  & 240 & 250 & Ch & $\psi_1$ & 0.17e-21 & 0.26e-17 & 0.58e-19 & 0.19e-20 & 0.35e-18 &\\
 & & Neural form  & 240 & 250 & Ch & $\psi_2$ & 0.36e-21 & 0.57e-17 & 0.24e-19 & 0.28-19 & 0.76e-18 & \multirow{ 2}{*}{0.91e-15}\\ 
 & &  Neural network  & 240 & 250 & Ch &$\psi_2$ & 0.67e-21 & 0.29e-12 & 0.19e-14 & 0.24e-20 & 0.29e-13 &\\  
\hline
6 & D-D & {Neural form } & 270 & 360 & Ch & & 0.27e-22 & 0.19e-13 & 0.31e-15 & 0.38e-19 & 0.21e-14 &\multirow{ 2}{*}{0.34e+00}\\
 & & Neural network  & 180 & 180 & Ch & & 0.41e-22 & 0.27e-05 & 0.27e-07 & 0.25e-19 & 0.27e-06 &\\
\cline{2-13}  
& D-N & {Neural form } & 360 & 360 & Ch & & 0.18e-20 & 0.34e-15 & 0.71e-17 & 0.66e-18 & 0.35e-16 &\multirow{ 2}{*}{0.44e-23}\\
 & & Neural network  & 360 & 360 & Ch & &  0.21e-21 & 0.62e-17 & 0.13e-18 & 0.67e-20 & 0.74e-18 &\\
\cline{2-13}  
& N-N &{Neural form } & 360 & 360 & Ch & & 0.82e-23 & 0.58e-06 & 0.58e-08 & 0.16e-18 & 0.58e-07 &\multirow{ 2}{*}{0.92e-04}\\
 & & Neural network  & 270 & 270 & Ch & &  0.17e-21 & 0.73e-17 & 0.23e-18 & 0.31e-19 & 0.93e-18 &\\
\cline{2-13}  
& C & {Neural form } & 270 & 360 & Ch & & 0.62e-20 & 0.88e-10 & 0.21e-11 & 0.34e-14 & 0.11e-10 &\multirow{ 2}{*}{0.26e-25}\\
 & & Neural network  & 360 & 360 & Ch & &  0.71e-20 & 0.95e-16 & 0.64e-17 & 0.14e-17 & 0.14e-16 &\\
\hline
\multicolumn{11}{l}{\scriptsize Abbreviations: D=Dirichlet, N=Neumann, D-N=Mixed Dirichlet-Neumann, R=Robin, C=Cauchy, Ch=Chebyshev, Eq=Equidistant}
\end{tabular}}
\caption{Descriptive statistics of the mean squared deviations over the test datasets for the best neural form and the best neural network over 100 experiments per test problem and condition type. The corresponding grid type, number of neural form parameters $|\theta|$, and number of training points $M_{\rm tr}$ are also reported. The provided $p$-values refer to Wilcoxon rank-sum tests between the two approaches.}
\label{tab:compactResStats}
\end{table}

\begin{table}[h!] 
\centering
\small
\begin{tabular}{cccccccc}  
\hline
Metric   & Method & min & max & mean & median & std & $p$-value \\ 
\hline
Min & Neural form &   0.45e-23 & 0.25e-18 & 0.20e-19 & 0.36e-21 & 0.62e-19 & \multirow{ 2}{*}{0.96e-01}\\
MSD &  Neural network  &   0.13e-22 & 0.60e-08 & 0.46e-09 & 0.17e-21 & 0.15e-08 &\\  
\hline
Max & {Neural form } &   0.39e-17 & 0.58e-06 & 0.50e-07 & 0.10e-11 & 0.15e-06 &\multirow{ 2}{*}{0.68e+00}\\
MSD  &  Neural network  &   0.26e-17 & 0.27e-05 & 0.36e-06 & 0.12e-15 & 0.86e-06 &\\  
\hline
Mean  & Neural form &   0.13e-21 & 0.58e-08 & 0.69e-09 & 0.65e-14 & 0.18e-08 & \multirow{ 2}{*}{0.89e+00}\\
MSD &  Neural network  &   0.58e-19 & 0.46e-06 & 0.38e-07 & 0.64e-17 & 0.11e-06 &\\
\hline
Median  & Neural form &   0.24-21 & 0.34e-14 & 0.25e-15 & 0.66e-18 & 0.84e-15 &\multirow{ 2}{*}{0.28e+00} \\
MSD &  Neural network  &   0.19e-20 & 0.45e-06 & 0.34e-07 & 0.38e-19 & 0.11e-06 &\\
\hline
St.D.  & {Neural form } &  0.47e-20 & 0.58e-07 & 0.55e-08 & 0.40e-13 & 0.15e-07 & \multirow{ 2}{*}{0.89e+00}\\
MSD  & Neural network  & 0.35e-18 & 0.47e-06 & 0.54e-07 & 0.16e-16 & 0.13e-06 &\\
\hline
\end{tabular}
\caption{Statistical measures calculated over the MSD columns of Table~\ref{tab:compactResStats} for neural forms and the baseline neural method. The reported $p$-values refer to Wilcoxon paired (signed-rank) tests for the two approaches.}
\label{tab:compactResStatsPaired}
\end{table}

A comparison of the augmented neural forms against the baseline neural method does not yield a conclusive indication of superiority for either, as can be readily determined by inspecting Table~\ref{tab:compactRes}. 
More specifically, the reported MSD values over the test data suggest that each method achieves better performance for approximately half of the test problems. 
Further evidence is reported in Table~\ref{tab:compactResStats}, which reports the basic descriptive statistics, namely the minimum, maximum, mean, median, and standard deviation of the MSD values over 100 independent experiments, conducted for each test problem and for both the neural forms and the baseline neural method. 
The last column reports $p$-values of Wilcoxon rank-sum tests performed between the MSD samples of the compared approaches for each test problem. The small $p$-values obtained in all but one case, indicate statistically significant differences between the compared methods for significance levels 0.05 and 0.01. 
This suggests that, although a dominant method may exist for each distinct test problem, the count over all test problems is almost equally distributed between the two methods.

Further expanding our analysis, we conducted paired (signed-rank) Wilcoxon tests for each one of reported MSD descriptive statistics in Table~\ref{tab:compactResStats}. More specifically, each MSD column of Table~\ref{tab:compactResStats} provides the paired samples of the two neural methods regarding the corresponding statistic. For example, the ``min'' column offers a sample of 15 values for the neural form, and 15 paired values for the neural network. For all these samples, the basic statistical measures were calculated and they are reported in Table~\ref{tab:compactResStatsPaired}. For example, the maximum among the minimum MSD values achieved by the neural form for all test problems was 0.25e-18, while for the basic neural method it was 0.60e-08. 
The corresponding paired samples of the two neural methods for each statistic were compared using the Wilcoxon paired (signed-rank) test, and the obtained $p$-values are reported in the last column of  Table~\ref{tab:compactResStatsPaired}. For example, consider the paired samples of the minimum MSD, where each pair consists of the minimum MSD obtained from neural forms and the corresponding minimum MSD of the neural network. Since the calculated $p$-value is 0.96e-1, there is no statistically significant difference in significance level 0.05 or 0.01 between the two methods. 

The $p$-values reported in Table~\ref{tab:compactResStatsPaired} for all metrics suggest the lack of significant differences. Therefore, although it appears that for each specific problem it is highly probable to have a winner method, there is no statistically verified dominance of one method over the other across all the test problems. Thus, the proposed neural forms provide solutions of comparable quality to the baseline neural method, while also maintaining the critical advantage of exact matching the initial/boundary conditions. Note that Chebyshev training grids were shown to successfully treat the Runge phenomenon in many cases, thereby enhancing the performance of neural forms with respect to the solution deviation metrics.

Finally, it is worth noting that widely used solvers such as the stochastic gradient descent, Adam~\cite{ADAM_2014}, and variations~\cite{RADAM_2019}, which have recently gained increasing popularity for neural network training in big data applications, may not achieve good performance. Indeed, Table~\ref{tab:compactResAdam} in~\ref{appendix:AdamOptimizer} reports the mean and maximum squared deviation from the exact solutions for the best neural form trained using \texttt{Merlin} as described in Section~\ref{sec:NetParams}, and the same neural form trained with Adam. Comparing the reported deviations, it becomes evident that for all test problems Adam either converged to solutions of remarkably lower quality or completely failed to obtain a valid solution.


\begin{table}[t]
\centering
\begin{tabular}{clcccc} \hline
 & & \multicolumn{4}{c}{Deviation metrics}\\
\cline{3-6}
Test Problem & Condition Type & MSD & MXD & MSED & MXED  \\
\hline
1 & Dirichlet         & 0.19e-16 & 0.30e-16 & 0.34e-13 & 0.63e-13 \\ 
\hline
2 & Dirichlet         & 0.11e-20 & 0.24e-19 & 0.62e-15 & 0.26e-14 \\ 
 & Dirichlet-Neumann  & 0.68e-20 & 0.29e-18 & 0.20e-14 & 0.36e-14 \\ 
 & Neumann            & 0.86e-19 & 0.36e-17 & 0.14e-11 & 0.81e-11 \\ 
 & Cauchy             & 0.32e-19 & 0.18e-17 & 0.16e-16 & 0.43e-16 \\ 
 & Robin              & 0.55e-23 & 0.28e-22 & 0.49e-18 & 0.15e-17 \\ 
\hline
\end{tabular}
\caption{Deviation metrics for Test Problem~1 and Test Problem~2 with diverse condition types.}
 \label{MAIN_NL_ERROR_ESTIMATION}
\end{table} 

\subsection{Deviation bounding}
\label{subsec:dev_bound}

\begin{figure}[t]
\centering
\includegraphics[scale=0.38]{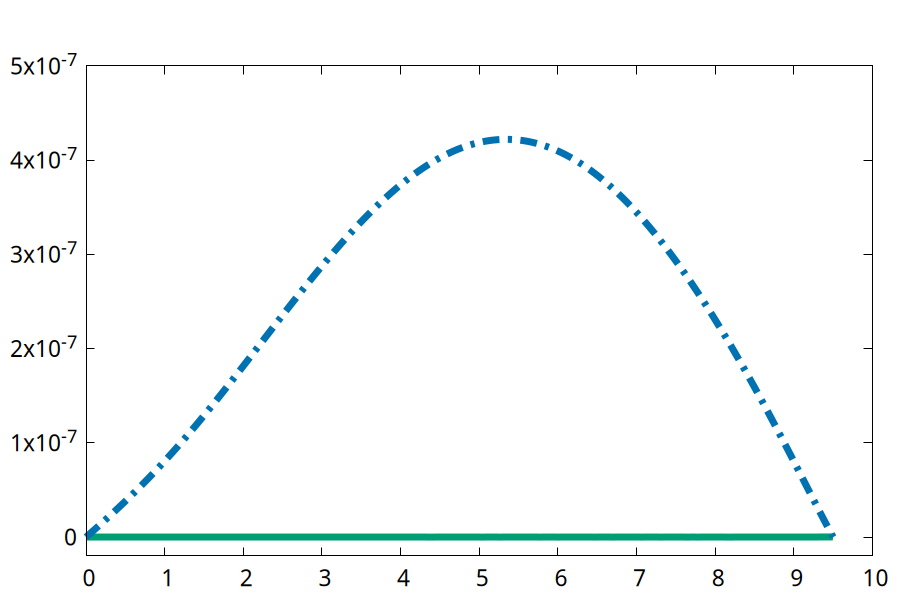}
\caption{Absolute deviation $|\Psi_T(x)-\psi(x)|$ (solid line), and the estimated absolute upper bound $|\eta_{ex}(x)|$ (dashed line) for Test Problem~2 with Dirichlet conditions.} 
\label{fig:MAIN_NL_ERROR_ESTIMATION}
\end{figure}

\noindent
In order to demonstrate the proposed deviation bounding technique presented in Section~\ref{UPPER_BOUND}, we applied it on Test Problems~1 and~2 for various types of conditions. Table~\ref{MAIN_NL_ERROR_ESTIMATION} reports the mean (MSD), and maximum (MXD) squared deviation from the exact solution, as well as the mean (MSED), and maximum (MXED) estimated deviation. In Test Problem~1, the neural form had 36 network parameters, augmented by 18 additional parameters for $\eta(x,\gamma)$, while the training dataset consisted of 80 equidistant points in $[0,1.5]$. In Test Problem~2, the neural form had 180 network parameters, augmented by 60 additional parameters for $\eta(x,\gamma)$, while the training dataset consisted of 270 equidistant points in $[0,9.5]$. In both cases, the reported MSD, MXD, MSED, and MXED metrics were calculated over 1000 equidistant points taken in $[0,1.5]$ for Test Problem~1, and in $[0,9.5]$ for Test Problem~2. For the latter case, Fig.~\ref{fig:MAIN_NL_ERROR_ESTIMATION} illustrates the absolute deviation and the corresponding estimated upper bound across the ODE's domain.

In terms of CPU time, both the augmented neural forms and the baseline neural method were quite demanding. However, the baseline method required slightly more processing time due to the need for adjusting the penalty coefficient. The \texttt{ode113\,} solver proved to be far more efficient, as expected for methods based on finite differences. However, since finite difference methods do not scale well in many dimensions, the augmented neural forms provide promising alternatives for PDEs.

\section{Conclusions}
\label{CONCLUDE}

\noindent
Augmented neural forms offer a reliable approach for solving ODEs under a variety of boundary and initial conditions. The present work contributes to the existing literature by introducing a systematic procedure for designing appropriate trial solutions with neural forms that exactly satisfy the prescribed problem conditions. Furthermore, a novel technique is presented that transforms a problem with Neumann or Robin boundary conditions into one with parametric Dirichlet conditions.  This transformation provides an alternative methodology for formulating proper trial solutions using neural forms.
Furthermore, an upper bound for the absolute deviation of the obtained solution from the exact one was introduced, providing a new evaluation metric for the quality of the approximation.
For initial value problems, higher accuracy may be achieved by employing domain decomposition techniques. 

The proposed augmented neural forms were evaluated on test problems consisting of first- and second-order ODEs, as well as first-order ODE systems, under a variety of boundary and initial conditions. The obtained solutions were compared to the exact ones, as well as to solutions obtained by the common penalized neural method, and to solutions provided by state-of-the-art numerical solvers such as \texttt{ode113} of \textsc{Matlab}\textsuperscript{\textregistered}. 
The comparison clearly demonstrated that the augmented neural forms can provide accurate solutions with superior generalization performance. 
Taking also into consideration their closed form type of solution and the exact satisfaction of the boundary/initial conditions, we can infer that the proposed augmented neural forms constitute a robust and reliable alternative for solving ODE's, while also laying the ground to the treatment of PDEs.

Moreover, our analysis indicates that the careful selection of optimizer for training the employed neural networks may offer advantages. Commonly utilized methods crafted for big data applications, such as stochastic gradient descent, Adam, and variants, may not perform well. Instead, pattern search and quasi-Newton methods were found to be very effective. 
In addition, we have noticed that Chebyshev grids performed better than equidistant grids for the majority of the considered test problems. This advantage stems from their capability to mitigate the Runge phenomenon which is related to overtraining of the neural networks.

Overall, the proposed approach is easy to implement and offers a differentiable closed-form solution that can be directly employed in subsequent calculations. Deep or shallow neural networks with any activation function meeting the universal approximation requirements may be used. 
The proposed ideas have the potential to provide a solid foundation for future initiatives. It might be of interest to consider the recently introduced functionally weighted neural network~\cite{BLEKAS_2017, BLEKAS_2020}, which is a special type of feedforward neural network with advanced extrapolation capabilities. Furthermore, the proposed ideas could be extended in a straightforward manner to treat PDEs in rectangular domains. The challenging problem of PDEs defined inside non-rectangular domains certainly merits attention and requires further investigation.


\bibliographystyle{elsarticle-num} 
\biboptions{sort&compress}
\bibliography{NNFORMS}


\newpage
\appendix

\section{Results for the \texttt{ode113} solver}
\label{appendix:ode113}

\begin{table}[h!]
\centering
\footnotesize
\begin{tabular}{@{\extracolsep{2pt}}ccccccccc@{}}
\hline
Test & Condition & Number of & Spline Polynomial & & \\
Problem & Type & Training Points & Degree & & \multicolumn{2}{c}{Training points} & \multicolumn{2}{c}{Test dataset} \\
\cline{1-5} \cline{6-7} \cline{8-9}
1 &  & & & & MSD & MXD & MSD & MXD \\
& D  & 300 & 3 & & 0.58e-28 & 0.18e-26 & 0.12e-16 & 0.76e-14 \\
&    &  & 5 & &  &  & 0.12e-23 & 0.99e-21 \\
&    &  & 7 & &  &  & 0.15e-13 & 0.15e-10 \\
\cline{1-5} \cline{6-7} \cline{8-9}
3 &  & & & & MSD & MXD & MSD & MXD \\
& C  & 215 & 3  & & 0.31e-29 & 0.28e-28 & 0.14e-15 & 0.18e-14 \\
&   &  & 5 & &  & & 0.12e-16 & 0.73e-14 \\
&   &  & 7 & &  & & 0.70e+18 & 0.33e+21 \\
\cline{1-5} \cline{6-7} \cline{8-9}
4 &  & & & & MSD & MXD & MSD & MXD \\
& D  & 109  & 3 & $\psi_1$ & 0.20e-24 & 0.51e-23 & 0.56e-16 & 0.27e-15 \\
&   &          &     & $\psi_2$ & 0.14e-23 & 0.41e-22 & 0.19e-23 & 0.41e-22 \\
&   &  & 5  & $\psi_1$ &  &  & 0.91e-24 & 0.63e-22 \\
&   &  &    & $\psi_2$ &  &  & 0.20e-23 & 0.75e-22 \\
&   &  & 7  & $\psi_1$ &  &  & 0.10e-08 & 0.54e-06 \\
&   &  &    & $\psi_2$ &  &  & 0.80e-06 & 0.42e-03 \\
\cline{1-5} \cline{6-7} \cline{8-9}
5 &  & & & & MSD & MXD & MSD & MXD \\
& D  & 321 & 3 & $\psi_1$ & 0.52e-29 & 0.86e-28 & 0.97e-17 & 0.14e-14 \\
&   &      &   & $\psi_2$ & 0.16e-28 & 0.31e-27 & 0.22e-16 & 0.31e-14 \\
&   &      & 5 & $\psi_1$ &          &          & 0.72e-24 & 0.37e-21 \\
&   &      &   & $\psi_2$ &          &          & 0.16e-23 & 0.84e-21 \\
&   &      & 7 & $\psi_1$ &          &          & 0.75e-09 & 0.59e-06 \\
&   &      &   & $\psi_2$ &          &          & 0.16e-11 & 0.13e-08 \\
\cline{1-5} \cline{6-7} \cline{8-9}
6 &  & & & & MSD & MXD & MSD & MXD \\
& D-D   & 203     & 3   & & 0.13e-26 & 0.56e-26 & 0.11e-16 & 0.34e-14 \\
&       &         & 5   & &          &          & 0.63e-22 & 0.52e-19 \\
&       &         & 7   & &          &          & 0.35e-16 & 0.27e-13 \\
& D-N   & 203     & 3   & & 0.11e-26 & 0.46e-26 & 0.11e-16 & 0.34e-14 \\
&       &         & 5   & &          &          & 0.63e-22 & 0.52e-19 \\
&       &         & 7   & &          &          & 0.35e-16 & 0.26e-13 \\
& N-N   & 203     & 3   & & 0.42e-27 & 0.14e-26 & 0.11e-16 & 0.34e-14 \\
&       &         & 5   & &          &          & 0.63e-22 & 0.52e-19 \\
&       &         & 7   & &          &          & 0.52e-16 & 0.40e-13 \\
& C     & 203     & 3   & & 0.27e-24 & 0.42e-23 & 0.11e-16 & 0.34e-14 \\
&       &         & 5   & &          &          & 0.64e-22 & 0.53e-19 \\
&       &         & 7   & &          &          & 0.41e-16 & 0.31e-13 \\
\hline
\multicolumn{9}{l}{\scriptsize Abbreviations: D=Dirichlet, N=Neumann, D-N=Mixed Dirichlet-Neumann, C=Cauchy}
\end{tabular}
\caption{Mean (MSD) and maximum (MXD) squared deviation of solutions obtained by \texttt{ode113} solver of \textsc{Matlab}\textsuperscript{\textregistered} per test problem and condition type, calculated over the training, and test datasets of 1000 equidistant points with cubic, quintic, and septic spline interpolation.}
\label{tab:ode113_results}
\end{table}

\newpage

\section{Results for the baseline neural method (standalone neural network)}
\label{appendix:NN}


\begin{table}[h!]
\setlength{\tabcolsep}{2pt}
\centering 
\scriptsize
\begin{tabular}{@{\extracolsep{3pt}}ccccccccccccc@{}}
\hline
\multicolumn{13}{c}{\textbf{Measure: Deviation~~~~Grid type: Equidistant}}\\
\hline
  & \multicolumn{12}{c}{Network Parameters} \\
  & \multicolumn{4}{c}{36} & \multicolumn{4}{c}{72} & \multicolumn{4}{c}{144} \\
\cline{2-5} \cline{6-9} \cline{10-13}
  & \multicolumn{2}{c}{Training} & \multicolumn{2}{c}{Test} & \multicolumn{2}{c}{Training} & \multicolumn{2}{c}{Test} & \multicolumn{2}{c}{Training} & \multicolumn{2}{c}{Test} \\
\cline{2-3} \cline{4-5} \cline{6-7} \cline{8-9} \cline{10-11} \cline{12-13}
$M_{\rm tr}$ & MSD & MXD & MSD & MXD & MSD & MXD & MSD & MXD & MSD & MXD & MSD & MXD \\
\hline
40 & 0.13e-10 &  0.53e-09 &  0.30e-10 &  0.18e-08 & & & & & & & & \\
80 & 0.44e-16 &  0.30e-14 &  0.56e-16 &  0.47e-14 &  0.12e-14 &  0.82e-13 &  0.15e-14 &  0.13e-12 & & & & \\
160 & 0.41e-18 &  0.36e-16 &  0.31e-18 &  0.36e-16 &  0.93e-18 &  0.73e-16 &  0.90e-18 &  0.87e-16 &  0.56e-18 &  0.36e-16 &  0.48e-18 &  0.43e-16 \\
\hline
\multicolumn{13}{c}{\textbf{Measure: Deviation~~~~Grid type: Chebyshev}}\\
\hline
 & \multicolumn{12}{c}{Network Parameters} \\
 & \multicolumn{4}{c}{36} & \multicolumn{4}{c}{72} & \multicolumn{4}{c}{144} \\
\cline{2-5}\cline{6-9}\cline{10-13}
 & \multicolumn{2}{c}{Training} & \multicolumn{2}{c}{Test} & \multicolumn{2}{c}{Training} & \multicolumn{2}{c}{Test}  & \multicolumn{2}{c}{Training} & \multicolumn{2}{c}{Test} \\
\cline{2-3} \cline{4-5} \cline{6-7} \cline{8-9} \cline{10-11} \cline{12-13}
$M_{\rm tr}$ & MSD & MXD & MSD & MXD & MSD & MXD & MSD & MXD & MSD & MXD & MSD & MXD \\
\hline
40 & 0.16e-17 &  0.33e-16 &  0.26e-18 &  0.36e-16 & & & & & & & & \\
80 & 0.16e-17 &  0.35e-16 &  0.25e-18 &  0.36e-16 &  0.16e-17 &  0.35e-16 &  0.26e-18 &  0.36e-16 & & & & \\
160 & 0.16e-17 &  0.35e-16 &  0.26e-18 &  0.36e-16 &  0.16e-17 &  0.35e-16 &  0.26e-18 &  0.36e-16 &  0.16e-17 &  0.35e-16 &  0.25e-18 &  0.35e-16 \\
\hline
\multicolumn{13}{c}{\textbf{Measure: Error~~~~Grid type: Equidistant}}\\
\hline
 & \multicolumn{12}{c}{Network Parameters} \\
 & \multicolumn{4}{c}{36} & \multicolumn{4}{c}{72} & \multicolumn{4}{c}{144} \\
\cline{2-5}\cline{6-9}\cline{10-13}
 & \multicolumn{2}{c}{Training} & \multicolumn{2}{c}{Test} & \multicolumn{2}{c}{Training} & \multicolumn{2}{c}{Test}  & \multicolumn{2}{c}{Training} & \multicolumn{2}{c}{Test} \\
\cline{2-3} \cline{4-5} \cline{6-7} \cline{8-9} \cline{10-11} \cline{12-13}
$M_{\rm tr}$ & MSE & MXE & MSE & MXE & MSE & MXE & MSE & MXE & MSE & MXE & MSE & MXE \\
\hline
40 & 0.60e-19 &  0.38e-18 &  0.18e-06 &  0.22e-04 & & & & & & & & \\
80 & 0.51e-18 &  0.19e-17 &  0.50e-12 &  0.11e-09 &  0.57e-18 &  0.56e-17 &  0.15e-10 &  0.32e-08 & & & & \\
160 & 0.65e-17 &  0.24e-15 &  0.76e-16 &  0.24e-13 &  0.88e-18 &  0.22e-16 &  0.42e-14 &  0.15e-11 &  0.77e-19 &  0.26e-17 &  0.77e-15 &  0.29e-12 \\
\hline
\multicolumn{13}{c}{\textbf{Measure: Error~~~~Grid type: Chebyshev}}\\
\hline
 & \multicolumn{12}{c}{Network Parameters} \\
 & \multicolumn{4}{c}{36} & \multicolumn{4}{c}{72} & \multicolumn{4}{c}{144} \\
\cline{2-5}\cline{6-9}\cline{10-13}
 & \multicolumn{2}{c}{Training} & \multicolumn{2}{c}{Test} & \multicolumn{2}{c}{Training} & \multicolumn{2}{c}{Test}  & \multicolumn{2}{c}{Training} & \multicolumn{2}{c}{Test} \\
\cline{2-3} \cline{4-5} \cline{6-7} \cline{8-9} \cline{10-11} \cline{12-13}
$M_{\rm tr}$ & MSE & MXE & MSE & MXE & MSE & MXE & MSE & MXE & MSE & MXE & MSE & MXE \\
\hline
40 & 0.19e-16 &  0.18e-15 &  0.12e-16 &  0.21e-15 & & & & & & & & \\
80 & 0.43e-18 &  0.46e-17 &  0.35e-18 &  0.50e-17 &  0.11e-17 &  0.62e-17 &  0.10e-17 &  0.65e-17 & & & & \\
160 & 0.30e-17 &  0.14e-16 &  0.27e-17 &  0.15e-16 &  0.16e-17 &  0.17e-16 &  0.11e-17 &  0.20e-16 &  0.95e-18 &  0.45e-17 &  0.12e-17 &  0.46e-17 \\
\hline
\end{tabular}
\caption{Performance measures for all scenarios of Test Problem 1.}
\label{tab:Res_Prob_1_SNN}
 \end{table} 



\begin{table}[h!]
\setlength{\tabcolsep}{2pt}
\centering
\scriptsize
\begin{tabular}{@{\extracolsep{3pt}}ccccccccccccc@{}}
\hline
\multicolumn{13}{c}{\textbf{Measure: Deviation~~~~Grid type: Equidistant}}\\
\hline
& \multicolumn{12}{c}{Network Parameters} \\
& \multicolumn{4}{c}{90} & \multicolumn{4}{c}{180} & \multicolumn{4}{c}{270} \\
\cline{2-5} \cline{6-9} \cline{10-13}
& \multicolumn{2}{c}{Training} & \multicolumn{2}{c}{Test} & \multicolumn{2}{c}{Training} & \multicolumn{2}{c}{Test} & \multicolumn{2}{c}{Training} & \multicolumn{2}{c}{Test} \\
\cline{2-3} \cline{4-5} \cline{6-7} \cline{8-9} \cline{10-11} \cline{12-13}
$M_{\rm tr}$ & MSD & MXD & MSD & MXD & MSD & MXD & MSD & MXD & MSD & MXD & MSD & MXD \\
\hline
90  &  0.10e-15 &  0.40e-14 &  0.12e-15 &  0.54e-14 & & & & & & & & \\
180 &  0.13e-21 &  0.56e-20 &  0.14e-21 &  0.69e-20 &  0.31e-21 &  0.33e-20 &  0.32e-21 &  0.39e-20 & & & & \\
270 &  0.33e-22 &  0.19e-21 &  0.33e-22 &  0.19e-21 &  0.11e-21 &  0.65e-21 &  0.11e-21 &  0.65e-21 &  0.20e-21 &  0.15e-20 &  0.20e-21 &  0.15e-20 \\
\hline
\multicolumn{13}{c}{\textbf{Measure: Deviation~~~~Grid type: Chebyshev}}\\
\hline
& \multicolumn{12}{c}{Network Parameters} \\
& \multicolumn{4}{c}{90} & \multicolumn{4}{c}{180} & \multicolumn{4}{c}{270} \\
\cline{2-5} \cline{6-9} \cline{10-13}
& \multicolumn{2}{c}{Training} & \multicolumn{2}{c}{Test} & \multicolumn{2}{c}{Training} & \multicolumn{2}{c}{Test} & \multicolumn{2}{c}{Training} & \multicolumn{2}{c}{Test} \\
\cline{2-3} \cline{4-5} \cline{6-7} \cline{8-9} \cline{10-11} \cline{12-13}
$M_{\rm tr}$ & MSD & MXD & MSD & MXD & MSD & MXD & MSD & MXD & MSD & MXD & MSD & MXD \\
\hline
90  &  0.10e-21 &  0.61e-21 &  0.14e-21 &  0.61e-21 & & & & & & & & \\
180 &  0.90e-23 &  0.68e-22 &  0.13e-22 &  0.69e-22 &  0.46e-21 &  0.34e-20 &  0.66e-21 &  0.34e-20 & & & & \\
270 &  0.88e-23 &  0.85e-22 &  0.13e-22 &  0.85e-22 &  0.95e-22 &  0.12e-20 &  0.13e-21 &  0.12e-20 &  0.37e-21 &  0.56e-20 &  0.55e-21 &  0.56e-20 \\
\hline
\multicolumn{13}{c}{\textbf{Measure: Error~~~~Grid type: Equidistant}}\\
\hline
& \multicolumn{12}{c}{Network Parameters} \\
& \multicolumn{4}{c}{90} & \multicolumn{4}{c}{180} & \multicolumn{4}{c}{270} \\
\cline{2-5} \cline{6-9} \cline{10-13}
& \multicolumn{2}{c}{Training} & \multicolumn{2}{c}{Test} & \multicolumn{2}{c}{Training}  & \multicolumn{2}{c}{Test} & \multicolumn{2}{c}{Training} & \multicolumn{2}{c}{Test} \\
\cline{2-3} \cline{4-5} \cline{6-7} \cline{8-9} \cline{10-11} \cline{12-13}
$M_{\rm tr}$ & MSE & MXE & MSE & MXE & MSE & MXE & MSE & MXE & MSE & MXE & MSE & MXE \\
\hline
90  &  0.35e-19 &  0.21e-18 &  0.12e-10 &  0.29e-08 & & & & & & & & \\
180 &  0.19e-19 &  0.22e-18 &  0.89e-16 &  0.38e-13 &  0.20e-18 &  0.11e-17 &  0.49e-16 &  0.22e-13 & & & & \\
270 &  0.25e-19 &  0.16e-18 &  0.33e-18 &  0.20e-15 &  0.32e-19 &  0.63e-18 &  0.10e-17 &  0.65e-15 &  0.12e-18 &  0.23e-17 &  0.12e-18 &  0.23e-17 \\
\hline
\multicolumn{13}{c}{\textbf{Measure: Error~~~~Grid type: Chebyshev}}\\
\hline
& \multicolumn{12}{c}{Network Parameters} \\
& \multicolumn{4}{c}{90} & \multicolumn{4}{c}{180} & \multicolumn{4}{c}{270} \\
\cline{2-5} \cline{6-9} \cline{10-13}
& \multicolumn{2}{c}{Training} & \multicolumn{2}{c}{Test} & \multicolumn{2}{c}{Training} & \multicolumn{2}{c}{Test} & \multicolumn{2}{c}{Training} & \multicolumn{2}{c}{Test} \\
\cline{2-3} \cline{4-5} \cline{6-7} \cline{8-9} \cline{10-11} \cline{12-13}
$M_{\rm tr}$ & MSE & MXE & MSE & MXE & MSE & MXE & MSE & MXE & MSE & MXE & MSE & MXE \\
\hline
90  &  0.32e-19 &  0.93e-19 &  0.37e-19 &  0.10e-18 & & & & & & & & \\
180 &  0.11e-19 &  0.51e-19 &  0.11e-19 &  0.53e-19 &  0.34e-18 &  0.13e-17 &  0.39e-18 &  0.13e-17 & & & & \\
270 &  0.20e-19 &  0.12e-18 &  0.17e-19 &  0.12e-18 &  0.20e-18 &  0.14e-17 &  0.15e-18 &  0.14e-17 &  0.12e-17 &  0.70e-17 &  0.81e-18 &  0.78e-17 \\
\hline
\end{tabular}
\caption{Performance measures for all scenarios of Test Problem 2 with Dirichlet conditions.}
\label{tab:Res_Prob_2_DD_SNN}
\end{table}

\begin{table}[h!]
\setlength{\tabcolsep}{2pt}
\centering
\scriptsize
\begin{tabular}{@{\extracolsep{3pt}}ccccccccccccc@{}}
\hline
\multicolumn{13}{c}{\textbf{Measure: Deviation~~~~Grid type: Equidistant}}\\
\hline
& \multicolumn{12}{c}{Network Parameters} \\
& \multicolumn{4}{c}{90} & \multicolumn{4}{c}{180} & \multicolumn{4}{c}{270} \\
\cline{2-5} \cline{6-9} \cline{10-13}
& \multicolumn{2}{c}{Training} & \multicolumn{2}{c}{Test} & \multicolumn{2}{c}{Training} & \multicolumn{2}{c}{Test} & \multicolumn{2}{c}{Training} & \multicolumn{2}{c}{Test} \\
\cline{2-3} \cline{4-5} \cline{6-7} \cline{8-9} \cline{10-11} \cline{12-13}
$M_{\rm tr}$ & MSD & MXD & MSD & MXD & MSD & MXD & MSD & MXD & MSD & MXD & MSD & MXD \\
\hline
90  &  0.10e-14 &  0.37e-13 &  0.84e-15 &  0.37e-13 & & & & & & & & \\
180 &  0.27e-19 &  0.12e-17 &  0.24e-19 &  0.12e-17 &  0.76e-20 &  0.34e-18 &  0.69e-20 &  0.34e-18 & & & & \\
270 &  0.20e-21 &  0.80e-20 &  0.19e-21 &  0.80e-20 &  0.48e-22 &  0.32e-21 &  0.48e-22 &  0.32e-21 &  0.21e-18 &  0.99e-17 &  0.19e-18 &  0.99e-17 \\
\hline
\multicolumn{13}{c}{\textbf{Measure: Deviation~~~~Grid type: Chebyshev}}\\
\hline
& \multicolumn{12}{c}{Network Parameters} \\
& \multicolumn{4}{c}{90} & \multicolumn{4}{c}{180} & \multicolumn{4}{c}{270} \\
\cline{2-5} \cline{6-9} \cline{10-13}
& \multicolumn{2}{c}{Training} & \multicolumn{2}{c}{Test} & \multicolumn{2}{c}{Training} & \multicolumn{2}{c}{Test} & \multicolumn{2}{c}{Training} & \multicolumn{2}{c}{Test} \\
\cline{2-3} \cline{4-5} \cline{6-7} \cline{8-9} \cline{10-11} \cline{12-13}
$M_{\rm tr}$ & MSD & MXD & MSD & MXD & MSD & MXD & MSD & MXD & MSD & MXD & MSD & MXD \\
\hline
90  &  0.95e-22 &  0.76e-21 &  0.13e-21 &  0.76e-21 & & & & & & & & \\
180 &  0.37e-22 &  0.33e-21 &  0.49e-22 &  0.33e-21 &  0.54e-22 &  0.49e-21 &  0.78e-22 &  0.49e-21 & & & & \\
270 &  0.36e-22 &  0.29e-21 &  0.52e-22 &  0.29e-21 &  0.39e-21 &  0.24e-20 &  0.58e-21 &  0.24e-20 &  0.12e-20 &  0.10e-19 &  0.17e-20 &  0.10e-19 \\
\hline
\multicolumn{13}{c}{\textbf{Measure: Error~~~~Grid type: Equidistant}}\\
\hline
& \multicolumn{12}{c}{Network Parameters} \\
& \multicolumn{4}{c}{90} & \multicolumn{4}{c}{180} & \multicolumn{4}{c}{270} \\
\cline{2-5} \cline{6-9} \cline{10-13}
& \multicolumn{2}{c}{Training} & \multicolumn{2}{c}{Test} & \multicolumn{2}{c}{Training}  & \multicolumn{2}{c}{Test} & \multicolumn{2}{c}{Training} & \multicolumn{2}{c}{Test} \\
\cline{2-3} \cline{4-5} \cline{6-7} \cline{8-9} \cline{10-11} \cline{12-13}
$M_{\rm tr}$ & MSE & MXE & MSE & MXE & MSE & MXE & MSE & MXE & MSE & MXE & MSE & MXE \\
\hline
90  &  0.15e-20 &  0.46e-19 &  0.11e-11 &  0.27e-09 & & & & & & & & \\
180 &  0.44e-20 &  0.15e-18 &  0.65e-16 &  0.27e-13 &  0.17e-19 &  0.13e-18 &  0.18e-16 &  0.78e-14 & & & & \\
270 &  0.12e-19 &  0.44e-18 &  0.67e-18 &  0.42e-15 &  0.34e-19 &  0.26e-18 &  0.60e-19 &  0.17e-16 &  0.32e-18 &  0.85e-17 &  0.77e-15 &  0.53e-12 \\
\hline
\multicolumn{13}{c}{\textbf{Measure: Error~~~~Grid type: Chebyshev}}\\
\hline
& \multicolumn{12}{c}{Network Parameters} \\
& \multicolumn{4}{c}{90} & \multicolumn{4}{c}{180} & \multicolumn{4}{c}{270} \\
\cline{2-5} \cline{6-9} \cline{10-13}
& \multicolumn{2}{c}{Training} & \multicolumn{2}{c}{Test} & \multicolumn{2}{c}{Training} & \multicolumn{2}{c}{Test} & \multicolumn{2}{c}{Training} & \multicolumn{2}{c}{Test} \\
\cline{2-3} \cline{4-5} \cline{6-7} \cline{8-9} \cline{10-11} \cline{12-13}
$M_{\rm tr}$ & MSE & MXE & MSE & MXE & MSE & MXE & MSE & MXE & MSE & MXE & MSE & MXE \\
\hline
90  &  0.26e-19 &  0.24e-18 &  0.24e-19 &  0.24e-18 & & & & & & & & \\
180 &  0.22e-19 &  0.18e-18 &  0.14e-19 &  0.20e-18 &  0.12e-18 &  0.13e-17 &  0.61e-19 &  0.17e-17 & & & & \\
270 &  0.22e-19 &  0.13e-18 &  0.23e-19 &  0.13e-18 &  0.19e-18 &  0.15e-17 &  0.14e-18 &  0.15e-17 &  0.96e-18 &  0.75e-17 &  0.67e-18 &  0.75e-17 \\
\hline
\end{tabular}
\caption{Performance measures for all scenarios of Test Problem 2 with mixed Dirichlet-Neumann conditions.}
\label{tab:Res_Prob_2_DN_SNN}
\end{table}


\begin{table}[h!]
\setlength{\tabcolsep}{2pt}
\centering
\scriptsize
\begin{tabular}{@{\extracolsep{3pt}}ccccccccccccc@{}}
\hline
\multicolumn{13}{c}{\textbf{Measure: Deviation~~~~Grid type: Equidistant}}\\
\hline
& \multicolumn{12}{c}{Network Parameters} \\
& \multicolumn{4}{c}{90} & \multicolumn{4}{c}{180} & \multicolumn{4}{c}{270} \\
\cline{2-5} \cline{6-9} \cline{10-13}
& \multicolumn{2}{c}{Training} & \multicolumn{2}{c}{Test} & \multicolumn{2}{c}{Training} & \multicolumn{2}{c}{Test} & \multicolumn{2}{c}{Training} & \multicolumn{2}{c}{Test} \\
\cline{2-3} \cline{4-5} \cline{6-7} \cline{8-9} \cline{10-11} \cline{12-13}
$M_{\rm tr}$ & MSD & MXD & MSD & MXD & MSD & MXD & MSD & MXD & MSD & MXD & MSD & MXD \\
\hline
90  &  0.38e-19 &  0.16e-17 &  0.29e-19 &  0.17e-17 & & & & & & & & \\
180 &  0.52e-20 &  0.35e-18 &  0.42e-20 &  0.35e-18 &  0.98e-20 &  0.46e-18 &  0.87e-20 &  0.46e-18 & & & & \\
270 &  0.67e-20 &  0.35e-18 &  0.62e-20 &  0.35e-18 &  0.15e-19 &  0.77e-18 &  0.14e-19 &  0.77e-18 &  0.32e-19 &  0.16e-17 &  0.30e-19 &  0.16e-17 \\
\hline
\multicolumn{13}{c}{\textbf{Measure: Deviation~~~~Grid type: Chebyshev}}\\
\hline
& \multicolumn{12}{c}{Network Parameters} \\
& \multicolumn{4}{c}{90} & \multicolumn{4}{c}{180} & \multicolumn{4}{c}{270} \\
\cline{2-5} \cline{6-9} \cline{10-13}
& \multicolumn{2}{c}{Training} & \multicolumn{2}{c}{Test} & \multicolumn{2}{c}{Training} & \multicolumn{2}{c}{Test} & \multicolumn{2}{c}{Training} & \multicolumn{2}{c}{Test} \\
\cline{2-3} \cline{4-5} \cline{6-7} \cline{8-9} \cline{10-11} \cline{12-13}
$M_{\rm tr}$ & MSD & MXD & MSD & MXD & MSD & MXD & MSD & MXD & MSD & MXD & MSD & MXD \\
\hline
90  &  0.11e-19 &  0.15e-18 &  0.28e-20 &  0.15e-18 & & & & & & & & \\
180 &  0.44e-19 &  0.61e-18 &  0.11e-19 &  0.62e-18 &  0.14e-19 &  0.20e-18 &  0.36e-20 &  0.20e-18 & & & & \\
270 &  0.75e-20 &  0.10e-18 &  0.19e-20 &  0.10e-18 &  0.58e-20 &  0.81e-19 &  0.15e-20 &  0.81e-19 &  0.43e-21 &  0.60e-20 &  0.11e-21 &  0.60e-20 \\
\hline
\multicolumn{13}{c}{\textbf{Measure: Error~~~~Grid type: Equidistant}}\\
\hline
& \multicolumn{12}{c}{Network Parameters} \\
& \multicolumn{4}{c}{90} & \multicolumn{4}{c}{180} & \multicolumn{4}{c}{270} \\
\cline{2-5} \cline{6-9} \cline{10-13}
& \multicolumn{2}{c}{Training} & \multicolumn{2}{c}{Test} & \multicolumn{2}{c}{Training}  & \multicolumn{2}{c}{Test} & \multicolumn{2}{c}{Training} & \multicolumn{2}{c}{Test} \\
\cline{2-3} \cline{4-5} \cline{6-7} \cline{8-9} \cline{10-11} \cline{12-13}
$M_{\rm tr}$ & MSE & MXE & MSE & MXE & MSE & MXE & MSE & MXE & MSE & MXE & MSE & MXE \\
\hline
90  &  0.28e-20 &  0.22e-19 &  0.10e-14 &  0.37e-12 & & & & & & & & \\
180 &  0.41e-20 &  0.77e-19 &  0.48e-16 &  0.20e-13 &  0.60e-20 &  0.18e-19 &  0.16e-17 &  0.63e-15 & & & & \\
270 &  0.80e-20 &  0.38e-18 &  0.66e-18 &  0.42e-15 &  0.11e-19 &  0.47e-18 &  0.95e-18 &  0.61e-15 &  0.71e-20 &  0.20e-19 &  0.77e-20 &  0.40e-18 \\
\hline
\multicolumn{13}{c}{\textbf{Measure: Error~~~~Grid type: Chebyshev}}\\
\hline
& \multicolumn{12}{c}{Network Parameters} \\
& \multicolumn{4}{c}{90} & \multicolumn{4}{c}{180} & \multicolumn{4}{c}{270} \\
\cline{2-5} \cline{6-9} \cline{10-13}
& \multicolumn{2}{c}{Training} & \multicolumn{2}{c}{Test} & \multicolumn{2}{c}{Training} & \multicolumn{2}{c}{Test} & \multicolumn{2}{c}{Training} & \multicolumn{2}{c}{Test} \\
\cline{2-3} \cline{4-5} \cline{6-7} \cline{8-9} \cline{10-11} \cline{12-13}
$M_{\rm tr}$ & MSE & MXE & MSE & MXE & MSE & MXE & MSE & MXE & MSE & MXE & MSE & MXE \\
\hline
90  &  0.67e-20 &  0.86e-19 &  0.35e-20 &  0.89e-19 & & & & & & & & \\
180 &  0.60e-20 &  0.59e-19 &  0.33e-20 &  0.70e-19 &  0.32e-20 &  0.37e-19 &  0.19e-20 &  0.37e-19 & & & & \\
270 &  0.15e-20 &  0.17e-19 &  0.10e-20 &  0.20e-19 &  0.22e-20 &  0.23e-19 &  0.15e-20 &  0.26e-19 &  0.37e-20 &  0.21e-19 &  0.28e-20 &  0.21e-19 \\
\hline
\end{tabular}
\caption{Performance measures for all scenarios of Test Problem 2 with Cauchy conditions.}
\label{tab:Res_Prob_2_C_SNN}
\end{table}

\begin{table}[h!]
\setlength{\tabcolsep}{2pt}
\centering
\scriptsize
\begin{tabular}{@{\extracolsep{3pt}}ccccccccccccc@{}}
\hline
\multicolumn{13}{c}{\textbf{Measure: Deviation~~~~Grid type: Equidistant}}\\
\hline
& \multicolumn{12}{c}{Network Parameters} \\
& \multicolumn{4}{c}{90} & \multicolumn{4}{c}{180} & \multicolumn{4}{c}{270} \\
\cline{2-5} \cline{6-9} \cline{10-13}
& \multicolumn{2}{c}{Training} & \multicolumn{2}{c}{Test} & \multicolumn{2}{c}{Training} & \multicolumn{2}{c}{Test} & \multicolumn{2}{c}{Training} & \multicolumn{2}{c}{Test} \\
\cline{2-3} \cline{4-5} \cline{6-7} \cline{8-9} \cline{10-11} \cline{12-13}
$M_{\rm tr}$ & MSD & MXD & MSD & MXD & MSD & MXD & MSD & MXD & MSD & MXD & MSD & MXD \\
\hline
90  &  0.39e-14 &  0.14e-12 &  0.32e-14 &  0.14e-12 & & & & & & & & \\
180 &  0.62e-20 &  0.26e-18 &  0.56e-20 &  0.26e-18 &  0.77e-19 &  0.34e-17 &  0.70e-19 &  0.34e-17 & & & & \\
270 &  0.50e-22 &  0.17e-20 &  0.48e-22 &  0.17e-20 &  0.70e-21 &  0.17e-19 &  0.68e-21 &  0.17e-19 &  0.24e-19 &  0.11e-17 &  0.22e-19 &  0.11e-17 \\
\hline
\multicolumn{13}{c}{\textbf{Measure: Deviation~~~~Grid type: Chebyshev}}\\
\hline
& \multicolumn{12}{c}{Network Parameters} \\
& \multicolumn{4}{c}{90} & \multicolumn{4}{c}{180} & \multicolumn{4}{c}{270} \\
\cline{2-5} \cline{6-9} \cline{10-13}
& \multicolumn{2}{c}{Training} & \multicolumn{2}{c}{Test} & \multicolumn{2}{c}{Training} & \multicolumn{2}{c}{Test} & \multicolumn{2}{c}{Training} & \multicolumn{2}{c}{Test} \\
\cline{2-3} \cline{4-5} \cline{6-7} \cline{8-9} \cline{10-11} \cline{12-13}
$M_{\rm tr}$ & MSD & MXD & MSD & MXD & MSD & MXD & MSD & MXD & MSD & MXD & MSD & MXD \\
\hline
90  &  0.23e-21 &  0.83e-21 &  0.18e-21 &  0.83e-21 & & & & & & & & \\
180 &  0.10e-20 &  0.33e-20 &  0.94e-21 &  0.33e-20 &  0.19e-22 &  0.15e-21 &  0.27e-22 &  0.15e-21 & & & & \\
270 &  0.15e-20 &  0.69e-20 &  0.15e-20 &  0.69e-20 &  0.55e-21 &  0.14e-20 &  0.50e-21 &  0.14e-20 &  0.82e-22 &  0.20e-21 &  0.81e-22 &  0.20e-21 \\
\hline
\multicolumn{13}{c}{\textbf{Measure: Error~~~~Grid type: Equidistant}}\\
\hline
& \multicolumn{12}{c}{Network Parameters} \\
& \multicolumn{4}{c}{90} & \multicolumn{4}{c}{180} & \multicolumn{4}{c}{270} \\
\cline{2-5} \cline{6-9} \cline{10-13}
& \multicolumn{2}{c}{Training} & \multicolumn{2}{c}{Test} & \multicolumn{2}{c}{Training}  & \multicolumn{2}{c}{Test} & \multicolumn{2}{c}{Training} & \multicolumn{2}{c}{Test} \\
\cline{2-3} \cline{4-5} \cline{6-7} \cline{8-9} \cline{10-11} \cline{12-13}
$M_{\rm tr}$ & MSE & MXE & MSE & MXE & MSE & MXE & MSE & MXE & MSE & MXE & MSE & MXE \\
\hline
90  &  0.17e-21 &  0.14e-20 &  0.41e-11 &  0.11e-08 & & & & & & & & \\
180 &  0.26e-19 &  0.13e-18 &  0.14e-16 &  0.63e-14 &  0.37e-20 &  0.58e-19 &  0.18e-15 &  0.80e-13 & & & & \\
270 &  0.26e-20 &  0.29e-19 &  0.13e-18 &  0.93e-16 &  0.14e-19 &  0.41e-18 &  0.14e-17 &  0.89e-15 &  0.20e-19 &  0.14e-17 &  0.88e-16 &  0.62e-13 \\
\hline
\multicolumn{13}{c}{\textbf{Measure: Error~~~~Grid type: Chebyshev}}\\
\hline
& \multicolumn{12}{c}{Network Parameters} \\
& \multicolumn{4}{c}{90} & \multicolumn{4}{c}{180} & \multicolumn{4}{c}{270} \\
\cline{2-5} \cline{6-9} \cline{10-13}
& \multicolumn{2}{c}{Training} & \multicolumn{2}{c}{Test} & \multicolumn{2}{c}{Training} & \multicolumn{2}{c}{Test} & \multicolumn{2}{c}{Training} & \multicolumn{2}{c}{Test} \\
\cline{2-3} \cline{4-5} \cline{6-7} \cline{8-9} \cline{10-11} \cline{12-13}
$M_{\rm tr}$ & MSE & MXE & MSE & MXE & MSE & MXE & MSE & MXE & MSE & MXE & MSE & MXE \\
\hline
90  &  0.54e-19 &  0.59e-18 &  0.34e-19 &  0.60e-18 & & & & & & & & \\
180 &  0.12e-18 &  0.97e-18 &  0.81e-19 &  0.11e-17 &  0.88e-20 &  0.97e-19 &  0.76e-20 &  0.10e-18 & & & & \\
270 &  0.10e-18 &  0.84e-18 &  0.93e-19 &  0.85e-18 &  0.18e-19 &  0.12e-18 &  0.17e-19 &  0.10e-18 &  0.57e-19 &  0.61e-18 &  0.39e-19 &  0.62e-18 \\
\hline
\end{tabular}
\caption{Performance measures for all scenarios of Test Problem 2 with Neumann conditions.}
\label{tab:Res_Prob_2_NN_SNN}
\end{table}


\begin{table}[h!]
\setlength{\tabcolsep}{2pt}
\centering
\scriptsize
\begin{tabular}{@{\extracolsep{3pt}}ccccccccccccc@{}}
\hline
\multicolumn{13}{c}{\textbf{Measure: Deviation~~~~Grid type: Equidistant}}\\
\hline
& \multicolumn{12}{c}{Network Parameters} \\
& \multicolumn{4}{c}{90} & \multicolumn{4}{c}{180} & \multicolumn{4}{c}{270} \\
\cline{2-5} \cline{6-9} \cline{10-13}
& \multicolumn{2}{c}{Training} & \multicolumn{2}{c}{Test} & \multicolumn{2}{c}{Training} & \multicolumn{2}{c}{Test} & \multicolumn{2}{c}{Training} & \multicolumn{2}{c}{Test} \\
\cline{2-3} \cline{4-5} \cline{6-7} \cline{8-9} \cline{10-11} \cline{12-13}
$M_{\rm tr}$ & MSD & MXD & MSD & MXD & MSD & MXD & MSD & MXD & MSD & MXD & MSD & MXD \\
\hline
90  &  0.82e-15 &  0.27e-13 &  0.71e-15 &  0.29e-13 & & & & & & & & \\
180 &  0.55e-19 &  0.23e-17 &  0.50e-19 &  0.24e-17 &  0.32e-17 &  0.13e-15 &  0.29e-17 &  0.14e-15 & & & & \\
270 &  0.28e-22 &  0.12e-20 &  0.26e-22 &  0.12e-20 &  0.13e-20 &  0.57e-19 &  0.12e-20 &  0.59e-19 &  0.46e-21 &  0.18e-19 &  0.43e-21 &  0.19e-19 \\
\hline
\multicolumn{13}{c}{\textbf{Measure: Deviation~~~~Grid type: Chebyshev}}\\
\hline
& \multicolumn{12}{c}{Network Parameters} \\
& \multicolumn{4}{c}{90} & \multicolumn{4}{c}{180} & \multicolumn{4}{c}{270} \\
\cline{2-5} \cline{6-9} \cline{10-13}
& \multicolumn{2}{c}{Training} & \multicolumn{2}{c}{Test} & \multicolumn{2}{c}{Training} & \multicolumn{2}{c}{Test} & \multicolumn{2}{c}{Training} & \multicolumn{2}{c}{Test} \\
\cline{2-3} \cline{4-5} \cline{6-7} \cline{8-9} \cline{10-11} \cline{12-13}
$M_{\rm tr}$ & MSD & MXD & MSD & MXD & MSD & MXD & MSD & MXD & MSD & MXD & MSD & MXD \\
\hline
90  &  0.10e-21 &  0.76e-21 &  0.15e-21 &  0.76e-21 & & & & & & & & \\
180 &  0.19e-21 &  0.19e-20 &  0.25e-21 &  0.19e-20 &  0.33e-22 &  0.32e-21 &  0.48e-22 &  0.32e-21 & & & & \\
270 &  0.11e-21 &  0.81e-21 &  0.15e-21 &  0.81e-21 &  0.20e-22 &  0.17e-21 &  0.29e-22 &  0.17e-21 &  0.33e-21 &  0.33e-20 &  0.50e-21 &  0.33e-20 \\
\hline
\multicolumn{13}{c}{\textbf{Measure: Error~~~~Grid type: Equidistant}}\\
\hline
& \multicolumn{12}{c}{Network Parameters} \\
& \multicolumn{4}{c}{90} & \multicolumn{4}{c}{180} & \multicolumn{4}{c}{270} \\
\cline{2-5} \cline{6-9} \cline{10-13}
& \multicolumn{2}{c}{Training} & \multicolumn{2}{c}{Test} & \multicolumn{2}{c}{Training}  & \multicolumn{2}{c}{Test} & \multicolumn{2}{c}{Training} & \multicolumn{2}{c}{Test} \\
\cline{2-3} \cline{4-5} \cline{6-7} \cline{8-9} \cline{10-11} \cline{12-13}
$M_{\rm tr}$ & MSE & MXE & MSE & MXE & MSE & MXE & MSE & MXE & MSE & MXE & MSE & MXE \\
\hline
90  &  0.25e-20 &  0.30e-19 &  0.23e-11 &  0.59e-09 & & & & & & & & \\
180 &  0.87e-20 &  0.34e-18 &  0.37e-15 &  0.16e-12 &  0.16e-19 &  0.11e-18 &  0.22e-13 &  0.13e-10 & & & & \\
270 &  0.20e-20 &  0.83e-19 &  0.30e-18 &  0.20e-15 &  0.55e-19 &  0.18e-17 &  0.14e-16 &  0.93e-14 &  0.78e-19 &  0.31e-17 &  0.47e-17 &  0.30e-14 \\
\hline
\multicolumn{13}{c}{\textbf{Measure: Error~~~~Grid type: Chebyshev}}\\
\hline
& \multicolumn{12}{c}{Network Parameters} \\
& \multicolumn{4}{c}{90} & \multicolumn{4}{c}{180} & \multicolumn{4}{c}{270} \\
\cline{2-5} \cline{6-9} \cline{10-13}
& \multicolumn{2}{c}{Training} & \multicolumn{2}{c}{Test} & \multicolumn{2}{c}{Training} & \multicolumn{2}{c}{Test} & \multicolumn{2}{c}{Training} & \multicolumn{2}{c}{Test} \\
\cline{2-3} \cline{4-5} \cline{6-7} \cline{8-9} \cline{10-11} \cline{12-13}
$M_{\rm tr}$ & MSE & MXE & MSE & MXE & MSE & MXE & MSE & MXE & MSE & MXE & MSE & MXE \\
\hline
90  &  0.48e-19 &  0.17e-18 &  0.52e-19 &  0.18e-18 & & & & & & & & \\
180 &  0.48e-19 &  0.19e-18 &  0.56e-19 &  0.19e-18 &  0.44e-19 &  0.18e-18 &  0.53e-19 &  0.19e-18 & & & & \\
270 &  0.43e-19 &  0.14e-18 &  0.41e-19 &  0.14e-18 &  0.23e-19 &  0.65e-19 &  0.25e-19 &  0.66e-19 &  0.90e-19 &  0.40e-18 &  0.89e-19 &  0.40e-18 \\
\hline
\end{tabular}
\caption{Performance measures for all scenarios of Test Problem 2 with Robin conditions.}
\label{tab:Res_Prob_2_R_SNN}
\end{table}



\begin{table}[h!]
\setlength{\tabcolsep}{2pt}
\centering
\scriptsize
\begin{tabular}{@{\extracolsep{3pt}}ccccccccccccc@{}}
\hline
\multicolumn{13}{c}{\textbf{Measure: Deviation~~~~Grid type: Equidistant}}\\
\hline
 & \multicolumn{12}{c}{Network Parameters} \\
 & \multicolumn{4}{c}{90} & \multicolumn{4}{c}{180} & \multicolumn{4}{c}{270} \\
\cline{2-5} \cline{6-9} \cline{10-13}
 & \multicolumn{2}{c}{Training} & \multicolumn{2}{c}{Test} & \multicolumn{2}{c}{Training} & \multicolumn{2}{c}{Test} & \multicolumn{2}{c}{Training} & \multicolumn{2}{c}{Test} \\
\cline{2-3} \cline{4-5} \cline{6-7} \cline{8-9} \cline{10-11} \cline{12-13}
$M_{\rm tr}$ & MSD & MXD & MSD & MXD & MSD & MXD & MSD & MXD & MSD & MXD & MSD & MXD \\
\hline
90  &  0.11e-17 &  0.51e-17 &  0.11e-17 &  0.51e-17 & & & & & & & & \\
180 &  0.20e-17 &  0.93e-17 &  0.20e-17 &  0.93e-17 &  0.39e-18 &  0.18e-17 &  0.39e-18 &  0.18e-17 & & & & \\
270 &  0.47e-17 &  0.21e-16 &  0.47e-17 &  0.21e-16 &  0.38e-18 &  0.18e-17 &  0.38e-18 &  0.18e-17 &  0.93e-19 &  0.43e-18 &  0.93e-19 &  0.43e-18 \\
\hline
\multicolumn{13}{c}{\textbf{Measure: Deviation~~~~Grid type: Chebyshev}}\\
\hline
 & \multicolumn{12}{c}{Network Parameters} \\
 & \multicolumn{4}{c}{90} & \multicolumn{4}{c}{180} & \multicolumn{4}{c}{270} \\
\cline{2-5} \cline{6-9} \cline{10-13}
 & \multicolumn{2}{c}{Training} & \multicolumn{2}{c}{Test} & \multicolumn{2}{c}{Training} & \multicolumn{2}{c}{Test} & \multicolumn{2}{c}{Training} & \multicolumn{2}{c}{Test} \\
\cline{2-3} \cline{4-5} \cline{6-7} \cline{8-9} \cline{10-11} \cline{12-13}
$M_{\rm tr}$ & MSD & MXD & MSD & MXD & MSD & MXD & MSD & MXD & MSD & MXD & MSD & MXD \\
\hline
90  &  0.11e-17 &  0.46e-17 &  0.98e-18 &  0.46e-17 & & & & & & & & \\
180 &  0.73e-18 &  0.32e-17 &  0.67e-18 &  0.32e-17 &  0.37e-18 &  0.16e-17 &  0.34e-18 &  0.16e-17 & & & & \\
270 &  0.25e-18 &  0.11e-17 &  0.24e-18 &  0.11e-17 &  0.76e-20 &  0.33e-19 &  0.66e-20 &  0.33e-19 &  0.28e-19 &  0.13e-18 &  0.27e-19 &  0.13e-18 \\
\hline
\multicolumn{13}{c}{\textbf{Measure: Error~~~~Grid type: Equidistant}}\\
\hline
 & \multicolumn{12}{c}{Network Parameters} \\
 & \multicolumn{4}{c}{90} & \multicolumn{4}{c}{180} & \multicolumn{4}{c}{270} \\
 \cline{2-5} \cline{6-9} \cline{10-13}
  & \multicolumn{2}{c}{Training} & \multicolumn{2}{c}{Test} & \multicolumn{2}{c}{Training}  & \multicolumn{2}{c}{Test} & \multicolumn{2}{c}{Training} & \multicolumn{2}{c}{Test} \\
 \cline{2-3} \cline{4-5} \cline{6-7} \cline{8-9} \cline{10-11} \cline{12-13}
 $M_{\rm tr}$ & MSE & MXE & MSE & MXE & MSE & MXE & MSE & MXE & MSE & MXE & MSE & MXE \\
 \hline
90  &  0.93e-19 &  0.38e-18 &  0.96e-19 &  0.53e-18 & & & & & & & & \\
180 &  0.37e-19 &  0.66e-18 &  0.37e-19 &  0.81e-18 &  0.35e-19 &  0.19e-18 &  0.36e-19 &  0.20e-18 & & & & \\
270 &  0.13e-18 &  0.24e-17 &  0.13e-18 &  0.24e-17 &  0.88e-20 &  0.16e-18 &  0.88e-20 &  0.16e-18 &  0.45e-19 &  0.28e-18 &  0.45e-19 &  0.36e-18 \\
 \hline
 \multicolumn{13}{c}{\textbf{Measure: Error~~~~Grid type: Chebyshev}}\\
 \hline
  & \multicolumn{12}{c}{Network Parameters} \\
  & \multicolumn{4}{c}{90} & \multicolumn{4}{c}{180} & \multicolumn{4}{c}{270} \\
 \cline{2-5} \cline{6-9} \cline{10-13}
  & \multicolumn{2}{c}{Training} & \multicolumn{2}{c}{Test} & \multicolumn{2}{c}{Training} & \multicolumn{2}{c}{Test} & \multicolumn{2}{c}{Training} & \multicolumn{2}{c}{Test} \\
 \cline{2-3} \cline{4-5} \cline{6-7} \cline{8-9} \cline{10-11} \cline{12-13}
 $M_{\rm tr}$ & MSE & MXE & MSE & MXE & MSE & MXE & MSE & MXE & MSE & MXE & MSE & MXE \\
 \hline
90  &  0.22e-18 &  0.75e-18 &  0.22e-18 &  0.83e-18 & & & & & & & & \\
180 &  0.84e-19 &  0.31e-18 &  0.93e-19 &  0.31e-18 &  0.87e-19 &  0.36e-18 &  0.77e-19 &  0.36e-18 & & & & \\
270 &  0.72e-19 &  0.40e-18 &  0.62e-19 &  0.41e-18 &  0.77e-19 &  0.34e-18 &  0.10e-18 &  0.34e-18 &  0.11e-18 &  0.61e-18 &  0.14e-18 &  0.62e-18 \\
 \hline
 \end{tabular}
 \caption{Performance measures for all scenarios of Test Problem 3.}
 \label{tab:Res_Prob_3_SNN}
 \end{table} 



\begin{table}[h!]
\setlength{\tabcolsep}{2pt}
\centering
\scriptsize
\begin{tabular}{@{\extracolsep{3pt}}cccccccccccccc@{}}
\hline
\multicolumn{14}{c}{\textbf{Measure: Deviation~~~~Grid type: Equidistant}}\\
\hline
&& \multicolumn{12}{c}{Network Parameters} \\
&& \multicolumn{4}{c}{60} & \multicolumn{4}{c}{120} & \multicolumn{4}{c}{240} \\
\cline{3-6}\cline{7-10}\cline{11-14}
&& \multicolumn{2}{c}{Training} & \multicolumn{2}{c}{Test} & \multicolumn{2}{c}{Training} & \multicolumn{2}{c}{Test} & \multicolumn{2}{c}{Training} & \multicolumn{2}{c}{Test} \\
\cline{3-4} \cline{5-6} \cline{7-8} \cline{9-10} \cline{11-12} \cline{13-14}
$M_{\rm tr}$& Solution & MSD & MXD & MSD & MXD & MSD & MXD & MSD & MXD & MSD & MXD & MSD & MXD \\
\hline
70 & $\psi_1$& 0.66e-07 &  0.11e-05 &  0.59e-07 &  0.11e-05 & & & & & & & & \\
&$\psi_2$& 0.47e-06 &  0.92e-05 &  0.41e-06 &  0.92e-05 & & & & & & & & \\
130 & $\psi_1$& 0.14e-07 &  0.25e-06 &  0.13e-07 &  0.25e-06 &  0.14e-07 &  0.25e-06 &  0.13e-07 &  0.25e-06 & & & & \\
&$\psi_2$& 0.99e-07 &  0.21e-05 &  0.93e-07 &  0.21e-05 &  0.98e-07 &  0.20e-05 &  0.92e-07 &  0.20e-05 & & & & \\
250 & $\psi_1$& 0.57e-07 &  0.11e-05 &  0.56e-07 &  0.11e-05 &  0.45e-07 &  0.84e-06 &  0.44e-07 &  0.84e-06 &  0.87e-07 &  0.16e-05 &  0.84e-07 &  0.16e-05 \\
&$\psi_2$& 0.40e-06 &  0.87e-05 &  0.39e-06 &  0.87e-05 &  0.32e-06 &  0.68e-05 &  0.31e-06 &  0.68e-05 &  0.61e-06 &  0.13e-04 &  0.59e-06 &  0.13e-04 \\
\hline
\multicolumn{14}{c}{\textbf{Measure: Deviation~~~~Grid type: Chebyshev}}\\
\hline
&& \multicolumn{12}{c}{Network Parameters}\\
&& \multicolumn{4}{c}{60} & \multicolumn{4}{c}{120} & \multicolumn{4}{c}{240} \\
\cline{3-6}\cline{7-10}\cline{11-14}
&& \multicolumn{2}{c}{Training} & \multicolumn{2}{c}{Test} & \multicolumn{2}{c}{Training} & \multicolumn{2}{c}{Test} & \multicolumn{2}{c}{Training} & \multicolumn{2}{c}{Test} \\
\cline{3-4} \cline{5-6} \cline{7-8} \cline{9-10} \cline{11-12} \cline{13-14}
$M_{\rm tr}$& Solution & MSD & MXD & MSD & MXD & MSD & MXD & MSD & MXD & MSD & MXD & MSD & MXD \\
\hline
70 & $\psi_1$& 0.94e-07 &  0.71e-06 &  0.38e-07 &  0.72e-06 & & & & & & & & \\
   & $\psi_2$& 0.71e-06 &  0.58e-05 &  0.26e-06 &  0.58e-05 & & & & & & & & \\
130& $\psi_1$& 0.79e-08 &  0.60e-07 &  0.32e-08 &  0.60e-07 &  0.38e-06 &  0.29e-05 &  0.15e-06 &  0.29e-05 & & & & \\
   & $\psi_2$& 0.59e-07 &  0.49e-06 &  0.22e-07 &  0.49e-06 &  0.29e-05 &  0.24e-04 &  0.11e-05 &  0.24e-04 & & & & \\
250& $\psi_1$& 0.71e-07 &  0.55e-06 &  0.29e-07 &  0.55e-06 &  0.43e-07 &  0.33e-06 &  0.17e-07 &  0.33e-06 &  0.91e-07 &  0.69e-06 &  0.36e-07 &  0.70e-06 \\
&$\psi_2$& 0.54e-06 &  0.44e-05 &  0.20e-06 &  0.44e-05 &  0.32e-06 &  0.27e-05 &  0.12e-06 &  0.27e-05 &  0.68e-06 &  0.57e-05 &  0.26e-06 &  0.57e-05 \\
\hline
\multicolumn{14}{c}{\textbf{Measure: Error~~~~Grid type: Equidistant}}\\
\hline
&& \multicolumn{12}{c}{Network Parameters}\\
&& \multicolumn{4}{c}{60} & \multicolumn{4}{c}{120} & \multicolumn{4}{c}{240} \\
\cline{3-6}\cline{7-10}\cline{11-14}
&& \multicolumn{2}{c}{Training} & \multicolumn{2}{c}{Test} & \multicolumn{2}{c}{Training} & \multicolumn{2}{c}{Test} & \multicolumn{2}{c}{Training} & \multicolumn{2}{c}{Test} \\
\cline{3-4} \cline{5-6} \cline{7-8} \cline{9-10} \cline{11-12} \cline{13-14}
$M_{\rm tr}$& & MSE & MXE & MSE & MXE & MSE & MXE & MSE & MXE & MSE & MXE & MSE & MXE \\
\hline
70 & & 0.15e-19 &  0.97e-18 &  0.17e-19 &  0.35e-17 & & & & & & & & \\
130 & & 0.49e-19 &  0.39e-17 &  0.46e-19 &  0.39e-17 &  0.90e-21 &  0.67e-19 &  0.84e-21 &  0.77e-19 & & & & \\
250 & & 0.11e-19 &  0.15e-17 &  0.10e-19 &  0.15e-17 &  0.11e-20 &  0.30e-18 &  0.11e-20 &  0.30e-18 &  0.40e-21 &  0.66e-19 &  0.37e-21 &  0.66e-19 \\
\hline
\multicolumn{14}{c}{\textbf{Measure: Error~~~~Grid type: Chebyshev}}\\
\hline
&& \multicolumn{12}{c}{Network Parameters}\\
&& \multicolumn{4}{c}{60} & \multicolumn{4}{c}{120} & \multicolumn{4}{c}{240} \\
\cline{3-6}\cline{7-10}\cline{11-14}
&& \multicolumn{2}{c}{Training} & \multicolumn{2}{c}{Test} & \multicolumn{2}{c}{Training} & \multicolumn{2}{c}{Test} & \multicolumn{2}{c}{Training} & \multicolumn{2}{c}{Test} \\
\cline{3-4} \cline{5-6} \cline{7-8} \cline{9-10} \cline{11-12} \cline{13-14}
$M_{\rm tr}$& & MSE & MXE & MSE & MXE & MSE & MXE & MSE & MXE & MSE & MXE & MSE & MXE \\
\hline
70 & & 0.20e-19 &  0.59e-18 &  0.20e-19 &  0.62e-18 & & & & & & & & \\
130 & & 0.61e-20 &  0.18e-18 &  0.70e-20 &  0.18e-18 &  0.16e-20 &  0.56e-19 &  0.17e-20 &  0.59e-19 & & & & \\
250 & & 0.41e-19 &  0.17e-17 &  0.40e-19 &  0.17e-17 &  0.95e-21 &  0.24e-19 &  0.92e-21 &  0.25e-19 &  0.12e-21 &  0.40e-20 &  0.13e-21 &  0.40e-20 \\
\hline
\end{tabular}
\caption{Performance measures for all scenarios of Test Problem 4.}
\label{tab:Res_Prob_4_SNN}
\end{table}



\begin{table}[h!]
\setlength{\tabcolsep}{2pt}
\centering
\scriptsize
\begin{tabular}{@{\extracolsep{3pt}}cccccccccccccc@{}}
\hline
\multicolumn{14}{c}{\textbf{Measure: Deviation~~~~Grid type: Equidistant}}\\
\hline
&& \multicolumn{12}{c}{Network Parameters} \\
&& \multicolumn{4}{c}{60} & \multicolumn{4}{c}{120} & \multicolumn{4}{c}{240} \\
\cline{3-6}\cline{7-10}\cline{11-14}
&& \multicolumn{2}{c}{Training} & \multicolumn{2}{c}{Test} & \multicolumn{2}{c}{Training} & \multicolumn{2}{c}{Test} & \multicolumn{2}{c}{Training} & \multicolumn{2}{c}{Test} \\
\cline{3-4} \cline{5-6} \cline{7-8} \cline{9-10} \cline{11-12} \cline{13-14}
$M_{\rm tr}$& Solution & MSD & MXD & MSD & MXD & MSD & MXD & MSD & MXD & MSD & MXD & MSD & MXD \\
\hline
70 & $\psi_1$& 0.13e-07 &  0.97e-07 &  0.13e-07 &  0.97e-07 & & & & & & & & \\
&$\psi_2$& 0.34e-07 &  0.47e-06 &  0.37e-07 &  0.60e-06 & & & & & & & & \\
130 & $\psi_1$& 0.67e-12 &  0.51e-11 &  0.67e-12 &  0.51e-11 &  0.57e-13 &  0.19e-11 &  0.60e-13 &  0.21e-11 & & & & \\
&$\psi_2$& 0.21e-11 &  0.49e-10 &  0.22e-11 &  0.54e-10 &  0.72e-13 &  0.55e-12 &  0.73e-13 &  0.55e-12 & & & & \\
250 & $\psi_1$& 0.31e-17 &  0.27e-16 &  0.31e-17 &  0.36e-16 &  0.36e-17 &  0.28e-16 &  0.36e-17 &  0.28e-16 &  0.29e-16 &  0.22e-15 &  0.29e-16 &  0.22e-15 \\
&$\psi_2$& 0.14e-16 &  0.79e-15 &  0.14e-16 &  0.80e-15 &  0.99e-17 &  0.20e-15 &  0.10e-16 &  0.20e-15 &  0.11e-15 &  0.42e-14 &  0.11e-15 &  0.43e-14 \\
\hline
\multicolumn{14}{c}{\textbf{Measure: Deviation~~~~Grid type: Chebyshev}}\\
\hline
&& \multicolumn{12}{c}{Network Parameters}\\
&& \multicolumn{4}{c}{60} & \multicolumn{4}{c}{120} & \multicolumn{4}{c}{240} \\
\cline{3-6}\cline{7-10}\cline{11-14}
&& \multicolumn{2}{c}{Training} & \multicolumn{2}{c}{Test} & \multicolumn{2}{c}{Training} & \multicolumn{2}{c}{Test} & \multicolumn{2}{c}{Training} & \multicolumn{2}{c}{Test} \\
\cline{3-4} \cline{5-6} \cline{7-8} \cline{9-10} \cline{11-12} \cline{13-14}
$M_{\rm tr}$& Solution & MSD & MXD & MSD & MXD & MSD & MXD & MSD & MXD & MSD & MXD & MSD & MXD \\
\hline
70 & $\psi_1$& 0.75e-19 &  0.39e-18 &  0.10e-18 &  0.39e-18 & & & & & & & & \\
&$\psi_2$& 0.11e-18 &  0.38e-18 &  0.13e-18 &  0.39e-18 & & & & & & & & \\
130 & $\psi_1$& 0.83e-19 &  0.36e-18 &  0.11e-18 &  0.37e-18 &  0.30e-20 &  0.18e-19 &  0.43e-20 &  0.18e-19 & & & & \\
&$\psi_2$& 0.47e-19 &  0.28e-18 &  0.54e-19 &  0.29e-18 &  0.45e-20 &  0.22e-19 &  0.56e-20 &  0.22e-19 & & & & \\
250 & $\psi_1$& 0.42e-18 &  0.21e-17 &  0.43e-18 &  0.21e-17 &  0.18e-20 &  0.11e-19 &  0.26e-20 &  0.11e-19 &  0.13e-21 &  0.95e-21 &  0.17e-21 &  0.95e-21 \\
&$\psi_2$& 0.83e-18 &  0.38e-17 &  0.88e-18 &  0.38e-17 &  0.58e-20 &  0.36e-19 &  0.82e-20 &  0.36e-19 &  0.48e-21 &  0.30e-20 &  0.67e-21 &  0.30e-20 \\
\hline
\multicolumn{14}{c}{\textbf{Measure: Error~~~~Grid type: Equidistant}}\\
\hline
&& \multicolumn{12}{c}{Network Parameters}\\
&& \multicolumn{4}{c}{60} & \multicolumn{4}{c}{120} & \multicolumn{4}{c}{240} \\
\cline{3-6}\cline{7-10}\cline{11-14}
&& \multicolumn{2}{c}{Training} & \multicolumn{2}{c}{Test} & \multicolumn{2}{c}{Training} & \multicolumn{2}{c}{Test} & \multicolumn{2}{c}{Training} & \multicolumn{2}{c}{Test} \\
\cline{3-4} \cline{5-6} \cline{7-8} \cline{9-10} \cline{11-12} \cline{13-14}
$M_{\rm tr}$& & MSE & MXE & MSE & MXE & MSE & MXE & MSE & MXE & MSE & MXE & MSE & MXE \\
\hline
70 & & 0.98e-17 &  0.58e-16 &  0.58e-05 &  0.18e-02 & & & & & & & & \\
130 & & 0.15e-16 &  0.19e-15 &  0.61e-09 &  0.22e-06 &  0.45e-18 &  0.11e-16 &  0.31e-10 &  0.11e-07 & & & & \\
250 & & 0.60e-16 &  0.28e-14 &  0.15e-13 &  0.96e-11 &  0.52e-18 &  0.27e-16 &  0.33e-14 &  0.22e-11 &  0.11e-18 &  0.28e-17 &  0.77e-13 &  0.54e-10 \\
\hline
\multicolumn{14}{c}{\textbf{Measure: Error~~~~Grid type: Chebyshev}}\\
\hline
&& \multicolumn{12}{c}{Network Parameters}\\
&& \multicolumn{4}{c}{60} & \multicolumn{4}{c}{120} & \multicolumn{4}{c}{240} \\
\cline{3-6}\cline{7-10}\cline{11-14}
&& \multicolumn{2}{c}{Training} & \multicolumn{2}{c}{Test} & \multicolumn{2}{c}{Training} & \multicolumn{2}{c}{Test} & \multicolumn{2}{c}{Training} & \multicolumn{2}{c}{Test} \\
\cline{3-4} \cline{5-6} \cline{7-8} \cline{9-10} \cline{11-12} \cline{13-14}
$M_{\rm tr}$& & MSE & MXE & MSE & MXE & MSE & MXE & MSE & MXE & MSE & MXE & MSE & MXE \\
\hline
70 & & 0.30e-16 &  0.19e-15 &  0.24e-16 &  0.20e-15 & & & & & & & & \\
130 & & 0.29e-16 &  0.13e-15 &  0.27e-16 &  0.13e-15 &  0.16e-17 &  0.52e-17 &  0.17e-17 &  0.49e-17 & & & & \\
250 & & 0.64e-16 &  0.60e-15 &  0.42e-16 &  0.55e-15 &  0.75e-18 &  0.66e-17 &  0.63e-18 &  0.46e-17 &  0.23e-18 &  0.20e-17 &  0.16e-18 &  0.16e-17 \\
\hline
\end{tabular}
\caption{Performance measures for all scenarios of Test Problem 5.}
\label{tab:Res_Prob_5_SNN}
\end{table}



\begin{table}[h!]
\setlength{\tabcolsep}{2pt}
\centering
\scriptsize
\begin{tabular}{@{\extracolsep{3pt}}ccccccccccccc@{}}
\hline
\multicolumn{13}{c}{\textbf{Measure: Deviation~~~~Grid type: Equidistant}}\\
\hline
& \multicolumn{12}{c}{Network Parameters} \\
& \multicolumn{4}{c}{180} & \multicolumn{4}{c}{270} & \multicolumn{4}{c}{360} \\
\cline{2-5} \cline{6-9} \cline{10-13}
& \multicolumn{2}{c}{Training} & \multicolumn{2}{c}{Test} & \multicolumn{2}{c}{Training} & \multicolumn{2}{c}{Test} & \multicolumn{2}{c}{Training} & \multicolumn{2}{c}{Test} \\
\cline{2-3} \cline{4-5} \cline{6-7} \cline{8-9} \cline{10-11} \cline{12-13}
$M_{\rm tr}$ & MSD & MXD & MSD & MXD & MSD & MXD & MSD & MXD & MSD & MXD & MSD & MXD \\
\hline
180 &  0.34e-16 &  0.16e-14 &  0.36e-16 &  0.17e-14 & & & & & & & & \\
270 &  0.19e-21 &  0.96e-20 &  0.19e-21 &  0.95e-20 &  0.70e-20 &  0.44e-19 &  0.70e-20 &  0.43e-19 & & & & \\
360 &  0.87e-20 &  0.37e-18 &  0.88e-20 &  0.37e-18 &  0.29e-11 &  0.14e-09 &  0.30e-11 &  0.16e-09 &  0.57e-20 &  0.33e-18 &  0.57e-20 &  0.33e-18 \\
\hline
\multicolumn{13}{c}{\textbf{Measure: Deviation~~~~Grid type: Chebyshev}}\\
\hline
& \multicolumn{12}{c}{Network Parameters} \\
& \multicolumn{4}{c}{180} & \multicolumn{4}{c}{270} & \multicolumn{4}{c}{360} \\
\cline{2-5} \cline{6-9} \cline{10-13}
& \multicolumn{2}{c}{Training} & \multicolumn{2}{c}{Test} & \multicolumn{2}{c}{Training} & \multicolumn{2}{c}{Test} & \multicolumn{2}{c}{Training} & \multicolumn{2}{c}{Test} \\
\cline{2-3} \cline{4-5} \cline{6-7} \cline{8-9} \cline{10-11} \cline{12-13}
$M_{\rm tr}$ & MSD & MXD & MSD & MXD & MSD & MXD & MSD & MXD & MSD & MXD & MSD & MXD \\
\hline
180 &  0.30e-22 &  0.21e-21 &  0.41e-22 &  0.21e-21 & & & & & & & & \\
270 &  0.24e-22 &  0.14e-21 &  0.32e-22 &  0.14e-21 &  0.22e-22 &  0.12e-21 &  0.29e-22 &  0.12e-21 & & & & \\
360 &  0.29e-22 &  0.14e-21 &  0.39e-22 &  0.14e-21 &  0.26e-21 &  0.15e-20 &  0.34e-21 &  0.15e-20 &  0.11e-21 &  0.76e-21 &  0.16e-21 &  0.76e-21 \\
\hline
\multicolumn{13}{c}{\textbf{Measure: Error~~~~Grid type: Equidistant}}\\
\hline
& \multicolumn{12}{c}{Network Parameters} \\
& \multicolumn{4}{c}{180} & \multicolumn{4}{c}{270} & \multicolumn{4}{c}{360} \\
\cline{2-5} \cline{6-9} \cline{10-13}
& \multicolumn{2}{c}{Training} & \multicolumn{2}{c}{Test} & \multicolumn{2}{c}{Training}  & \multicolumn{2}{c}{Test} & \multicolumn{2}{c}{Training} & \multicolumn{2}{c}{Test} \\
\cline{2-3} \cline{4-5} \cline{6-7} \cline{8-9} \cline{10-11} \cline{12-13}
$M_{\rm tr}$ & MSE & MXE & MSE & MXE & MSE & MXE & MSE & MXE & MSE & MXE & MSE & MXE \\
\hline
180 &  0.21e-18 &  0.33e-17 &  0.89e-14 &  0.61e-11 & & & & & & & & \\
270 &  0.24e-19 &  0.14e-18 &  0.48e-17 &  0.34e-14 &  0.23e-17 &  0.16e-16 &  0.62e-17 &  0.17e-14 & & & & \\
360 &  0.98e-18 &  0.18e-16 &  0.33e-17 &  0.20e-14 &  0.39e-20 &  0.12e-18 &  0.11e-13 &  0.11e-10 &  0.96e-19 &  0.40e-17 &  0.23e-17 &  0.19e-14 \\
\hline
\multicolumn{13}{c}{\textbf{Measure: Error~~~~Grid type: Chebyshev}}\\
\hline
& \multicolumn{12}{c}{Network Parameters} \\
& \multicolumn{4}{c}{180} & \multicolumn{4}{c}{270} & \multicolumn{4}{c}{360} \\
\cline{2-5} \cline{6-9} \cline{10-13}
& \multicolumn{2}{c}{Training} & \multicolumn{2}{c}{Test} & \multicolumn{2}{c}{Training} & \multicolumn{2}{c}{Test} & \multicolumn{2}{c}{Training} & \multicolumn{2}{c}{Test} \\
\cline{2-3} \cline{4-5} \cline{6-7} \cline{8-9} \cline{10-11} \cline{12-13}
$M_{\rm tr}$ & MSE & MXE & MSE & MXE & MSE & MXE & MSE & MXE & MSE & MXE & MSE & MXE \\
\hline
180 &  0.21e-20 &  0.12e-19 &  0.18e-20 &  0.12e-19 & & & & & & & & \\
270 &  0.23e-20 &  0.13e-19 &  0.18e-20 &  0.13e-19 &  0.48e-20 &  0.20e-19 &  0.42e-20 &  0.22e-19 & & & & \\
360 &  0.23e-20 &  0.87e-20 &  0.22e-20 &  0.87e-20 &  0.17e-19 &  0.64e-19 &  0.16e-19 &  0.62e-19 &  0.18e-19 &  0.11e-18 &  0.17e-19 &  0.12e-18 \\
\hline
\end{tabular}
\caption{Performance measures for all scenarios of Test Problem 6 with Dirichlet conditions.}
\label{tab:Res_Prob_6_DD_SNN}
\end{table}

\begin{table}[h!]
\setlength{\tabcolsep}{2pt}
\centering
\scriptsize
\begin{tabular}{@{\extracolsep{3pt}}ccccccccccccc@{}}
\hline
\multicolumn{13}{c}{\textbf{Measure: Deviation~~~~Grid type: Equidistant}}\\
\hline
& \multicolumn{12}{c}{Network Parameters} \\
& \multicolumn{4}{c}{180} & \multicolumn{4}{c}{270} & \multicolumn{4}{c}{360} \\
\cline{2-5} \cline{6-9} \cline{10-13}
& \multicolumn{2}{c}{Training} & \multicolumn{2}{c}{Test} & \multicolumn{2}{c}{Training} & \multicolumn{2}{c}{Test} & \multicolumn{2}{c}{Training} & \multicolumn{2}{c}{Test} \\
\cline{2-3} \cline{4-5} \cline{6-7} \cline{8-9} \cline{10-11} \cline{12-13}
$M_{\rm tr}$ & MSD & MXD & MSD & MXD & MSD & MXD & MSD & MXD & MSD & MXD & MSD & MXD \\
\hline
180 &  0.25e-19 &  0.13e-17 &  0.23e-19 &  0.13e-17 & & & & & & & & \\
270 &  0.96e-21 &  0.16e-19 &  0.96e-21 &  0.16e-19 &  0.78e-21 &  0.90e-20 &  0.77e-21 &  0.90e-20 & & & & \\
360 &  0.52e-21 &  0.38e-20 &  0.51e-21 &  0.38e-20 &  0.30e-21 &  0.14e-20 &  0.30e-21 &  0.14e-20 &  0.15e-21 &  0.46e-20 &  0.15e-21 &  0.46e-20 \\
\hline
\multicolumn{13}{c}{\textbf{Measure: Deviation~~~~Grid type: Chebyshev}}\\
\hline
& \multicolumn{12}{c}{Network Parameters} \\
& \multicolumn{4}{c}{180} & \multicolumn{4}{c}{270} & \multicolumn{4}{c}{360} \\
\cline{2-5} \cline{6-9} \cline{10-13}
& \multicolumn{2}{c}{Training} & \multicolumn{2}{c}{Test} & \multicolumn{2}{c}{Training} & \multicolumn{2}{c}{Test} & \multicolumn{2}{c}{Training} & \multicolumn{2}{c}{Test} \\
\cline{2-3} \cline{4-5} \cline{6-7} \cline{8-9} \cline{10-11} \cline{12-13}
$M_{\rm tr}$ & MSD & MXD & MSD & MXD & MSD & MXD & MSD & MXD & MSD & MXD & MSD & MXD \\
\hline
180 &  0.30e-20 &  0.20e-19 &  0.42e-20 &  0.20e-19 & & & & & & & & \\
270 &  0.69e-21 &  0.37e-20 &  0.94e-21 &  0.37e-20 &  0.10e-20 &  0.53e-20 &  0.13e-20 &  0.53e-20 & & & & \\
360 &  0.60e-21 &  0.37e-20 &  0.81e-21 &  0.37e-20 &  0.46e-21 &  0.23e-20 &  0.57e-21 &  0.23e-20 &  0.20e-21 &  0.13e-20 &  0.29e-21 &  0.13e-20 \\
\hline
\multicolumn{13}{c}{\textbf{Measure: Error~~~~Grid type: Equidistant}}\\
\hline
& \multicolumn{12}{c}{Network Parameters} \\
& \multicolumn{4}{c}{180} & \multicolumn{4}{c}{270} & \multicolumn{4}{c}{360} \\
\cline{2-5} \cline{6-9} \cline{10-13}
& \multicolumn{2}{c}{Training} & \multicolumn{2}{c}{Test} & \multicolumn{2}{c}{Training}  & \multicolumn{2}{c}{Test} & \multicolumn{2}{c}{Training} & \multicolumn{2}{c}{Test} \\
\cline{2-3} \cline{4-5} \cline{6-7} \cline{8-9} \cline{10-11} \cline{12-13}
$M_{\rm tr}$ & MSE & MXE & MSE & MXE & MSE & MXE & MSE & MXE & MSE & MXE & MSE & MXE \\
\hline
180 &  0.34e-19 &  0.67e-18 &  0.65e-16 &  0.28e-13 & & & & & & & & \\
270 &  0.54e-19 &  0.89e-18 &  0.13e-17 &  0.82e-15 &  0.49e-19 &  0.39e-18 &  0.16e-18 &  0.66e-16 & & & & \\
360 &  0.49e-19 &  0.17e-17 &  0.16e-18 &  0.87e-16 &  0.24e-19 &  0.70e-18 &  0.90e-19 &  0.56e-16 &  0.28e-19 &  0.78e-18 &  0.70e-19 &  0.33e-16 \\
\hline
\multicolumn{13}{c}{\textbf{Measure: Error~~~~Grid type: Chebyshev}}\\
\hline
& \multicolumn{12}{c}{Network Parameters} \\
& \multicolumn{4}{c}{180} & \multicolumn{4}{c}{270} & \multicolumn{4}{c}{360} \\
\cline{2-5} \cline{6-9} \cline{10-13}
& \multicolumn{2}{c}{Training} & \multicolumn{2}{c}{Test} & \multicolumn{2}{c}{Training} & \multicolumn{2}{c}{Test} & \multicolumn{2}{c}{Training} & \multicolumn{2}{c}{Test} \\
\cline{2-3} \cline{4-5} \cline{6-7} \cline{8-9} \cline{10-11} \cline{12-13}
$M_{\rm tr}$ & MSE & MXE & MSE & MXE & MSE & MXE & MSE & MXE & MSE & MXE & MSE & MXE \\
\hline
180 &  0.17e-18 &  0.68e-18 &  0.17e-18 &  0.69e-18 & & & & & & & & \\
270 &  0.38e-19 &  0.33e-18 &  0.29e-19 &  0.36e-18 &  0.70e-19 &  0.30e-18 &  0.64e-19 &  0.32e-18 & & & & \\
360 &  0.87e-19 &  0.58e-18 &  0.69e-19 &  0.62e-18 &  0.59e-19 &  0.21e-18 &  0.59e-19 &  0.21e-18 &  0.19e-19 &  0.74e-19 &  0.18e-19 &  0.75e-19 \\
\hline
\end{tabular}
\caption{Performance measures for all scenarios of Test Problem 6 with Dirichlet-Neumann conditions.}
\label{tab:Res_Prob_6_DN_SNN}
\end{table}


\begin{table}[h!]
\setlength{\tabcolsep}{2pt}
\centering
\scriptsize
\begin{tabular}{@{\extracolsep{3pt}}ccccccccccccc@{}}
\hline
\multicolumn{13}{c}{\textbf{Measure: Deviation~~~~Grid type: Equidistant}}\\
\hline
& \multicolumn{12}{c}{Network Parameters} \\
& \multicolumn{4}{c}{180} & \multicolumn{4}{c}{270} & \multicolumn{4}{c}{360} \\
\cline{2-5} \cline{6-9} \cline{10-13}
& \multicolumn{2}{c}{Training} & \multicolumn{2}{c}{Test} & \multicolumn{2}{c}{Training} & \multicolumn{2}{c}{Test} & \multicolumn{2}{c}{Training} & \multicolumn{2}{c}{Test} \\
\cline{2-3} \cline{4-5} \cline{6-7} \cline{8-9} \cline{10-11} \cline{12-13}
$M_{\rm tr}$ & MSD & MXD & MSD & MXD & MSD & MXD & MSD & MXD & MSD & MXD & MSD & MXD \\
\hline
180 &  0.13e-18 &  0.11e-16 &  0.10e-18 &  0.11e-16 & & & & & & & & \\
270 &  0.25e-17 &  0.13e-15 &  0.23e-17 &  0.13e-15 &  0.13e-18 &  0.34e-16 &  0.40e-19 &  0.34e-16 & & & & \\
360 &  0.10e-18 &  0.57e-17 &  0.98e-19 &  0.57e-17 &  0.24e-18 &  0.13e-16 &  0.23e-18 &  0.13e-16 &  0.11e-18 &  0.57e-17 &  0.11e-18 &  0.57e-17 \\
\hline
\multicolumn{13}{c}{\textbf{Measure: Deviation~~~~Grid type: Chebyshev}}\\
\hline
& \multicolumn{12}{c}{Network Parameters} \\
& \multicolumn{4}{c}{180} & \multicolumn{4}{c}{270} & \multicolumn{4}{c}{360} \\
\cline{2-5} \cline{6-9} \cline{10-13}
& \multicolumn{2}{c}{Training} & \multicolumn{2}{c}{Test} & \multicolumn{2}{c}{Training} & \multicolumn{2}{c}{Test} & \multicolumn{2}{c}{Training} & \multicolumn{2}{c}{Test} \\
\cline{2-3} \cline{4-5} \cline{6-7} \cline{8-9} \cline{10-11} \cline{12-13}
$M_{\rm tr}$ & MSD & MXD & MSD & MXD & MSD & MXD & MSD & MXD & MSD & MXD & MSD & MXD \\
\hline
180 &  0.63e-17 &  0.88e-16 &  0.16e-17 &  0.88e-16 & & & & & & & & \\
270 &  0.24e-19 &  0.32e-18 &  0.75e-20 &  0.32e-18 &  0.21e-17 &  0.29e-16 &  0.53e-18 &  0.29e-16 & & & & \\
360 &  0.94e-17 &  0.13e-15 &  0.24e-17 &  0.13e-15 &  0.88e-19 &  0.12e-17 &  0.23e-19 &  0.12e-17 &  0.72e-19 &  0.99e-18 &  0.18e-19 &  0.99e-18 \\
\hline
\multicolumn{13}{c}{\textbf{Measure: Error~~~~Grid type: Equidistant}}\\
\hline
& \multicolumn{12}{c}{Network Parameters} \\
& \multicolumn{4}{c}{180} & \multicolumn{4}{c}{270} & \multicolumn{4}{c}{360} \\
\cline{2-5} \cline{6-9} \cline{10-13}
& \multicolumn{2}{c}{Training} & \multicolumn{2}{c}{Test} & \multicolumn{2}{c}{Training}  & \multicolumn{2}{c}{Test} & \multicolumn{2}{c}{Training} & \multicolumn{2}{c}{Test} \\
\cline{2-3} \cline{4-5} \cline{6-7} \cline{8-9} \cline{10-11} \cline{12-13}
$M_{\rm tr}$ & MSE & MXE & MSE & MXE & MSE & MXE & MSE & MXE & MSE & MXE & MSE & MXE \\
\hline
180 &  0.17e-19 &  0.17e-18 &  0.51e-15 &  0.24e-12 & & & & & & & & \\
270 &  0.41e-19 &  0.74e-18 &  0.54e-17 &  0.37e-14 &  0.68e-20 &  0.15e-18 &  0.16e-13 &  0.16e-10 & & & & \\
360 &  0.36e-19 &  0.69e-18 &  0.69e-19 &  0.25e-16 &  0.22e-19 &  0.33e-18 &  0.16e-18 &  0.12e-15 &  0.85e-20 &  0.34e-18 &  0.36e-19 &  0.21e-16 \\
\hline
\multicolumn{13}{c}{\textbf{Measure: Error~~~~Grid type: Chebyshev}}\\
\hline
& \multicolumn{12}{c}{Network Parameters} \\
& \multicolumn{4}{c}{180} & \multicolumn{4}{c}{270} & \multicolumn{4}{c}{360} \\
\cline{2-5} \cline{6-9} \cline{10-13}
& \multicolumn{2}{c}{Training} & \multicolumn{2}{c}{Test} & \multicolumn{2}{c}{Training} & \multicolumn{2}{c}{Test} & \multicolumn{2}{c}{Training} & \multicolumn{2}{c}{Test} \\
\cline{2-3} \cline{4-5} \cline{6-7} \cline{8-9} \cline{10-11} \cline{12-13}
$M_{\rm tr}$ & MSE & MXE & MSE & MXE & MSE & MXE & MSE & MXE & MSE & MXE & MSE & MXE \\
\hline
180 &  0.54e-19 &  0.15e-18 &  0.57e-19 &  0.15e-18 & & & & & & & & \\
270 &  0.66e-19 &  0.32e-18 &  0.57e-19 &  0.34e-18 &  0.39e-19 &  0.14e-18 &  0.39e-19 &  0.14e-18 & & & & \\
360 &  0.40e-19 &  0.14e-18 &  0.39e-19 &  0.15e-18 &  0.31e-19 &  0.12e-18 &  0.32e-19 &  0.12e-18 &  0.15e-19 &  0.72e-19 &  0.12e-19 &  0.71e-19 \\
\hline
\end{tabular}
\caption{Performance measures for all scenarios of Test Problem 6 with Cauchy conditions.}
\label{tab:Res_Prob_6_C_SNN}
\end{table}

\begin{table}[h!]
\setlength{\tabcolsep}{2pt}
\centering
\scriptsize
\begin{tabular}{@{\extracolsep{3pt}}ccccccccccccc@{}}
\hline
\multicolumn{13}{c}{\textbf{Measure: Deviation~~~~Grid type: Equidistant}}\\
\hline
& \multicolumn{12}{c}{Network Parameters} \\
& \multicolumn{4}{c}{180} & \multicolumn{4}{c}{270} & \multicolumn{4}{c}{360} \\
\cline{2-5} \cline{6-9} \cline{10-13}
& \multicolumn{2}{c}{Training} & \multicolumn{2}{c}{Test} & \multicolumn{2}{c}{Training} & \multicolumn{2}{c}{Test} & \multicolumn{2}{c}{Training} & \multicolumn{2}{c}{Test} \\
\cline{2-3} \cline{4-5} \cline{6-7} \cline{8-9} \cline{10-11} \cline{12-13}
$M_{\rm tr}$ & MSD & MXD & MSD & MXD & MSD & MXD & MSD & MXD & MSD & MXD & MSD & MXD \\
\hline
180 &  0.50e-19 &  0.32e-17 &  0.34e-19 &  0.32e-17 & & & & & & & & \\
270 &  0.10e-20 &  0.30e-19 &  0.10e-20 &  0.30e-19 &  0.20e-20 &  0.17e-19 &  0.20e-20 &  0.17e-19 & & & & \\
360 &  0.73e-21 &  0.36e-20 &  0.73e-21 &  0.37e-20 &  0.40e-21 &  0.54e-20 &  0.39e-21 &  0.54e-20 &  0.55e-21 &  0.22e-20 &  0.55e-21 &  0.22e-20 \\
\hline
\multicolumn{13}{c}{\textbf{Measure: Deviation~~~~Grid type: Chebyshev}}\\
\hline
& \multicolumn{12}{c}{Network Parameters} \\
& \multicolumn{4}{c}{180} & \multicolumn{4}{c}{270} & \multicolumn{4}{c}{360} \\
\cline{2-5} \cline{6-9} \cline{10-13}
& \multicolumn{2}{c}{Training} & \multicolumn{2}{c}{Test} & \multicolumn{2}{c}{Training} & \multicolumn{2}{c}{Test} & \multicolumn{2}{c}{Training} & \multicolumn{2}{c}{Test} \\
\cline{2-3} \cline{4-5} \cline{6-7} \cline{8-9} \cline{10-11} \cline{12-13}
$M_{\rm tr}$ & MSD & MXD & MSD & MXD & MSD & MXD & MSD & MXD & MSD & MXD & MSD & MXD \\
\hline
180 &  0.14e-20 &  0.10e-19 &  0.17e-20 &  0.10e-19 & & & & & & & & \\
270 &  0.31e-20 &  0.16e-19 &  0.30e-20 &  0.16e-19 &  0.13e-21 &  0.94e-21 &  0.17e-21 &  0.94e-21 & & & & \\
360 &  0.42e-21 &  0.20e-20 &  0.50e-21 &  0.20e-20 &  0.17e-21 &  0.12e-20 &  0.18e-21 &  0.12e-20 &  0.22e-21 &  0.12e-20 &  0.29e-21 &  0.12e-20 \\
\hline
\multicolumn{13}{c}{\textbf{Measure: Error~~~~Grid type: Equidistant}}\\
\hline
& \multicolumn{12}{c}{Network Parameters} \\
& \multicolumn{4}{c}{180} & \multicolumn{4}{c}{270} & \multicolumn{4}{c}{360} \\
\cline{2-5} \cline{6-9} \cline{10-13}
& \multicolumn{2}{c}{Training} & \multicolumn{2}{c}{Test} & \multicolumn{2}{c}{Training}  & \multicolumn{2}{c}{Test} & \multicolumn{2}{c}{Training} & \multicolumn{2}{c}{Test} \\
\cline{2-3} \cline{4-5} \cline{6-7} \cline{8-9} \cline{10-11} \cline{12-13}
$M_{\rm tr}$ & MSE & MXE & MSE & MXE & MSE & MXE & MSE & MXE & MSE & MXE & MSE & MXE \\
\hline
180 &  0.47e-19 &  0.11e-17 &  0.95e-16 &  0.39e-13 & & & & & & & & \\
270 &  0.35e-19 &  0.38e-18 &  0.23e-17 &  0.16e-14 &  0.47e-19 &  0.10e-17 &  0.12e-17 &  0.72e-15 & & & & \\
360 &  0.51e-19 &  0.15e-17 &  0.12e-18 &  0.50e-16 &  0.29e-19 &  0.98e-18 &  0.74e-19 &  0.34e-16 &  0.16e-19 &  0.19e-18 &  0.24e-19 &  0.62e-17 \\
\hline
\multicolumn{13}{c}{\textbf{Measure: Error~~~~Grid type: Chebyshev}}\\
\hline
& \multicolumn{12}{c}{Network Parameters} \\
& \multicolumn{4}{c}{180} & \multicolumn{4}{c}{270} & \multicolumn{4}{c}{360} \\
\cline{2-5} \cline{6-9} \cline{10-13}
& \multicolumn{2}{c}{Training} & \multicolumn{2}{c}{Test} & \multicolumn{2}{c}{Training} & \multicolumn{2}{c}{Test} & \multicolumn{2}{c}{Training} & \multicolumn{2}{c}{Test} \\
\cline{2-3} \cline{4-5} \cline{6-7} \cline{8-9} \cline{10-11} \cline{12-13}
$M_{\rm tr}$ & MSE & MXE & MSE & MXE & MSE & MXE & MSE & MXE & MSE & MXE & MSE & MXE \\
\hline
180 &  0.58e-19 &  0.29e-18 &  0.49e-19 &  0.37e-18 & & & & & & & & \\
270 &  0.51e-19 &  0.13e-18 &  0.59e-19 &  0.13e-18 &  0.13e-19 &  0.42e-19 &  0.13e-19 &  0.41e-19 & & & & \\
360 &  0.10e-18 &  0.94e-18 &  0.73e-19 &  0.98e-18 &  0.31e-19 &  0.39e-18 &  0.16e-19 &  0.42e-18 &  0.35e-19 &  0.30e-18 &  0.26e-19 &  0.32e-18 \\
\hline
\end{tabular}
\caption{Performance measures for all scenarios of Test Problem 6 with Neumann conditions.}
\label{tab:Res_Prob_6_NN_SNN}
\end{table}




\clearpage

\section{Results for the proposed augmented neural forms}
\label{appendix:NF}


\begin{table}[h!]
\setlength{\tabcolsep}{2pt}
\centering 
\scriptsize
\begin{tabular}{@{\extracolsep{3pt}}ccccccccccccc@{}}
\hline
\multicolumn{13}{c}{\textbf{Measure: Deviation~~~~Grid type: Equidistant}}\\
\hline
  & \multicolumn{12}{c}{Network Parameters} \\
  & \multicolumn{4}{c}{36} & \multicolumn{4}{c}{72} & \multicolumn{4}{c}{144} \\
\cline{2-5} \cline{6-9} \cline{10-13}
  & \multicolumn{2}{c}{Training} & \multicolumn{2}{c}{Test} & \multicolumn{2}{c}{Training} & \multicolumn{2}{c}{Test} & \multicolumn{2}{c}{Training} & \multicolumn{2}{c}{Test} \\
\cline{2-3} \cline{4-5} \cline{6-7} \cline{8-9} \cline{10-11} \cline{12-13}
$M_{\rm tr}$ & MSD & MXD & MSD & MXD & MSD & MXD & MSD & MXD & MSD & MXD & MSD & MXD \\
\hline
40 & 0.20e-10 &  0.79e-09 &  0.62e-10 &  0.41e-08 & & & & & & & & \\
80 & 0.64e-14 &  0.45e-12 &  0.80e-14 &  0.68e-12 &  0.74e-11 &  0.50e-09 &  0.10e-10 &  0.88e-09 & & & & \\
160 & 0.35e-18 &  0.36e-16 &  0.23e-18 &  0.36e-16 &  0.81e-18 &  0.60e-16 &  0.77e-18 &  0.73e-16 &  0.44e-17 &  0.43e-15 &  0.47e-17 &  0.49e-15 \\
\hline
\multicolumn{13}{c}{\textbf{Measure: Deviation~~~~Grid type: Chebyshev}}\\
\hline
 & \multicolumn{12}{c}{Network Parameters} \\
 & \multicolumn{4}{c}{36} & \multicolumn{4}{c}{72} & \multicolumn{4}{c}{144} \\
\cline{2-5}\cline{6-9}\cline{10-13}
 & \multicolumn{2}{c}{Training} & \multicolumn{2}{c}{Test} & \multicolumn{2}{c}{Training} & \multicolumn{2}{c}{Test}  & \multicolumn{2}{c}{Training} & \multicolumn{2}{c}{Test} \\
\cline{2-3} \cline{4-5} \cline{6-7} \cline{8-9} \cline{10-11} \cline{12-13}
$M_{\rm tr}$ & MSD & MXD & MSD & MXD & MSD & MXD & MSD & MXD & MSD & MXD & MSD & MXD \\
\hline
40 & 0.38e-17 &  0.36e-16 &  0.27e-18 &  0.36e-16 & & & & & & & & \\
80 & 0.27e-17 &  0.36e-16 &  0.26e-18 &  0.36e-16 &  0.27e-17 &  0.36e-16 &  0.26e-18 &  0.36e-16 & & & & \\
160 & 0.22e-17 &  0.36e-16 &  0.26e-18 &  0.36e-16 &  0.22e-17 &  0.36e-16 &  0.26e-18 &  0.36e-16 &  0.22e-17 &  0.36e-16 &  0.25e-18 &  0.36e-16 \\
\hline
\multicolumn{13}{c}{\textbf{Measure: Error~~~~Grid type: Equidistant}}\\
\hline
 & \multicolumn{12}{c}{Network Parameters} \\
 & \multicolumn{4}{c}{36} & \multicolumn{4}{c}{72} & \multicolumn{4}{c}{144} \\
\cline{2-5}\cline{6-9}\cline{10-13}
 & \multicolumn{2}{c}{Training} & \multicolumn{2}{c}{Test} & \multicolumn{2}{c}{Training} & \multicolumn{2}{c}{Test}  & \multicolumn{2}{c}{Training} & \multicolumn{2}{c}{Test} \\
\cline{2-3} \cline{4-5} \cline{6-7} \cline{8-9} \cline{10-11} \cline{12-13}
$M_{\rm tr}$ & MSE & MXE & MSE & MXE & MSE & MXE & MSE & MXE & MSE & MXE & MSE & MXE \\
\hline
40 & 0.21e-17 &  0.89e-17 &  0.44e-06 &  0.69e-04 & & & & & & & & \\
80 & 0.95e-17 &  0.10e-15 &  0.79e-10 &  0.17e-07 &  0.83e-18 &  0.48e-17 &  0.11e-06 &  0.31e-04 & & & & \\
160 & 0.40e-17 &  0.50e-16 &  0.17e-16 &  0.45e-14 &  0.13e-17 &  0.11e-16 &  0.29e-14 &  0.11e-11 &  0.22e-18 &  0.15e-17 &  0.53e-13 &  0.20e-10 \\
\hline
\multicolumn{13}{c}{\textbf{Measure: Error~~~~Grid type: Chebyshev}}\\
\hline
 & \multicolumn{12}{c}{Network Parameters} \\
 & \multicolumn{4}{c}{36} & \multicolumn{4}{c}{72} & \multicolumn{4}{c}{144} \\
\cline{2-5}\cline{6-9}\cline{10-13}
 & \multicolumn{2}{c}{Training} & \multicolumn{2}{c}{Test} & \multicolumn{2}{c}{Training} & \multicolumn{2}{c}{Test}  & \multicolumn{2}{c}{Training} & \multicolumn{2}{c}{Test} \\
\cline{2-3} \cline{4-5} \cline{6-7} \cline{8-9} \cline{10-11} \cline{12-13}
$M_{\rm tr}$ & MSE & MXE & MSE & MXE & MSE & MXE & MSE & MXE & MSE & MXE & MSE & MXE \\
\hline
40 & 0.13e-18 &  0.17e-17 &  0.29e-14 &  0.42e-12 & & & & & & & & \\
80 & 0.18e-16 &  0.50e-16 &  0.21e-16 &  0.51e-16 &  0.76e-17 &  0.29e-16 &  0.88e-17 &  0.29e-16 & & & & \\
160 & 0.54e-17 &  0.25e-16 &  0.50e-17 &  0.27e-16 &  0.30e-17 &  0.17e-16 &  0.25e-17 &  0.18e-16 &  0.48e-18 &  0.45e-17 &  0.30e-18 &  0.45e-17 \\
\hline
\end{tabular}
\caption{Performance measures for all scenarios of Test Problem 1.}
\label{tab:Res_Prob_1}
 \end{table} 



\begin{table}[h!]
\setlength{\tabcolsep}{2pt}
\centering
\scriptsize
\begin{tabular}{@{\extracolsep{3pt}}ccccccccccccc@{}}
\hline
\multicolumn{13}{c}{\textbf{Measure: Deviation~~~~Grid type: Equidistant}}\\
\hline
& \multicolumn{12}{c}{Network Parameters} \\
& \multicolumn{4}{c}{90} & \multicolumn{4}{c}{180} & \multicolumn{4}{c}{270} \\
\cline{2-5} \cline{6-9} \cline{10-13}
& \multicolumn{2}{c}{Training} & \multicolumn{2}{c}{Test} & \multicolumn{2}{c}{Training} & \multicolumn{2}{c}{Test} & \multicolumn{2}{c}{Training} & \multicolumn{2}{c}{Test} \\
\cline{2-3} \cline{4-5} \cline{6-7} \cline{8-9} \cline{10-11} \cline{12-13}
$M_{\rm tr}$ & MSD & MXD & MSD & MXD & MSD & MXD & MSD & MXD & MSD & MXD & MSD & MXD \\
\hline
90  &  0.48e-15 &  0.19e-13 &  0.56e-15 &  0.25e-13 & & & & & & & & \\
180 &  0.24e-20 &  0.12e-18 &  0.26e-20 &  0.15e-18 &  0.19e-18 &  0.93e-17 &  0.21e-18 &  0.11e-16 & & & & \\
270 &  0.78e-22 &  0.56e-21 &  0.78e-22 &  0.56e-21 &  0.11e-20 &  0.20e-19 &  0.11e-20 &  0.25e-19 &  0.21e-19 &  0.98e-19 &  0.22e-19 &  0.11e-18 \\
\hline
\multicolumn{13}{c}{\textbf{Measure: Deviation~~~~Grid type: Chebyshev}}\\
\hline
& \multicolumn{12}{c}{Network Parameters} \\
& \multicolumn{4}{c}{90} & \multicolumn{4}{c}{180} & \multicolumn{4}{c}{270} \\
\cline{2-5} \cline{6-9} \cline{10-13}
& \multicolumn{2}{c}{Training} & \multicolumn{2}{c}{Test} & \multicolumn{2}{c}{Training} & \multicolumn{2}{c}{Test} & \multicolumn{2}{c}{Training} & \multicolumn{2}{c}{Test} \\
\cline{2-3} \cline{4-5} \cline{6-7} \cline{8-9} \cline{10-11} \cline{12-13}
$M_{\rm tr}$ & MSD & MXD & MSD & MXD & MSD & MXD & MSD & MXD & MSD & MXD & MSD & MXD \\
\hline
90  &  0.21e-21 &  0.24e-20 &  0.31e-21 &  0.24e-20 & & & & & & & & \\
180 &  0.82e-21 &  0.60e-20 &  0.11e-20 &  0.62e-20 &  0.12e-21 &  0.13e-20 &  0.17e-21 &  0.13e-20 & & & & \\
270 &  0.82e-23 &  0.98e-22 &  0.12e-22 &  0.98e-22 &  0.43e-21 &  0.50e-20 &  0.63e-21 &  0.50e-20 &  0.26e-20 &  0.37e-19 &  0.39e-20 &  0.37e-19 \\
\hline
\multicolumn{13}{c}{\textbf{Measure: Error~~~~Grid type: Equidistant}}\\
\hline
& \multicolumn{12}{c}{Network Parameters} \\
& \multicolumn{4}{c}{90} & \multicolumn{4}{c}{180} & \multicolumn{4}{c}{270} \\
\cline{2-5} \cline{6-9} \cline{10-13}
& \multicolumn{2}{c}{Training} & \multicolumn{2}{c}{Test} & \multicolumn{2}{c}{Training}  & \multicolumn{2}{c}{Test} & \multicolumn{2}{c}{Training} & \multicolumn{2}{c}{Test} \\
\cline{2-3} \cline{4-5} \cline{6-7} \cline{8-9} \cline{10-11} \cline{12-13}
$M_{\rm tr}$ & MSE & MXE & MSE & MXE & MSE & MXE & MSE & MXE & MSE & MXE & MSE & MXE \\
\hline
90  &  0.86e-19 &  0.73e-18 &  0.53e-10 &  0.13e-07 & & & & & & & & \\
180 &  0.19e-18 &  0.82e-17 &  0.19e-14 &  0.79e-12 &  0.23e-18 &  0.27e-17 &  0.15e-12 &  0.66e-10 & & & & \\
270 &  0.31e-19 &  0.54e-18 &  0.52e-18 &  0.31e-15 &  0.32e-17 &  0.47e-16 &  0.97e-15 &  0.66e-12 &  0.65e-17 &  0.33e-16 &  0.47e-14 &  0.25e-11 \\
\hline
\multicolumn{13}{c}{\textbf{Measure: Error~~~~Grid type: Chebyshev}}\\
\hline
& \multicolumn{12}{c}{Network Parameters} \\
& \multicolumn{4}{c}{90} & \multicolumn{4}{c}{180} & \multicolumn{4}{c}{270} \\
\cline{2-5} \cline{6-9} \cline{10-13}
& \multicolumn{2}{c}{Training} & \multicolumn{2}{c}{Test} & \multicolumn{2}{c}{Training} & \multicolumn{2}{c}{Test} & \multicolumn{2}{c}{Training} & \multicolumn{2}{c}{Test} \\
\cline{2-3} \cline{4-5} \cline{6-7} \cline{8-9} \cline{10-11} \cline{12-13}
$M_{\rm tr}$ & MSE & MXE & MSE & MXE & MSE & MXE & MSE & MXE & MSE & MXE & MSE & MXE \\
\hline
90  &  0.58e-18 &  0.20e-17 &  0.65e-18 &  0.21e-17 & & & & & & & & \\
180 &  0.19e-17 &  0.12e-16 &  0.15e-17 &  0.14e-16 &  0.35e-17 &  0.21e-16 &  0.29e-17 &  0.21e-16 & & & & \\
270 &  0.56e-19 &  0.33e-18 &  0.34e-19 &  0.33e-18 &  0.23e-17 &  0.19e-16 &  0.17e-17 &  0.21e-16 &  0.26e-16 &  0.14e-15 &  0.18e-16 &  0.14e-15 \\
\hline
\end{tabular}
\caption{Performance measures for all scenarios of Test Problem 2 with Dirichlet conditions.}
\label{tab:Res_Prob_2_DD}
\end{table}

\begin{table}[h!]
\setlength{\tabcolsep}{2pt}
\centering
\scriptsize
\begin{tabular}{@{\extracolsep{3pt}}ccccccccccccc@{}}
\hline
\multicolumn{13}{c}{\textbf{Measure: Deviation~~~~Grid type: Equidistant}}\\
\hline
& \multicolumn{12}{c}{Network Parameters} \\
& \multicolumn{4}{c}{90} & \multicolumn{4}{c}{180} & \multicolumn{4}{c}{270} \\
\cline{2-5} \cline{6-9} \cline{10-13}
& \multicolumn{2}{c}{Training} & \multicolumn{2}{c}{Test} & \multicolumn{2}{c}{Training} & \multicolumn{2}{c}{Test} & \multicolumn{2}{c}{Training} & \multicolumn{2}{c}{Test} \\
\cline{2-3} \cline{4-5} \cline{6-7} \cline{8-9} \cline{10-11} \cline{12-13}
$M_{\rm tr}$ & MSD & MXD & MSD & MXD & MSD & MXD & MSD & MXD & MSD & MXD & MSD & MXD \\
\hline
90  &  0.24e-13 &  0.86e-12 &  0.20e-13 &  0.86e-12 & & & & & & & & \\
180 &  0.51e-18 &  0.23e-16 &  0.46e-18 &  0.23e-16 &  0.61e-16 &  0.27e-14 &  0.55e-16 &  0.27e-14 & & & & \\
270 &  0.59e-19 &  0.28e-17 &  0.56e-19 &  0.28e-17 &  0.70e-20 &  0.29e-18 &  0.67e-20 &  0.29e-18 &  0.82e-19 &  0.12e-17 &  0.80e-19 &  0.12e-17 \\
\hline
\multicolumn{13}{c}{\textbf{Measure: Deviation~~~~Grid type: Chebyshev}}\\
\hline
& \multicolumn{12}{c}{Network Parameters} \\
& \multicolumn{4}{c}{90} & \multicolumn{4}{c}{180} & \multicolumn{4}{c}{270} \\
\cline{2-5} \cline{6-9} \cline{10-13}
& \multicolumn{2}{c}{Training} & \multicolumn{2}{c}{Test} & \multicolumn{2}{c}{Training} & \multicolumn{2}{c}{Test} & \multicolumn{2}{c}{Training} & \multicolumn{2}{c}{Test} \\
\cline{2-3} \cline{4-5} \cline{6-7} \cline{8-9} \cline{10-11} \cline{12-13}
$M_{\rm tr}$ & MSD & MXD & MSD & MXD & MSD & MXD & MSD & MXD & MSD & MXD & MSD & MXD \\
\hline
90  &  0.17e-21 &  0.14e-20 &  0.24e-21 &  0.14e-20 & & & & & & & & \\
180 &  0.25e-20 &  0.18e-19 &  0.31e-20 &  0.18e-19 &  0.15e-20 &  0.14e-19 &  0.20e-20 &  0.15e-19 & & & & \\
270 &  0.81e-20 &  0.89e-19 &  0.12e-19 &  0.89e-19 &  0.76e-21 &  0.67e-20 &  0.11e-20 &  0.69e-20 &  0.56e-20 &  0.37e-19 &  0.78e-20 &  0.37e-19 \\
\hline
\multicolumn{13}{c}{\textbf{Measure: Error~~~~Grid type: Equidistant}}\\
\hline
& \multicolumn{12}{c}{Network Parameters} \\
& \multicolumn{4}{c}{90} & \multicolumn{4}{c}{180} & \multicolumn{4}{c}{270} \\
\cline{2-5} \cline{6-9} \cline{10-13}
& \multicolumn{2}{c}{Training} & \multicolumn{2}{c}{Test} & \multicolumn{2}{c}{Training}  & \multicolumn{2}{c}{Test} & \multicolumn{2}{c}{Training} & \multicolumn{2}{c}{Test} \\
\cline{2-3} \cline{4-5} \cline{6-7} \cline{8-9} \cline{10-11} \cline{12-13}
$M_{\rm tr}$ & MSE & MXE & MSE & MXE & MSE & MXE & MSE & MXE & MSE & MXE & MSE & MXE \\
\hline
90  &  0.74e-18 &  0.70e-17 &  0.25e-10 &  0.59e-08 & & & & & & & & \\
180 &  0.69e-18 &  0.12e-16 &  0.12e-14 &  0.50e-12 &  0.49e-17 &  0.12e-15 &  0.15e-12 &  0.60e-10 & & & & \\
270 &  0.58e-17 &  0.62e-16 &  0.23e-15 &  0.15e-12 &  0.29e-17 &  0.20e-16 &  0.27e-16 &  0.15e-13 &  0.12e-16 &  0.39e-15 &  0.12e-15 &  0.63e-13 \\
\hline
\multicolumn{13}{c}{\textbf{Measure: Error~~~~Grid type: Chebyshev}}\\
\hline
& \multicolumn{12}{c}{Network Parameters} \\
& \multicolumn{4}{c}{90} & \multicolumn{4}{c}{180} & \multicolumn{4}{c}{270} \\
\cline{2-5} \cline{6-9} \cline{10-13}
& \multicolumn{2}{c}{Training} & \multicolumn{2}{c}{Test} & \multicolumn{2}{c}{Training} & \multicolumn{2}{c}{Test} & \multicolumn{2}{c}{Training} & \multicolumn{2}{c}{Test} \\
\cline{2-3} \cline{4-5} \cline{6-7} \cline{8-9} \cline{10-11} \cline{12-13}
$M_{\rm tr}$ & MSE & MXE & MSE & MXE & MSE & MXE & MSE & MXE & MSE & MXE & MSE & MXE \\
\hline
90  &  0.16e-18 &  0.59e-18 &  0.17e-18 &  0.73e-18 & & & & & & & & \\
180 &  0.12e-17 &  0.12e-16 &  0.93e-18 &  0.13e-16 &  0.31e-17 &  0.21e-16 &  0.35e-17 &  0.21e-16 & & & & \\
270 &  0.70e-17 &  0.77e-16 &  0.45e-17 &  0.80e-16 &  0.75e-17 &  0.30e-16 &  0.81e-17 &  0.31e-16 &  0.23e-16 &  0.16e-15 &  0.18e-16 &  0.16e-15 \\
\hline
\end{tabular}
\caption{Performance measures for all scenarios of Test Problem 2 with mixed Dirichlet-Neumann conditions.}
\label{tab:Res_Prob_2_DN}
\end{table}


\begin{table}[h!]
\setlength{\tabcolsep}{2pt}
\centering
\scriptsize
\begin{tabular}{@{\extracolsep{3pt}}ccccccccccccc@{}}
\hline
\multicolumn{13}{c}{\textbf{Measure: Deviation~~~~Grid type: Equidistant}}\\
\hline
& \multicolumn{12}{c}{Network Parameters} \\
& \multicolumn{4}{c}{90} & \multicolumn{4}{c}{180} & \multicolumn{4}{c}{270} \\
\cline{2-5} \cline{6-9} \cline{10-13}
& \multicolumn{2}{c}{Training} & \multicolumn{2}{c}{Test} & \multicolumn{2}{c}{Training} & \multicolumn{2}{c}{Test} & \multicolumn{2}{c}{Training} & \multicolumn{2}{c}{Test} \\
\cline{2-3} \cline{4-5} \cline{6-7} \cline{8-9} \cline{10-11} \cline{12-13}
$M_{\rm tr}$ & MSD & MXD & MSD & MXD & MSD & MXD & MSD & MXD & MSD & MXD & MSD & MXD \\
\hline
90  &  0.21e-17 &  0.18e-15 &  0.39e-18 &  0.18e-15 & & & & & & & & \\
180 &  0.24e-18 &  0.13e-16 &  0.21e-18 &  0.13e-16 &  0.24e-14 &  0.12e-12 &  0.21e-14 &  0.12e-12 & & & & \\
270 &  0.39e-21 &  0.17e-19 &  0.37e-21 &  0.18e-19 &  0.92e-17 &  0.48e-15 &  0.85e-17 &  0.48e-15 &  0.12e-12 &  0.61e-11 &  0.11e-12 &  0.61e-11 \\
\hline
\multicolumn{13}{c}{\textbf{Measure: Deviation~~~~Grid type: Chebyshev}}\\
\hline
& \multicolumn{12}{c}{Network Parameters} \\
& \multicolumn{4}{c}{90} & \multicolumn{4}{c}{180} & \multicolumn{4}{c}{270} \\
\cline{2-5} \cline{6-9} \cline{10-13}
& \multicolumn{2}{c}{Training} & \multicolumn{2}{c}{Test} & \multicolumn{2}{c}{Training} & \multicolumn{2}{c}{Test} & \multicolumn{2}{c}{Training} & \multicolumn{2}{c}{Test} \\
\cline{2-3} \cline{4-5} \cline{6-7} \cline{8-9} \cline{10-11} \cline{12-13}
$M_{\rm tr}$ & MSD & MXD & MSD & MXD & MSD & MXD & MSD & MXD & MSD & MXD & MSD & MXD \\
\hline
90  &  0.11e-18 &  0.16e-17 &  0.29e-19 &  0.16e-17 & & & & & & & & \\
180 &  0.43e-22 &  0.58e-21 &  0.13e-22 &  0.58e-21 &  0.94e-17 &  0.13e-15 &  0.24e-17 &  0.13e-15 & & & & \\
270 &  0.25e-18 &  0.34e-17 &  0.62e-19 &  0.34e-17 &  0.22e-17 &  0.31e-16 &  0.56e-18 &  0.31e-16 &  0.19e-18 &  0.27e-17 &  0.45e-19 &  0.27e-17 \\
\hline
\multicolumn{13}{c}{\textbf{Measure: Error~~~~Grid type: Equidistant}}\\
\hline
& \multicolumn{12}{c}{Network Parameters} \\
& \multicolumn{4}{c}{90} & \multicolumn{4}{c}{180} & \multicolumn{4}{c}{270} \\
\cline{2-5} \cline{6-9} \cline{10-13}
& \multicolumn{2}{c}{Training} & \multicolumn{2}{c}{Test} & \multicolumn{2}{c}{Training}  & \multicolumn{2}{c}{Test} & \multicolumn{2}{c}{Training} & \multicolumn{2}{c}{Test} \\
\cline{2-3} \cline{4-5} \cline{6-7} \cline{8-9} \cline{10-11} \cline{12-13}
$M_{\rm tr}$ & MSE & MXE & MSE & MXE & MSE & MXE & MSE & MXE & MSE & MXE & MSE & MXE \\
\hline
90  &  0.15e-19 &  0.22e-18 &  0.60e-13 &  0.15e-10 & & & & & & & & \\
180 &  0.53e-18 &  0.87e-17 &  0.25e-15 &  0.97e-13 &  0.17e-19 &  0.73e-19 &  0.50e-16 &  0.16e-13 & & & & \\
270 &  0.17e-18 &  0.16e-17 &  0.93e-18 &  0.47e-15 &  0.68e-17 &  0.41e-16 &  0.23e-15 &  0.17e-12 &  0.25e-16 &  0.30e-15 &  0.31e-14 &  0.20e-11 \\
\hline
\multicolumn{13}{c}{\textbf{Measure: Error~~~~Grid type: Chebyshev}}\\
\hline
& \multicolumn{12}{c}{Network Parameters} \\
& \multicolumn{4}{c}{90} & \multicolumn{4}{c}{180} & \multicolumn{4}{c}{270} \\
\cline{2-5} \cline{6-9} \cline{10-13}
& \multicolumn{2}{c}{Training} & \multicolumn{2}{c}{Test} & \multicolumn{2}{c}{Training} & \multicolumn{2}{c}{Test} & \multicolumn{2}{c}{Training} & \multicolumn{2}{c}{Test} \\
\cline{2-3} \cline{4-5} \cline{6-7} \cline{8-9} \cline{10-11} \cline{12-13}
$M_{\rm tr}$ & MSE & MXE & MSE & MXE & MSE & MXE & MSE & MXE & MSE & MXE & MSE & MXE \\
\hline
90  &  0.47e-18 &  0.26e-17 &  0.48e-18 &  0.26e-17 & & & & & & & & \\
180 &  0.13e-18 &  0.55e-18 &  0.11e-18 &  0.62e-18 &  0.10e-15 &  0.50e-15 &  0.10e-15 &  0.52e-15 & & & & \\
270 &  0.74e-18 &  0.40e-17 &  0.53e-18 &  0.42e-17 &  0.68e-17 &  0.55e-16 &  0.55e-17 &  0.55e-16 &  0.17e-16 &  0.10e-15 &  0.14e-16 &  0.10e-15 \\
\hline
\end{tabular}
\caption{Performance measures for all scenarios of Test Problem 2 with Cauchy conditions.}
\label{tab:Res_Prob_2_C}
\end{table}

\begin{table}[h!]
\setlength{\tabcolsep}{2pt}
\centering
\scriptsize
\begin{tabular}{@{\extracolsep{3pt}}ccccccccccccc@{}}
\hline
\multicolumn{13}{c}{\textbf{Measure: Deviation~~~~Grid type: Equidistant}}\\
\hline
& \multicolumn{12}{c}{Network Parameters} \\
& \multicolumn{4}{c}{90} & \multicolumn{4}{c}{180} & \multicolumn{4}{c}{270} \\
\cline{2-5} \cline{6-9} \cline{10-13}
& \multicolumn{2}{c}{Training} & \multicolumn{2}{c}{Test} & \multicolumn{2}{c}{Training} & \multicolumn{2}{c}{Test} & \multicolumn{2}{c}{Training} & \multicolumn{2}{c}{Test} \\
\cline{2-3} \cline{4-5} \cline{6-7} \cline{8-9} \cline{10-11} \cline{12-13}
$M_{\rm tr}$ & MSD & MXD & MSD & MXD & MSD & MXD & MSD & MXD & MSD & MXD & MSD & MXD \\
\hline
90  &  0.13e-13 &  0.48e-12 &  0.11e-13 &  0.48e-12 & & & & & & & & \\
180 &  0.98e-17 &  0.44e-15 &  0.89e-17 &  0.44e-15 &  0.13e-16 &  0.56e-15 &  0.11e-16 &  0.56e-15 & & & & \\
270 &  0.54e-19 &  0.19e-17 &  0.52e-19 &  0.19e-17 &  0.27e-20 &  0.12e-19 &  0.27e-20 &  0.12e-19 &  0.33e-20 &  0.16e-18 &  0.31e-20 &  0.16e-18 \\
\hline
\multicolumn{13}{c}{\textbf{Measure: Deviation~~~~Grid type: Chebyshev}}\\
\hline
& \multicolumn{12}{c}{Network Parameters} \\
& \multicolumn{4}{c}{90} & \multicolumn{4}{c}{180} & \multicolumn{4}{c}{270} \\
\cline{2-5} \cline{6-9} \cline{10-13}
& \multicolumn{2}{c}{Training} & \multicolumn{2}{c}{Test} & \multicolumn{2}{c}{Training} & \multicolumn{2}{c}{Test} & \multicolumn{2}{c}{Training} & \multicolumn{2}{c}{Test} \\
\cline{2-3} \cline{4-5} \cline{6-7} \cline{8-9} \cline{10-11} \cline{12-13}
$M_{\rm tr}$ & MSD & MXD & MSD & MXD & MSD & MXD & MSD & MXD & MSD & MXD & MSD & MXD \\
\hline
90  &  0.24e-21 &  0.81e-21 &  0.20e-21 &  0.81e-21 & & & & & & & & \\
180 &  0.61e-22 &  0.45e-21 &  0.90e-22 &  0.46e-21 &  0.34e-20 &  0.17e-19 &  0.29e-20 &  0.17e-19 & & & & \\
270 &  0.61e-23 &  0.25e-22 &  0.65e-23 &  0.26e-22 &  0.25e-20 &  0.79e-20 &  0.22e-20 &  0.79e-20 &  0.13e-20 &  0.67e-20 &  0.18e-20 &  0.68e-20 \\
\hline
\multicolumn{13}{c}{\textbf{Measure: Error~~~~Grid type: Equidistant}}\\
\hline
& \multicolumn{12}{c}{Network Parameters} \\
& \multicolumn{4}{c}{90} & \multicolumn{4}{c}{180} & \multicolumn{4}{c}{270} \\
\cline{2-5} \cline{6-9} \cline{10-13}
& \multicolumn{2}{c}{Training} & \multicolumn{2}{c}{Test} & \multicolumn{2}{c}{Training}  & \multicolumn{2}{c}{Test} & \multicolumn{2}{c}{Training} & \multicolumn{2}{c}{Test} \\
\cline{2-3} \cline{4-5} \cline{6-7} \cline{8-9} \cline{10-11} \cline{12-13}
$M_{\rm tr}$ & MSE & MXE & MSE & MXE & MSE & MXE & MSE & MXE & MSE & MXE & MSE & MXE \\
\hline
90  &  0.53e-19 &  0.71e-18 &  0.14e-10 &  0.35e-08 & & & & & & & & \\
180 &  0.34e-18 &  0.89e-17 &  0.23e-13 &  0.10e-10 &  0.48e-18 &  0.50e-17 &  0.30e-13 &  0.13e-10 & & & & \\
270 &  0.99e-18 &  0.27e-16 &  0.16e-15 &  0.10e-12 &  0.65e-19 &  0.51e-18 &  0.70e-19 &  0.27e-17 &  0.11e-18 &  0.27e-17 &  0.13e-16 &  0.89e-14 \\
\hline
\multicolumn{13}{c}{\textbf{Measure: Error~~~~Grid type: Chebyshev}}\\
\hline
& \multicolumn{12}{c}{Network Parameters} \\
& \multicolumn{4}{c}{90} & \multicolumn{4}{c}{180} & \multicolumn{4}{c}{270} \\
\cline{2-5} \cline{6-9} \cline{10-13}
& \multicolumn{2}{c}{Training} & \multicolumn{2}{c}{Test} & \multicolumn{2}{c}{Training} & \multicolumn{2}{c}{Test} & \multicolumn{2}{c}{Training} & \multicolumn{2}{c}{Test} \\
\cline{2-3} \cline{4-5} \cline{6-7} \cline{8-9} \cline{10-11} \cline{12-13}
$M_{\rm tr}$ & MSE & MXE & MSE & MXE & MSE & MXE & MSE & MXE & MSE & MXE & MSE & MXE \\
\hline
90  &  0.25e-18 &  0.15e-17 &  0.20e-18 &  0.16e-17 & & & & & & & & \\
180 &  0.29e-19 &  0.12e-18 &  0.29e-19 &  0.12e-18 &  0.53e-17 &  0.60e-16 &  0.30e-17 &  0.63e-16 & & & & \\
270 &  0.37e-19 &  0.29e-18 &  0.26e-19 &  0.31e-18 &  0.21e-18 &  0.18e-17 &  0.15e-18 &  0.18e-17 &  0.10e-17 &  0.76e-17 &  0.78e-18 &  0.78e-17 \\
\hline
\end{tabular}
\caption{Performance measures for all scenarios of Test Problem 2 with Neumann conditions.}
\label{tab:Res_Prob_2_NN}
\end{table}


\begin{table}[h!]
\setlength{\tabcolsep}{2pt}
\centering
\scriptsize
\begin{tabular}{@{\extracolsep{3pt}}ccccccccccccc@{}}
\hline
\multicolumn{13}{c}{\textbf{Measure: Deviation~~~~Grid type: Equidistant}}\\
\hline
& \multicolumn{12}{c}{Network Parameters} \\
& \multicolumn{4}{c}{90} & \multicolumn{4}{c}{180} & \multicolumn{4}{c}{270} \\
\cline{2-5} \cline{6-9} \cline{10-13}
& \multicolumn{2}{c}{Training} & \multicolumn{2}{c}{Test} & \multicolumn{2}{c}{Training} & \multicolumn{2}{c}{Test} & \multicolumn{2}{c}{Training} & \multicolumn{2}{c}{Test} \\
\cline{2-3} \cline{4-5} \cline{6-7} \cline{8-9} \cline{10-11} \cline{12-13}
$M_{\rm tr}$ & MSD & MXD & MSD & MXD & MSD & MXD & MSD & MXD & MSD & MXD & MSD & MXD \\
\hline
90  &  0.15e-13 &  0.48e-12 &  0.13e-13 &  0.53e-12 & & & & & & & & \\
180 &  0.65e-20 &  0.25e-18 &  0.60e-20 &  0.26e-18 &  0.13e-15 &  0.68e-15 &  0.13e-15 &  0.68e-15 & & & & \\
270 &  0.30e-21 &  0.11e-19 &  0.29e-21 &  0.12e-19 &  0.13e-21 &  0.58e-20 &  0.12e-21 &  0.60e-20 &  0.12e-20 &  0.36e-19 &  0.12e-20 &  0.38e-19 \\
\hline
\multicolumn{13}{c}{\textbf{Measure: Deviation~~~~Grid type: Chebyshev}}\\
\hline
& \multicolumn{12}{c}{Network Parameters} \\
& \multicolumn{4}{c}{90} & \multicolumn{4}{c}{180} & \multicolumn{4}{c}{270} \\
\cline{2-5} \cline{6-9} \cline{10-13}
& \multicolumn{2}{c}{Training} & \multicolumn{2}{c}{Test} & \multicolumn{2}{c}{Training} & \multicolumn{2}{c}{Test} & \multicolumn{2}{c}{Training} & \multicolumn{2}{c}{Test} \\
\cline{2-3} \cline{4-5} \cline{6-7} \cline{8-9} \cline{10-11} \cline{12-13}
$M_{\rm tr}$ & MSD & MXD & MSD & MXD & MSD & MXD & MSD & MXD & MSD & MXD & MSD & MXD \\
\hline
90  &  0.20e-22 &  0.15e-21 &  0.28e-22 &  0.16e-21 & & & & & & & & \\
180 &  0.87e-22 &  0.77e-21 &  0.13e-21 &  0.79e-21 &  0.53e-22 &  0.32e-21 &  0.75e-22 &  0.33e-21 & & & & \\
270 &  0.60e-22 &  0.46e-21 &  0.87e-22 &  0.47e-21 &  0.40e-23 &  0.37e-22 &  0.59e-23 &  0.37e-22 &  0.32e-22 &  0.31e-21 &  0.45e-22 &  0.31e-21 \\
\hline
\multicolumn{13}{c}{\textbf{Measure: Error~~~~Grid type: Equidistant}}\\
\hline
& \multicolumn{12}{c}{Network Parameters} \\
& \multicolumn{4}{c}{90} & \multicolumn{4}{c}{180} & \multicolumn{4}{c}{270} \\
\cline{2-5} \cline{6-9} \cline{10-13}
& \multicolumn{2}{c}{Training} & \multicolumn{2}{c}{Test} & \multicolumn{2}{c}{Training}  & \multicolumn{2}{c}{Test} & \multicolumn{2}{c}{Training} & \multicolumn{2}{c}{Test} \\
\cline{2-3} \cline{4-5} \cline{6-7} \cline{8-9} \cline{10-11} \cline{12-13}
$M_{\rm tr}$ & MSE & MXE & MSE & MXE & MSE & MXE & MSE & MXE & MSE & MXE & MSE & MXE \\
\hline
90  &  0.47e-19 &  0.49e-18 &  0.43e-10 &  0.11e-07 & & & & & & & & \\
180 &  0.14e-18 &  0.10e-17 &  0.41e-16 &  0.19e-13 &  0.24e-19 &  0.38e-18 &  0.17e-13 &  0.71e-11 & & & & \\
270 &  0.14e-18 &  0.25e-17 &  0.30e-17 &  0.18e-14 &  0.12e-19 &  0.51e-18 &  0.14e-17 &  0.95e-15 &  0.26e-19 &  0.15e-18 &  0.85e-17 &  0.61e-14 \\
\hline
\multicolumn{13}{c}{\textbf{Measure: Error~~~~Grid type: Chebyshev}}\\
\hline
& \multicolumn{12}{c}{Network Parameters} \\
& \multicolumn{4}{c}{90} & \multicolumn{4}{c}{180} & \multicolumn{4}{c}{270} \\
\cline{2-5} \cline{6-9} \cline{10-13}
& \multicolumn{2}{c}{Training} & \multicolumn{2}{c}{Test} & \multicolumn{2}{c}{Training} & \multicolumn{2}{c}{Test} & \multicolumn{2}{c}{Training} & \multicolumn{2}{c}{Test} \\
\cline{2-3} \cline{4-5} \cline{6-7} \cline{8-9} \cline{10-11} \cline{12-13}
$M_{\rm tr}$ & MSE & MXE & MSE & MXE & MSE & MXE & MSE & MXE & MSE & MXE & MSE & MXE \\
\hline
90  &  0.18e-19 &  0.92e-19 &  0.17e-19 &  0.95e-19 & & & & & & & & \\
180 &  0.16e-18 &  0.75e-18 &  0.20e-18 &  0.76e-18 &  0.16e-19 &  0.48e-19 &  0.17e-19 &  0.48e-19 & & & & \\
270 &  0.17e-18 &  0.94e-18 &  0.14e-18 &  0.95e-18 &  0.11e-19 &  0.40e-19 &  0.14e-19 &  0.42e-19 &  0.30e-18 &  0.42e-17 &  0.15e-18 &  0.47e-17 \\
\hline
\end{tabular}
\caption{Performance measures for all scenarios of Test Problem 2 with Robin conditions.}
\label{tab:Res_Prob_2_R}
\end{table}



\begin{table}[h!]
\setlength{\tabcolsep}{2pt}
\centering
\scriptsize
\begin{tabular}{@{\extracolsep{3pt}}ccccccccccccc@{}}
\hline
\multicolumn{13}{c}{\textbf{Measure: Deviation~~~~Grid type: Equidistant}}\\
\hline
 & \multicolumn{12}{c}{Network Parameters} \\
 & \multicolumn{4}{c}{90} & \multicolumn{4}{c}{180} & \multicolumn{4}{c}{270} \\
\cline{2-5} \cline{6-9} \cline{10-13}
 & \multicolumn{2}{c}{Training} & \multicolumn{2}{c}{Test} & \multicolumn{2}{c}{Training} & \multicolumn{2}{c}{Test} & \multicolumn{2}{c}{Training} & \multicolumn{2}{c}{Test} \\
\cline{2-3} \cline{4-5} \cline{6-7} \cline{8-9} \cline{10-11} \cline{12-13}
$M_{\rm tr}$ & MSD & MXD & MSD & MXD & MSD & MXD & MSD & MXD & MSD & MXD & MSD & MXD \\
\hline
90  &  0.15e-16 &  0.67e-16 &  0.15e-16 &  0.67e-16 & & & & & & & & \\
180 &  0.18e-16 &  0.83e-16 &  0.18e-16 &  0.83e-16 &  0.45e-18 &  0.21e-17 &  0.45e-18 &  0.21e-17 & & & & \\
270 &  0.66e-17 &  0.30e-16 &  0.66e-17 &  0.30e-16 &  0.21e-16 &  0.10e-15 &  0.22e-16 &  0.10e-15 &  0.20e-17 &  0.93e-17 &  0.20e-17 &  0.93e-17 \\
\hline
\multicolumn{13}{c}{\textbf{Measure: Deviation~~~~Grid type: Chebyshev}}\\
\hline
 & \multicolumn{12}{c}{Network Parameters} \\
 & \multicolumn{4}{c}{90} & \multicolumn{4}{c}{180} & \multicolumn{4}{c}{270} \\
\cline{2-5} \cline{6-9} \cline{10-13}
 & \multicolumn{2}{c}{Training} & \multicolumn{2}{c}{Test} & \multicolumn{2}{c}{Training} & \multicolumn{2}{c}{Test} & \multicolumn{2}{c}{Training} & \multicolumn{2}{c}{Test} \\
\cline{2-3} \cline{4-5} \cline{6-7} \cline{8-9} \cline{10-11} \cline{12-13}
$M_{\rm tr}$ & MSD & MXD & MSD & MXD & MSD & MXD & MSD & MXD & MSD & MXD & MSD & MXD \\
\hline
90  &  0.89e-17 &  0.39e-16 &  0.85e-17 &  0.39e-16 & & & & & & & & \\
180 &  0.36e-17 &  0.16e-16 &  0.35e-17 &  0.16e-16 &  0.48e-17 &  0.21e-16 &  0.45e-17 &  0.21e-16 & & & & \\
270 &  0.59e-17 &  0.26e-16 &  0.56e-17 &  0.26e-16 &  0.85e-17 &  0.37e-16 &  0.78e-17 &  0.37e-16 &  0.28e-18 &  0.12e-17 &  0.27e-18 &  0.12e-17 \\
\hline
\multicolumn{13}{c}{\textbf{Measure: Error~~~~Grid type: Equidistant}}\\
\hline
 & \multicolumn{12}{c}{Network Parameters} \\
 & \multicolumn{4}{c}{90} & \multicolumn{4}{c}{180} & \multicolumn{4}{c}{270} \\
 \cline{2-5} \cline{6-9} \cline{10-13}
  & \multicolumn{2}{c}{Training} & \multicolumn{2}{c}{Test} & \multicolumn{2}{c}{Training}  & \multicolumn{2}{c}{Test} & \multicolumn{2}{c}{Training} & \multicolumn{2}{c}{Test} \\
 \cline{2-3} \cline{4-5} \cline{6-7} \cline{8-9} \cline{10-11} \cline{12-13}
 $M_{\rm tr}$ & MSE & MXE & MSE & MXE & MSE & MXE & MSE & MXE & MSE & MXE & MSE & MXE \\
 \hline
90  &  0.28e-18 &  0.33e-17 &  0.29e-18 &  0.61e-17 & & & & & & & & \\
180 &  0.80e-18 &  0.69e-17 &  0.80e-18 &  0.71e-17 &  0.57e-19 &  0.97e-18 &  0.56e-19 &  0.99e-18 & & & & \\
270 &  0.10e-18 &  0.28e-17 &  0.10e-18 &  0.28e-17 &  0.21e-18 &  0.62e-17 &  0.21e-18 &  0.62e-17 &  0.59e-19 &  0.78e-18 &  0.59e-19 &  0.81e-18 \\
 \hline
 \multicolumn{13}{c}{\textbf{Measure: Error~~~~Grid type: Chebyshev}}\\
 \hline
  & \multicolumn{12}{c}{Network Parameters} \\
  & \multicolumn{4}{c}{90} & \multicolumn{4}{c}{180} & \multicolumn{4}{c}{270} \\
 \cline{2-5} \cline{6-9} \cline{10-13}
  & \multicolumn{2}{c}{Training} & \multicolumn{2}{c}{Test} & \multicolumn{2}{c}{Training} & \multicolumn{2}{c}{Test} & \multicolumn{2}{c}{Training} & \multicolumn{2}{c}{Test} \\
 \cline{2-3} \cline{4-5} \cline{6-7} \cline{8-9} \cline{10-11} \cline{12-13}
 $M_{\rm tr}$ & MSE & MXE & MSE & MXE & MSE & MXE & MSE & MXE & MSE & MXE & MSE & MXE \\
 \hline
90  &  0.18e-17 &  0.71e-17 &  0.22e-17 &  0.77e-17 & & & & & & & & \\
180 &  0.50e-18 &  0.18e-17 &  0.43e-18 &  0.18e-17 &  0.69e-18 &  0.27e-17 &  0.83e-18 &  0.28e-17 & & & & \\
270 &  0.74e-18 &  0.31e-17 &  0.83e-18 &  0.32e-17 &  0.40e-18 &  0.20e-17 &  0.47e-18 &  0.19e-17 &  0.54e-19 &  0.22e-18 &  0.68e-19 &  0.22e-18 \\
 \hline
 \end{tabular}
 \caption{Performance measures for all scenarios of Test Problem 3.}
 \label{tab:Res_Prob_3}
 \end{table} 



\begin{table}[h!]
\setlength{\tabcolsep}{2pt}
\centering
\scriptsize
\begin{tabular}{@{\extracolsep{3pt}}cccccccccccccc@{}}
\hline
\multicolumn{14}{c}{\textbf{Measure: Deviation~~~~Grid type: Equidistant}}\\
\hline
&& \multicolumn{12}{c}{Network Parameters} \\
&& \multicolumn{4}{c}{60} & \multicolumn{4}{c}{120} & \multicolumn{4}{c}{240} \\
\cline{3-6}\cline{7-10}\cline{11-14}
&& \multicolumn{2}{c}{Training} & \multicolumn{2}{c}{Test} & \multicolumn{2}{c}{Training} & \multicolumn{2}{c}{Test} & \multicolumn{2}{c}{Training} & \multicolumn{2}{c}{Test} \\
\cline{3-4} \cline{5-6} \cline{7-8} \cline{9-10} \cline{11-12} \cline{13-14}
$M_{\rm tr}$& Solution & MSD & MXD & MSD & MXD & MSD & MXD & MSD & MXD & MSD & MXD & MSD & MXD \\
\hline
70&$\psi_1$& 0.14e-16& 0.24e-15& 0.13e-16& 0.24e-15&&&&&&&& \\
&$\psi_2$& 0.10e-15& 0.20e-14& 0.89e-16& 0.20e-14&&&&&&&& \\
130&$\psi_1$& 0.59e-16& 0.11e-14& 0.56e-16& 0.11e-14& 0.81e-17& 0.15e-15& 0.76e-17& 0.15e-15&&&& \\
&$\psi_2$& 0.42e-15& 0.87e-14& 0.39e-15& 0.87e-14& 0.57e-16& 0.12e-14& 0.53e-16& 0.12e-14&&&& \\
250&$\psi_1$& 0.41e-16& 0.76e-15& 0.40e-16& 0.76e-15& 0.37e-18& 0.68e-17& 0.36e-18& 0.68e-17& 0.92e-18& 0.17e-16& 0.89e-18& 0.17e-16 \\
&$\psi_2$& 0.29e-15& 0.62e-14& 0.28e-15& 0.62e-14& 0.26e-17& 0.55e-16& 0.25e-17& 0.55e-16& 0.65e-17& 0.14e-15& 0.63e-17& 0.14e-15 \\
\hline
\multicolumn{14}{c}{\textbf{Measure: Deviation~~~~Grid type: Chebyshev}}\\
\hline
&& \multicolumn{12}{c}{Network Parameters}\\
&& \multicolumn{4}{c}{60} & \multicolumn{4}{c}{120} & \multicolumn{4}{c}{240} \\
\cline{3-6}\cline{7-10}\cline{11-14}
&& \multicolumn{2}{c}{Training} & \multicolumn{2}{c}{Test} & \multicolumn{2}{c}{Training} & \multicolumn{2}{c}{Test} & \multicolumn{2}{c}{Training} & \multicolumn{2}{c}{Test} \\
\cline{3-4} \cline{5-6} \cline{7-8} \cline{9-10} \cline{11-12} \cline{13-14}
$M_{\rm tr}$& Solution & MSD & MXD & MSD & MXD & MSD & MXD & MSD & MXD & MSD & MXD & MSD & MXD \\
\hline
70&$\psi_1$& 0.44e-14& 0.33e-13& 0.18e-14& 0.33e-13&&&&&&&& \\
&$\psi_2$& 0.33e-13& 0.27e-12& 0.12e-13& 0.27e-12&&&&&&&& \\
130&$\psi_1$& 0.33e-16& 0.25e-15& 0.13e-16& 0.25e-15& 0.64e-17& 0.49e-16& 0.26e-17& 0.49e-16&&&& \\
&$\psi_2$& 0.25e-15& 0.21e-14& 0.93e-16& 0.21e-14& 0.48e-16& 0.40e-15& 0.18e-16& 0.40e-15&&&& \\
250&$\psi_1$& 0.12e-16& 0.92e-16& 0.48e-17& 0.92e-16& 0.12e-17& 0.91e-17& 0.48e-18& 0.91e-17& 0.15e-18& 0.12e-17& 0.61e-19& 0.12e-17 \\
&$\psi_2$& 0.91e-16& 0.75e-15& 0.34e-16& 0.75e-15& 0.90e-17& 0.74e-16& 0.33e-17& 0.74e-16& 0.12e-17& 0.95e-17& 0.43e-18& 0.95e-17 \\
\hline
\multicolumn{14}{c}{\textbf{Measure: Error~~~~Grid type: Equidistant}}\\
\hline
&& \multicolumn{12}{c}{Network Parameters}\\
&& \multicolumn{4}{c}{60} & \multicolumn{4}{c}{120} & \multicolumn{4}{c}{240} \\
\cline{3-6}\cline{7-10}\cline{11-14}
&& \multicolumn{2}{c}{Training} & \multicolumn{2}{c}{Test} & \multicolumn{2}{c}{Training} & \multicolumn{2}{c}{Test} & \multicolumn{2}{c}{Training} & \multicolumn{2}{c}{Test} \\
\cline{3-4} \cline{5-6} \cline{7-8} \cline{9-10} \cline{11-12} \cline{13-14}
$M_{\rm tr}$& & MSE & MXE & MSE & MXE & MSE & MXE & MSE & MXE & MSE & MXE & MSE & MXE \\
\hline
70&& 0.36e-20& 0.19e-18& 0.36e-20& 0.28e-18&&&&&&&& \\
130&& 0.24e-20& 0.39e-18& 0.22e-20& 0.39e-18& 0.69e-21& 0.87e-19& 0.63e-21& 0.87e-19&&&& \\
250&& 0.23e-20& 0.81e-18& 0.21e-20& 0.81e-18& 0.24e-21& 0.37e-19& 0.23e-21& 0.37e-19& 0.11e-21& 0.30e-19& 0.10e-21& 0.30e-19 \\
\hline
\multicolumn{14}{c}{\textbf{Measure: Error~~~~Grid type: Chebyshev}}\\
\hline
&& \multicolumn{12}{c}{Network Parameters}\\
&& \multicolumn{4}{c}{60} & \multicolumn{4}{c}{120} & \multicolumn{4}{c}{240} \\
\cline{3-6}\cline{7-10}\cline{11-14}
&& \multicolumn{2}{c}{Training} & \multicolumn{2}{c}{Test} & \multicolumn{2}{c}{Training} & \multicolumn{2}{c}{Test} & \multicolumn{2}{c}{Training} & \multicolumn{2}{c}{Test} \\
\cline{3-4} \cline{5-6} \cline{7-8} \cline{9-10} \cline{11-12} \cline{13-14}
$M_{\rm tr}$& & MSE & MXE & MSE & MXE & MSE & MXE & MSE & MXE & MSE & MXE & MSE & MXE \\
\hline
70&& 0.10e-19& 0.47e-18& 0.93e-20& 0.47e-18&&&&&&&& \\
130&& 0.48e-20& 0.16e-18& 0.50e-20& 0.16e-18& 0.20e-21& 0.69e-20& 0.19e-21& 0.71e-20&&&& \\
250&& 0.38e-20& 0.15e-18& 0.38e-20& 0.15e-18& 0.49e-21& 0.13e-19& 0.45e-21& 0.13e-19& 0.94e-22& 0.45e-20& 0.87e-22& 0.45e-20 \\
\hline
\end{tabular}
\caption{Performance measures for all scenarios of Test Problem 4.}
\label{tab:Res_Prob_4}
\end{table}



\begin{table}[h!]
\setlength{\tabcolsep}{2pt}
\centering
\scriptsize
\begin{tabular}{@{\extracolsep{3pt}}cccccccccccccc@{}}
\hline
\multicolumn{14}{c}{\textbf{Measure: Deviation~~~~Grid type: Equidistant}}\\
\hline
&& \multicolumn{12}{c}{Network Parameters} \\
&& \multicolumn{4}{c}{60} & \multicolumn{4}{c}{120} & \multicolumn{4}{c}{240} \\
\cline{3-6}\cline{7-10}\cline{11-14}
&& \multicolumn{2}{c}{Training} & \multicolumn{2}{c}{Test} & \multicolumn{2}{c}{Training} & \multicolumn{2}{c}{Test} & \multicolumn{2}{c}{Training} & \multicolumn{2}{c}{Test} \\
\cline{3-4} \cline{5-6} \cline{7-8} \cline{9-10} \cline{11-12} \cline{13-14}
$M_{\rm tr}$& Solution & MSD & MXD & MSD & MXD & MSD & MXD & MSD & MXD & MSD & MXD & MSD & MXD \\
\hline
70 & $\psi_1$ &  0.28e-08 &  0.22e-07 &  0.29e-08 &  0.22e-07 & & & & & & & & \\
 & $\psi_2$ &  0.81e-08 &  0.12e-06 &  0.86e-08 &  0.15e-06 & & & & & & & & \\
130 & $\psi_1$ &  0.51e-13 &  0.13e-11 &  0.52e-13 &  0.15e-11 &  0.21e-11 &  0.23e-10 &  0.21e-11 &  0.23e-10 & & & & \\
 & $\psi_2$ &  0.74e-13 &  0.57e-12 &  0.75e-13 &  0.57e-12 &  0.47e-11 &  0.46e-10 &  0.47e-11 &  0.46e-10 & & & & \\
250 & $\psi_1$ &  0.14e-15 &  0.41e-14 &  0.14e-15 &  0.41e-14 &  0.27e-17 &  0.21e-16 &  0.27e-17 &  0.21e-16 &  0.11e-15 &  0.31e-14 &  0.11e-15 &  0.31e-14 \\
 & $\psi_2$ &  0.23e-15 &  0.17e-14 &  0.23e-15 &  0.17e-14 &  0.87e-17 &  0.24e-15 &  0.87e-17 &  0.24e-15 &  0.17e-15 &  0.13e-14 &  0.17e-15 &  0.13e-14 \\
\hline
\multicolumn{14}{c}{\textbf{Measure: Deviation~~~~Grid type: Chebyshev}}\\
\hline
&& \multicolumn{12}{c}{Network Parameters}\\
&& \multicolumn{4}{c}{60} & \multicolumn{4}{c}{120} & \multicolumn{4}{c}{240} \\
\cline{3-6}\cline{7-10}\cline{11-14}
&& \multicolumn{2}{c}{Training} & \multicolumn{2}{c}{Test} & \multicolumn{2}{c}{Training} & \multicolumn{2}{c}{Test} & \multicolumn{2}{c}{Training} & \multicolumn{2}{c}{Test} \\
\cline{3-4} \cline{5-6} \cline{7-8} \cline{9-10} \cline{11-12} \cline{13-14}
$M_{\rm tr}$& Solution & MSD & MXD & MSD & MXD & MSD & MXD & MSD & MXD & MSD & MXD & MSD & MXD \\
\hline
70 & $\psi_1$ &  0.92e-19 &  0.37e-18 &  0.10e-18 &  0.38e-18 & & & & & & & & \\
 & $\psi_2$ &  0.44e-19 &  0.24e-18 &  0.43e-19 &  0.27e-18 & & & & & & & & \\
130 & $\psi_1$ &  0.53e-19 &  0.50e-18 &  0.72e-19 &  0.51e-18 &  0.13e-20 &  0.52e-20 &  0.13e-20 &  0.52e-20 & & & & \\
 & $\psi_2$ &  0.76e-19 &  0.42e-18 &  0.97e-19 &  0.43e-18 &  0.20e-20 &  0.14e-19 &  0.20e-20 &  0.14e-19 & & & & \\
250 & $\psi_1$ &  0.60e-19 &  0.26e-18 &  0.76e-19 &  0.26e-18 &  0.28e-20 &  0.16e-19 &  0.40e-20 &  0.16e-19 &  0.30e-21 &  0.17e-20 &  0.42e-21 &  0.17e-20 \\
 & $\psi_2$ &  0.96e-19 &  0.51e-18 &  0.11e-18 &  0.51e-18 &  0.40e-20 &  0.19e-19 &  0.54e-20 &  0.19e-19 &  0.29e-21 &  0.18e-20 &  0.36e-21 &  0.18e-20 \\
\hline
\multicolumn{14}{c}{\textbf{Measure: Error~~~~Grid type: Equidistant}}\\
\hline
&& \multicolumn{12}{c}{Network Parameters}\\
&& \multicolumn{4}{c}{60} & \multicolumn{4}{c}{120} & \multicolumn{4}{c}{240} \\
\cline{3-6}\cline{7-10}\cline{11-14}
&& \multicolumn{2}{c}{Training} & \multicolumn{2}{c}{Test} & \multicolumn{2}{c}{Training} & \multicolumn{2}{c}{Test} & \multicolumn{2}{c}{Training} & \multicolumn{2}{c}{Test} \\
\cline{3-4} \cline{5-6} \cline{7-8} \cline{9-10} \cline{11-12} \cline{13-14}
$M_{\rm tr}$& & MSE & MXE & MSE & MXE & MSE & MXE & MSE & MXE & MSE & MXE & MSE & MXE \\
\hline
70  & &  0.15e-17 &  0.17e-16 &  0.11e-05 &  0.26e-03 & & & & & & & & \\
130 & &  0.98e-18 &  0.16e-16 &  0.17e-10 &  0.60e-08 &  0.31e-18 &  0.11e-16 &  0.55e-09 &  0.19e-06 & & & & \\
250 & &  0.31e-16 &  0.21e-14 &  0.68e-13 &  0.43e-10 &  0.28e-18 &  0.66e-17 &  0.41e-14 &  0.28e-11 &  0.11e-18 &  0.80e-18 &  0.54e-13 &  0.37e-10 \\
\hline
\multicolumn{14}{c}{\textbf{Measure: Error~~~~Grid type: Chebyshev}}\\
\hline
&& \multicolumn{12}{c}{Network Parameters}\\
&& \multicolumn{4}{c}{60} & \multicolumn{4}{c}{120} & \multicolumn{4}{c}{240} \\
\cline{3-6}\cline{7-10}\cline{11-14}
&& \multicolumn{2}{c}{Training} & \multicolumn{2}{c}{Test} & \multicolumn{2}{c}{Training} & \multicolumn{2}{c}{Test} & \multicolumn{2}{c}{Training} & \multicolumn{2}{c}{Test} \\
\cline{3-4} \cline{5-6} \cline{7-8} \cline{9-10} \cline{11-12} \cline{13-14}
$M_{\rm tr}$& & MSE & MXE & MSE & MXE & MSE & MXE & MSE & MXE & MSE & MXE & MSE & MXE \\
\hline
70  & &  0.86e-16 &  0.77e-15 &  0.49e-16 &  0.90e-15 & & & & & & & & \\
130 & &  0.33e-16 &  0.24e-15 &  0.24e-16 &  0.23e-15 &  0.13e-17 &  0.15e-16 &  0.65e-18 &  0.13e-16 & & & & \\
250 & &  0.83e-17 &  0.43e-16 &  0.75e-17 &  0.41e-16 &  0.71e-18 &  0.26e-17 &  0.78e-18 &  0.26e-17 &  0.13e-18 &  0.41e-18 &  0.12e-18 &  0.36e-18 \\
\hline
\end{tabular}
\caption{Performance measures for all scenarios of Test Problem 5.}
\label{tab:Res_Prob_5}
\end{table}



\begin{table}[h!]
\setlength{\tabcolsep}{2pt}
\centering
\scriptsize
\begin{tabular}{@{\extracolsep{3pt}}ccccccccccccc@{}}
\hline
\multicolumn{13}{c}{\textbf{Measure: Deviation~~~~Grid type: Equidistant}}\\
\hline
& \multicolumn{12}{c}{Network Parameters} \\
& \multicolumn{4}{c}{180} & \multicolumn{4}{c}{270} & \multicolumn{4}{c}{360} \\
\cline{2-5} \cline{6-9} \cline{10-13}
& \multicolumn{2}{c}{Training} & \multicolumn{2}{c}{Test} & \multicolumn{2}{c}{Training} & \multicolumn{2}{c}{Test} & \multicolumn{2}{c}{Training} & \multicolumn{2}{c}{Test} \\
\cline{2-3} \cline{4-5} \cline{6-7} \cline{8-9} \cline{10-11} \cline{12-13}
$M_{\rm tr}$ & MSD & MXD & MSD & MXD & MSD & MXD & MSD & MXD & MSD & MXD & MSD & MXD \\
\hline
180 &  0.15e-19 &  0.82e-18 &  0.16e-19 &  0.85e-18 & & & & & & & & \\
270 &  0.49e-18 &  0.25e-16 &  0.51e-18 &  0.25e-16 &  0.25e-21 &  0.11e-19 &  0.25e-21 &  0.10e-19 & & & & \\
360 &  0.40e-22 &  0.11e-20 &  0.40e-22 &  0.11e-20 &  0.22e-20 &  0.99e-20 &  0.22e-20 &  0.99e-20 &  0.62e-20 &  0.26e-19 &  0.62e-20 &  0.26e-19 \\
\hline
\multicolumn{13}{c}{\textbf{Measure: Deviation~~~~Grid type: Chebyshev}}\\
\hline
& \multicolumn{12}{c}{Network Parameters} \\
& \multicolumn{4}{c}{180} & \multicolumn{4}{c}{270} & \multicolumn{4}{c}{360} \\
\cline{2-5} \cline{6-9} \cline{10-13}
& \multicolumn{2}{c}{Training} & \multicolumn{2}{c}{Test} & \multicolumn{2}{c}{Training} & \multicolumn{2}{c}{Test} & \multicolumn{2}{c}{Training} & \multicolumn{2}{c}{Test} \\
\cline{2-3} \cline{4-5} \cline{6-7} \cline{8-9} \cline{10-11} \cline{12-13}
$M_{\rm tr}$ & MSD & MXD & MSD & MXD & MSD & MXD & MSD & MXD & MSD & MXD & MSD & MXD \\
\hline
180 &  0.18e-21 &  0.11e-20 &  0.25e-21 &  0.12e-20 & & & & & & & & \\
270 &  0.18e-21 &  0.11e-20 &  0.24e-21 &  0.11e-20 &  0.38e-22 &  0.41e-21 &  0.52e-22 &  0.41e-21 & & & & \\
360 &  0.21e-22 &  0.12e-21 &  0.27e-22 &  0.12e-21 &  0.20e-22 &  0.10e-21 &  0.27e-22 &  0.10e-21 &  0.33e-21 &  0.20e-20 &  0.42e-21 &  0.20e-20 \\
\hline
\multicolumn{13}{c}{\textbf{Measure: Error~~~~Grid type: Equidistant}}\\
\hline
& \multicolumn{12}{c}{Network Parameters} \\
& \multicolumn{4}{c}{180} & \multicolumn{4}{c}{270} & \multicolumn{4}{c}{360} \\
\cline{2-5} \cline{6-9} \cline{10-13}
& \multicolumn{2}{c}{Training} & \multicolumn{2}{c}{Test} & \multicolumn{2}{c}{Training}  & \multicolumn{2}{c}{Test} & \multicolumn{2}{c}{Training} & \multicolumn{2}{c}{Test} \\
\cline{2-3} \cline{4-5} \cline{6-7} \cline{8-9} \cline{10-11} \cline{12-13}
$M_{\rm tr}$ & MSE & MXE & MSE & MXE & MSE & MXE & MSE & MXE & MSE & MXE & MSE & MXE \\
\hline
180 &  0.28e-19 &  0.41e-18 &  0.40e-15 &  0.17e-12 & & & & & & & & \\
270 &  0.10e-19 &  0.11e-18 &  0.18e-06 &  0.16e-03 &  0.84e-20 &  0.22e-18 &  0.14e-17 &  0.97e-15 & & & & \\
360 &  0.86e-20 &  0.13e-18 &  0.26e-17 &  0.23e-14 &  0.20e-19 &  0.58e-18 &  0.79e-18 &  0.52e-15 &  0.18e-19 &  0.40e-18 &  0.96e-16 &  0.92e-13 \\
\hline
\multicolumn{13}{c}{\textbf{Measure: Error~~~~Grid type: Chebyshev}}\\
\hline
& \multicolumn{12}{c}{Network Parameters} \\
& \multicolumn{4}{c}{180} & \multicolumn{4}{c}{270} & \multicolumn{4}{c}{360} \\
\cline{2-5} \cline{6-9} \cline{10-13}
& \multicolumn{2}{c}{Training} & \multicolumn{2}{c}{Test} & \multicolumn{2}{c}{Training} & \multicolumn{2}{c}{Test} & \multicolumn{2}{c}{Training} & \multicolumn{2}{c}{Test} \\
\cline{2-3} \cline{4-5} \cline{6-7} \cline{8-9} \cline{10-11} \cline{12-13}
$M_{\rm tr}$ & MSE & MXE & MSE & MXE & MSE & MXE & MSE & MXE & MSE & MXE & MSE & MXE \\
\hline
180 &  0.26e-19 &  0.17e-18 &  0.21e-19 &  0.19e-18 & & & & & & & & \\
270 &  0.31e-19 &  0.23e-18 &  0.22e-19 &  0.25e-18 &  0.12e-19 &  0.52e-19 &  0.13e-19 &  0.52e-19 & & & & \\
360 &  0.19e-19 &  0.11e-18 &  0.16e-19 &  0.11e-18 &  0.98e-20 &  0.85e-19 &  0.62e-20 &  0.93e-19 &  0.55e-19 &  0.42e-18 &  0.45e-19 &  0.43e-18 \\
\hline
\end{tabular}
\caption{Performance measures for all scenarios of Test Problem 6 with Dirichlet conditions.}
\label{tab:Res_Prob_6_DD}
\end{table}

\begin{table}[h!]
\setlength{\tabcolsep}{2pt}
\centering
\scriptsize
\begin{tabular}{@{\extracolsep{3pt}}ccccccccccccc@{}}
\hline
\multicolumn{13}{c}{\textbf{Measure: Deviation~~~~Grid type: Equidistant}}\\
\hline
& \multicolumn{12}{c}{Network Parameters} \\
& \multicolumn{4}{c}{180} & \multicolumn{4}{c}{270} & \multicolumn{4}{c}{360} \\
\cline{2-5} \cline{6-9} \cline{10-13}
& \multicolumn{2}{c}{Training} & \multicolumn{2}{c}{Test} & \multicolumn{2}{c}{Training} & \multicolumn{2}{c}{Test} & \multicolumn{2}{c}{Training} & \multicolumn{2}{c}{Test} \\
\cline{2-3} \cline{4-5} \cline{6-7} \cline{8-9} \cline{10-11} \cline{12-13}
$M_{\rm tr}$ & MSD & MXD & MSD & MXD & MSD & MXD & MSD & MXD & MSD & MXD & MSD & MXD \\
\hline
180 &  0.10e-18 &  0.54e-17 &  0.10e-18 &  0.55e-17 & & & & & & & & \\
270 &  0.80e-21 &  0.16e-19 &  0.80e-21 &  0.16e-19 &  0.22e-20 &  0.16e-18 &  0.18e-20 &  0.16e-18 & & & & \\
360 &  0.59e-20 &  0.36e-19 &  0.59e-20 &  0.36e-19 &  0.82e-16 &  0.39e-15 &  0.82e-16 &  0.39e-15 &  0.24e-20 &  0.30e-19 &  0.24e-20 &  0.30e-19 \\
\hline
\multicolumn{13}{c}{\textbf{Measure: Deviation~~~~Grid type: Chebyshev}}\\
\hline
& \multicolumn{12}{c}{Network Parameters} \\
& \multicolumn{4}{c}{180} & \multicolumn{4}{c}{270} & \multicolumn{4}{c}{360} \\
\cline{2-5} \cline{6-9} \cline{10-13}
& \multicolumn{2}{c}{Training} & \multicolumn{2}{c}{Test} & \multicolumn{2}{c}{Training} & \multicolumn{2}{c}{Test} & \multicolumn{2}{c}{Training} & \multicolumn{2}{c}{Test} \\
\cline{2-3} \cline{4-5} \cline{6-7} \cline{8-9} \cline{10-11} \cline{12-13}
$M_{\rm tr}$ & MSD & MXD & MSD & MXD & MSD & MXD & MSD & MXD & MSD & MXD & MSD & MXD \\
\hline
180 &  0.61e-19 &  0.37e-18 &  0.75e-19 &  0.38e-18 & & & & & & & & \\
270 &  0.62e-20 &  0.60e-19 &  0.88e-20 &  0.60e-19 &  0.16e-20 &  0.98e-20 &  0.22e-20 &  0.98e-20 & & & & \\
360 &  0.37e-20 &  0.29e-19 &  0.52e-20 &  0.29e-19 &  0.12e-20 &  0.89e-20 &  0.16e-20 &  0.89e-20 &  0.13e-20 &  0.95e-20 &  0.18e-20 &  0.96e-20 \\
\hline
\multicolumn{13}{c}{\textbf{Measure: Error~~~~Grid type: Equidistant}}\\
\hline
& \multicolumn{12}{c}{Network Parameters} \\
& \multicolumn{4}{c}{180} & \multicolumn{4}{c}{270} & \multicolumn{4}{c}{360} \\
\cline{2-5} \cline{6-9} \cline{10-13}
& \multicolumn{2}{c}{Training} & \multicolumn{2}{c}{Test} & \multicolumn{2}{c}{Training}  & \multicolumn{2}{c}{Test} & \multicolumn{2}{c}{Training} & \multicolumn{2}{c}{Test} \\
\cline{2-3} \cline{4-5} \cline{6-7} \cline{8-9} \cline{10-11} \cline{12-13}
$M_{\rm tr}$ & MSE & MXE & MSE & MXE & MSE & MXE & MSE & MXE & MSE & MXE & MSE & MXE \\
\hline
180 &  0.49e-18 &  0.32e-17 &  0.28e-15 &  0.12e-12 & & & & & & & & \\
270 &  0.14e-18 &  0.92e-18 &  0.15e-17 &  0.93e-15 &  0.25e-18 &  0.43e-17 &  0.16e-17 &  0.81e-15 & & & & \\
360 &  0.12e-17 &  0.54e-16 &  0.34e-17 &  0.17e-14 &  0.14e-17 &  0.58e-16 &  0.29e-17 &  0.11e-14 &  0.40e-18 &  0.93e-17 &  0.72e-18 &  0.25e-15 \\
\hline
\multicolumn{13}{c}{\textbf{Measure: Error~~~~Grid type: Chebyshev}}\\
\hline
& \multicolumn{12}{c}{Network Parameters} \\
& \multicolumn{4}{c}{180} & \multicolumn{4}{c}{270} & \multicolumn{4}{c}{360} \\
\cline{2-5} \cline{6-9} \cline{10-13}
& \multicolumn{2}{c}{Training} & \multicolumn{2}{c}{Test} & \multicolumn{2}{c}{Training} & \multicolumn{2}{c}{Test} & \multicolumn{2}{c}{Training} & \multicolumn{2}{c}{Test} \\
\cline{2-3} \cline{4-5} \cline{6-7} \cline{8-9} \cline{10-11} \cline{12-13}
$M_{\rm tr}$ & MSE & MXE & MSE & MXE & MSE & MXE & MSE & MXE & MSE & MXE & MSE & MXE \\
\hline
180 &  0.54e-17 &  0.59e-16 &  0.37e-17 &  0.61e-16 & & & & & & & & \\
270 &  0.11e-17 &  0.54e-17 &  0.11e-17 &  0.54e-17 &  0.25e-18 &  0.13e-17 &  0.30e-18 &  0.13e-17 & & & & \\
360 &  0.45e-18 &  0.19e-17 &  0.43e-18 &  0.19e-17 &  0.49e-18 &  0.32e-17 &  0.39e-18 &  0.32e-17 &  0.14e-18 &  0.68e-18 &  0.18e-18 &  0.68e-18 \\
\hline
\end{tabular}
\caption{Performance measures for all scenarios of Test Problem 6 with Dirichlet-Neumann conditions.}
\label{tab:Res_Prob_6_DN}
\end{table}


\begin{table}[h!]
\setlength{\tabcolsep}{2pt}
\centering
\scriptsize
\begin{tabular}{@{\extracolsep{3pt}}ccccccccccccc@{}}
\hline
\multicolumn{13}{c}{\textbf{Measure: Deviation~~~~Grid type: Equidistant}}\\
\hline
& \multicolumn{12}{c}{Network Parameters} \\
& \multicolumn{4}{c}{180} & \multicolumn{4}{c}{270} & \multicolumn{4}{c}{360} \\
\cline{2-5} \cline{6-9} \cline{10-13}
& \multicolumn{2}{c}{Training} & \multicolumn{2}{c}{Test} & \multicolumn{2}{c}{Training} & \multicolumn{2}{c}{Test} & \multicolumn{2}{c}{Training} & \multicolumn{2}{c}{Test} \\
\cline{2-3} \cline{4-5} \cline{6-7} \cline{8-9} \cline{10-11} \cline{12-13}
$M_{\rm tr}$ & MSD & MXD & MSD & MXD & MSD & MXD & MSD & MXD & MSD & MXD & MSD & MXD \\
\hline
180 &  0.16e-16 &  0.64e-15 &  0.14e-16 &  0.64e-15 & & & & & & & & \\
270 &  0.37e-19 &  0.16e-17 &  0.35e-19 &  0.17e-17 &  0.83e-01 &  0.25E+01 &  0.79e-01 &  0.25E+01 & & & & \\
360 &  0.23e-18 &  0.12e-16 &  0.22e-18 &  0.12e-16 &  0.74e-17 &  0.39e-15 &  0.71e-17 &  0.39e-15 &  0.20e-18 &  0.10e-16 &  0.19e-18 &  0.10e-16 \\
\hline
\multicolumn{13}{c}{\textbf{Measure: Deviation~~~~Grid type: Chebyshev}}\\
\hline
& \multicolumn{12}{c}{Network Parameters} \\
& \multicolumn{4}{c}{180} & \multicolumn{4}{c}{270} & \multicolumn{4}{c}{360} \\
\cline{2-5} \cline{6-9} \cline{10-13}
& \multicolumn{2}{c}{Training} & \multicolumn{2}{c}{Test} & \multicolumn{2}{c}{Training} & \multicolumn{2}{c}{Test} & \multicolumn{2}{c}{Training} & \multicolumn{2}{c}{Test} \\
\cline{2-3} \cline{4-5} \cline{6-7} \cline{8-9} \cline{10-11} \cline{12-13}
$M_{\rm tr}$ & MSD & MXD & MSD & MXD & MSD & MXD & MSD & MXD & MSD & MXD & MSD & MXD \\
\hline
180 &  0.19e-17 &  0.26e-16 &  0.47e-18 &  0.26e-16 & & & & & & & & \\
270 &  0.91e-16 &  0.13e-14 &  0.23e-16 &  0.13e-14 &  0.25e-15 &  0.34e-14 &  0.62e-16 &  0.34e-14 & & & & \\
360 &  0.81e-17 &  0.11e-15 &  0.20e-17 &  0.11e-15 &  0.24e-19 &  0.34e-18 &  0.62e-20 &  0.34e-18 &  0.98e-20 &  0.13e-18 &  0.31e-20 &  0.13e-18 \\
\hline
\multicolumn{13}{c}{\textbf{Measure: Error~~~~Grid type: Equidistant}}\\
\hline
& \multicolumn{12}{c}{Network Parameters} \\
& \multicolumn{4}{c}{180} & \multicolumn{4}{c}{270} & \multicolumn{4}{c}{360} \\
\cline{2-5} \cline{6-9} \cline{10-13}
& \multicolumn{2}{c}{Training} & \multicolumn{2}{c}{Test} & \multicolumn{2}{c}{Training}  & \multicolumn{2}{c}{Test} & \multicolumn{2}{c}{Training} & \multicolumn{2}{c}{Test} \\
\cline{2-3} \cline{4-5} \cline{6-7} \cline{8-9} \cline{10-11} \cline{12-13}
$M_{\rm tr}$ & MSE & MXE & MSE & MXE & MSE & MXE & MSE & MXE & MSE & MXE & MSE & MXE \\
\hline
180 &  0.93e-19 &  0.11e-17 &  0.46e-14 &  0.21e-11 & & & & & & & & \\
270 &  0.16e-18 &  0.40e-17 &  0.28e-15 &  0.20e-12 &  0.76e-19 &  0.32e-18 &  0.46e-04 &  0.46e-01 & & & & \\
360 &  0.44e-19 &  0.14e-17 &  0.13e-17 &  0.11e-14 &  0.39e-18 &  0.17e-16 &  0.23e-17 &  0.15e-14 &  0.15e-19 &  0.31e-18 &  0.13e-16 &  0.11e-13 \\
\hline
\multicolumn{13}{c}{\textbf{Measure: Error~~~~Grid type: Chebyshev}}\\
\hline
& \multicolumn{12}{c}{Network Parameters} \\
& \multicolumn{4}{c}{180} & \multicolumn{4}{c}{270} & \multicolumn{4}{c}{360} \\
\cline{2-5} \cline{6-9} \cline{10-13}
& \multicolumn{2}{c}{Training} & \multicolumn{2}{c}{Test} & \multicolumn{2}{c}{Training} & \multicolumn{2}{c}{Test} & \multicolumn{2}{c}{Training} & \multicolumn{2}{c}{Test} \\
\cline{2-3} \cline{4-5} \cline{6-7} \cline{8-9} \cline{10-11} \cline{12-13}
$M_{\rm tr}$ & MSE & MXE & MSE & MXE & MSE & MXE & MSE & MXE & MSE & MXE & MSE & MXE \\
\hline
180 &  0.13e-18 &  0.96e-18 &  0.92e-19 &  0.93e-18 & & & & & & & & \\
270 &  0.76e-18 &  0.45e-17 &  0.68e-18 &  0.49e-17 &  0.19e-17 &  0.98e-17 &  0.17e-17 &  0.99e-17 & & & & \\
360 &  0.19e-18 &  0.80e-18 &  0.19e-18 &  0.80e-18 &  0.17e-19 &  0.11e-18 &  0.13e-19 &  0.12e-18 &  0.51e-19 &  0.21e-18 &  0.52e-19 &  0.21e-18 \\
\hline
\end{tabular}
\caption{Performance measures for all scenarios of Test Problem 6 with Cauchy conditions.}
\label{tab:Res_Prob_6_C}
\end{table}

\begin{table}[h!]
\setlength{\tabcolsep}{2pt}
\centering
\scriptsize
\begin{tabular}{@{\extracolsep{3pt}}ccccccccccccc@{}}
\hline
\multicolumn{13}{c}{\textbf{Measure: Deviation~~~~Grid type: Equidistant}}\\
\hline
& \multicolumn{12}{c}{Network Parameters} \\
& \multicolumn{4}{c}{180} & \multicolumn{4}{c}{270} & \multicolumn{4}{c}{360} \\
\cline{2-5} \cline{6-9} \cline{10-13}
& \multicolumn{2}{c}{Training} & \multicolumn{2}{c}{Test} & \multicolumn{2}{c}{Training} & \multicolumn{2}{c}{Test} & \multicolumn{2}{c}{Training} & \multicolumn{2}{c}{Test} \\
\cline{2-3} \cline{4-5} \cline{6-7} \cline{8-9} \cline{10-11} \cline{12-13}
$M_{\rm tr}$ & MSD & MXD & MSD & MXD & MSD & MXD & MSD & MXD & MSD & MXD & MSD & MXD \\
\hline
180 &  0.13e-18 &  0.10e-16 &  0.85e-19 &  0.10e-16 & & & & & & & & \\
270 &  0.27e-19 &  0.61e-17 &  0.13e-19 &  0.61e-17 &  0.40e-19 &  0.30e-17 &  0.31e-19 &  0.30e-17 & & & & \\
360 &  0.60e-20 &  0.35e-19 &  0.59e-20 &  0.35e-19 &  0.12e-18 &  0.66e-17 &  0.11e-18 &  0.66e-17 &  0.14e-17 &  0.50e-15 &  0.56e-18 &  0.50e-15 \\
\hline
\multicolumn{13}{c}{\textbf{Measure: Deviation~~~~Grid type: Chebyshev}}\\
\hline
& \multicolumn{12}{c}{Network Parameters} \\
& \multicolumn{4}{c}{180} & \multicolumn{4}{c}{270} & \multicolumn{4}{c}{360} \\
\cline{2-5} \cline{6-9} \cline{10-13}
& \multicolumn{2}{c}{Training} & \multicolumn{2}{c}{Test} & \multicolumn{2}{c}{Training} & \multicolumn{2}{c}{Test} & \multicolumn{2}{c}{Training} & \multicolumn{2}{c}{Test} \\
\cline{2-3} \cline{4-5} \cline{6-7} \cline{8-9} \cline{10-11} \cline{12-13}
$M_{\rm tr}$ & MSD & MXD & MSD & MXD & MSD & MXD & MSD & MXD & MSD & MXD & MSD & MXD \\
\hline
180 &  0.18e-20 &  0.13e-19 &  0.25e-20 &  0.13e-19 & & & & & & & & \\
270 &  0.36e-20 &  0.21e-19 &  0.37e-20 &  0.21e-19 &  0.95e-21 &  0.36e-20 &  0.91e-21 &  0.36e-20 & & & & \\
360 &  0.62e-21 &  0.39e-20 &  0.81e-21 &  0.39e-20 &  0.31e-21 &  0.11e-20 &  0.27e-21 &  0.11e-20 &  0.63e-23 &  0.56e-22 &  0.82e-23 &  0.57e-22 \\
\hline
\multicolumn{13}{c}{\textbf{Measure: Error~~~~Grid type: Equidistant}}\\
\hline
& \multicolumn{12}{c}{Network Parameters} \\
& \multicolumn{4}{c}{180} & \multicolumn{4}{c}{270} & \multicolumn{4}{c}{360} \\
\cline{2-5} \cline{6-9} \cline{10-13}
& \multicolumn{2}{c}{Training} & \multicolumn{2}{c}{Test} & \multicolumn{2}{c}{Training}  & \multicolumn{2}{c}{Test} & \multicolumn{2}{c}{Training} & \multicolumn{2}{c}{Test} \\
\cline{2-3} \cline{4-5} \cline{6-7} \cline{8-9} \cline{10-11} \cline{12-13}
$M_{\rm tr}$ & MSE & MXE & MSE & MXE & MSE & MXE & MSE & MXE & MSE & MXE & MSE & MXE \\
\hline
180 &  0.36e-19 &  0.48e-18 &  0.23e-15 &  0.10e-12 & & & & & & & & \\
270 &  0.46e-19 &  0.85e-18 &  0.40e-16 &  0.30e-13 &  0.37e-19 &  0.98e-18 &  0.14e-15 &  0.96e-13 & & & & \\
360 &  0.14e-18 &  0.13e-17 &  0.45e-18 &  0.26e-15 &  0.25e-19 &  0.58e-18 &  0.16e-14 &  0.92e-12 &  0.62e-19 &  0.65e-18 &  0.20e-14 &  0.19e-11 \\
\hline
\multicolumn{13}{c}{\textbf{Measure: Error~~~~Grid type: Chebyshev}}\\
\hline
& \multicolumn{12}{c}{Network Parameters} \\
& \multicolumn{4}{c}{180} & \multicolumn{4}{c}{270} & \multicolumn{4}{c}{360} \\
\cline{2-5} \cline{6-9} \cline{10-13}
& \multicolumn{2}{c}{Training} & \multicolumn{2}{c}{Test} & \multicolumn{2}{c}{Training} & \multicolumn{2}{c}{Test} & \multicolumn{2}{c}{Training} & \multicolumn{2}{c}{Test} \\
\cline{2-3} \cline{4-5} \cline{6-7} \cline{8-9} \cline{10-11} \cline{12-13}
$M_{\rm tr}$ & MSE & MXE & MSE & MXE & MSE & MXE & MSE & MXE & MSE & MXE & MSE & MXE \\
\hline
180 &  0.13e-18 &  0.61e-18 &  0.13e-18 &  0.83e-18 & & & & & & & & \\
270 &  0.18e-18 &  0.12e-17 &  0.13e-18 &  0.13e-17 &  0.20e-18 &  0.60e-18 &  0.22e-18 &  0.63e-18 & & & & \\
360 &  0.39e-19 &  0.41e-18 &  0.30e-19 &  0.45e-18 &  0.13e-18 &  0.62e-18 &  0.13e-18 &  0.63e-18 &  0.63e-20 &  0.25e-19 &  0.64e-20 &  0.26e-19 \\
\hline
\end{tabular}
\caption{Performance measures for all scenarios of Test Problem 6 with Neumann conditions.}
\label{tab:Res_Prob_6_NN}
\end{table}





\clearpage
\section{Comparisons using Adam solver}
\label{appendix:AdamOptimizer}

\begin{table}[h!] 
\centering
\small
\begin{tabular}{cccccccccccc}  
\hline
 & & & & & & & \multicolumn{2}{c}{Training dataset} & & \multicolumn{2}{c}{Test dataset} \\
\cline{8-9} \cline{11-12}
Problem & Conditions & Method & $|\theta|$ & $M_{\rm tr}$ & Grid &  & MSD & MXD & & MSD & MXD \\ 
\hline
1 & D & Merlin & 144 & 160 & Ch & & 0.22e-17 & 0.36e-16 & & 0.25e-18 & 0.36e-16 \\
  & & Adam & & & & & 0.19e-08 & 0.99e-08 & & 0.20e-08 & 0.10e-07 \\
\hline
2 & D-D & Merlin & 90 & 270 & Ch & & 0.83e-23 & 0.98e-22 & & 0.12e-22 & 0.98e-22 \\
  & & Adam & & & & & 0.20e-03 & 0.87e-03 & & 0.24e-03 & 0.87e-03 \\
\cline{2-12}
& D-N & Merlin & 90 & 90 & Ch & & 0.17e-21 & 0.14e-20 & & 0.24e-21 & 0.14e-20 \\
  & & Adam & & & & & 0.16e-01 & 0.66e-01 & & 0.24e-01 & 0.66e-01 \\
\cline{2-12}
& N-N & Merlin & 90 & 270 & Ch & & 0.61e-23 & 0.25e-22 & & 0.65e-23 & 0.26e-22 \\
  & & Adam & & & & & 0.76e-01 & 0.12e+00 & & 0.86e-01 & 0.12e+00 \\
\cline{2-12}  
& C & Merlin & 90 & 180 & Ch & & 0.43e-22 & 0.58e-21 & & 0.13e-22 & 0.58e-21 \\
  & & Adam & & & & & 0.12e-00 & 0.14e+01 & & 0.37e-01 & 0.14e+01 \\
\cline{2-12}   
& R & Merlin & 180 & 270 & Ch & & 0.40e-23 & 0.37-22 & & 0.59e-23 & 0.37e-22 \\
  & & Adam & & & & & 0.22e-01 & 0.59e-01 & & 0.31e-01 & 0.59e-01 \\
\hline
3 & C & Merlin & 180 & 180 & Eq & & 0.45e-18 & 0.21e-17 & & 0.45e-18 & 0.21e-17 \\
  & & Adam & & & & & 0.12e+00 & 0.68e+00 & & 0.12e+00 & 0.68e+00 \\
\hline
4 & D & Merlin & 240 & 250 & Ch & $\psi_1$ & 0.15e-18 & 0.12e-17 & & 0.61e-19 & 0.12e-17 \\
  & & & & & & $\psi_2$ & 0.12e-17 & 0.95e-17 & & 0.43e-18 & 0.95e-17 \\
  & & Adam & & & & $\psi_1$ & 0.56e-03 & 0.41e-02 & & 0.23e-03 & 0.41e-02 \\
  & & & & & & $\psi_2$ & 0.39e-02 & 0.32e-01 & & 0.14e-02 & 0.32e-01 \\
\hline
5 & D & Merlin & 240 & 250 & Ch & $\psi_1$ & 0.30e-21 & 0.17e-20 & & 0.42e-21 & 0.17e-20 \\
  & & & & & & $\psi_2$ & 0.29e-21 & 0.18e-20 & & 0.36e-21 & 0.18e-20 \\
  & & Adam & & & & $\psi_1$ & 0.85e-06 & 0.24e-05 & & 0.84e-06 & 0.24e-05 \\
  & & & & & & $\psi_2$ & 0.17e-05 & 0.50e-05 & & 0.17e-05 & 0.50e-05 \\
\hline
6 & D-D & Merlin & 270 & 360 & Ch & & 0.20e-22 & 0.10e-21 & & 0.27e-22 & 0.10e-21 \\
  & & Adam & & & & & 0.15e-03 & 0.76e-03 & & 0.17e-03 & 0.76e-03 \\
\cline{2-12}  
& D-N & Merlin & 360 & 360 & Ch & & 0.13e-20 & 0.95e-20 & & 0.18e-20 & 0.96e-20 \\
  & & Adam & & & & & 0.19e-02 & 0.88e-02 & & 0.27e-02 & 0.88e-02 \\
\cline{2-12}  
& N-N & Merlin & 360 & 360 & Ch & & 0.63e-23 & 0.56e-22 & & 0.82e-23 & 0.57e-22 \\
  & & Adam & & & & & 0.40e-01 & 0.70e-01 & & 0.44e-01 & 0.70e-01 \\
\cline{2-12}  
& C & Merlin & 270 & 360 & Ch & & 0.24e-19 & 0.34e-18 & & 0.62e-20 & 0.34e-18 \\
  & & Adam & & & & & 0.25e+00 & 0.26e+01 & & 0.84e-01 & 0.26e+01 \\
\hline
\multicolumn{11}{l}{\scriptsize Abbreviations: D=Dirichlet, N=Neumann, D-N=Mixed Dirichlet-Neumann, R=Robin, C=Cauchy, Ch=Chebyshev, Eq=Equidistant}
\end{tabular}
\caption{Mean (MSD) and maximum (MXD) squared deviation of the best neural form optimized using \texttt{Merlin}, and the corresponding Adam solver per test problem and condition type. The corresponding grid type, number of neural form parameters $|\theta|$, and number of training points $M_{\rm tr}$ are also reported.}
\label{tab:compactResAdam}
\end{table}

\end{document}